%% file: main.tex
\documentclass[journal]{IEEEtran}

\usepackage{amsmath,amssymb,amsfonts}
\usepackage{algorithmic}
\usepackage{multirow}
\usepackage{makecell}
\usepackage{bm}
\usepackage{amsthm}
\usepackage{color, xspace}

\usepackage{algorithm}
\newcommand{\INPUT}{\item[\textbf{Input:}]}
\newcommand{\OUTPUT}{\item[\textbf{Output:}]}

\usepackage{hyperref}
\hypersetup{
    colorlinks=true,
    linkcolor=green,
    citecolor=green,
    urlcolor=blue
}
\usepackage{booktabs}
\usepackage{tikz}
\usetikzlibrary{fadings}
\usetikzlibrary{mindmap,trees}
\usetikzlibrary{arrows,automata,shapes,positioning,shadows,trees}
\usetikzlibrary{shapes,shadows}
\usetikzlibrary{decorations.pathmorphing}
\usetikzlibrary{shapes.arrows}
\usetikzlibrary{calc,shapes, positioning}
\usetikzlibrary{decorations.text}
\usetikzlibrary{matrix,chains,positioning,decorations.pathreplacing,arrows}
\usetikzlibrary{bayesnet}
\usepackage[edges]{forest}
\usetikzlibrary{shadows.blur}
\usetikzlibrary{shapes.geometric}

\begin{document}
\title{A Comprehensive Survey on Imbalanced Data Learning}

\author{Xinyi~Gao,~\IEEEmembership{Student Member,~IEEE,}
Dongting~Xie,
Yihang~Zhang,
Zhengren~Wang,
Chong~Chen,\\
Conghui~He,
Hongzhi~Yin,~\IEEEmembership{Senior Member,~IEEE,}
Wentao~Zhang,~\IEEEmembership{Member,~IEEE}

\thanks{This work is supported by the National Key R\&D Program of China (2024YFA1014003), National Natural Science Foundation of China (92470121, 62402016), and High-performance Computing Platform of Peking University.
}

\thanks{Xinyi Gao and Hongzhi Yin are with the School of Electrical Engineering and Computer Science, The University of Queensland, Brisbane, Australia. 
Dongting Xie is with the School of Information Science and Engineering, Yunnan University, Kunming, China. 
Yihang Zhang is with the School of Mathematical Sciences, Peking University, Beijing, China. 
Zhengren Wang is with the Academy for Advanced Interdisciplinary Studies, Peking University, Beijing, China. 
Chong Chen is with Huawei Cloud BU, Beijing, China. 
Conghui He is with the Shanghai AI Laboratory, Shanghai, China. 
Wentao Zhang is with the Center of Machine Learning Research, Peking University, Beijing, China.}

\thanks{Corresponding author: Wentao Zhang (e-mail: wentao.zhang@pku.edu.cn) and Hongzhi Yin (e-mail: h.yin1@uq.edu.au).}

}

% \markboth{Journal of \LaTeX\ Class Files,~Vol.~14, No.~8, August~2015}%
% {Xinyi Gao \MakeLowercase{\textit{et al.}}: Bare Demo of IEEEtran.cls for IEEE Journals}

\maketitle
\begin{abstract}
With the expansion of data availability, machine learning (ML) has achieved remarkable breakthroughs in both academia and industry. However, imbalanced data distributions are prevalent in various types of raw data and severely hinder the performance of ML by biasing the decision-making processes. To deepen the understanding of imbalanced data and facilitate the related research and applications, this survey systematically analyzes various real-world data formats and concludes existing researches for different data formats into four distinct categories: data re-balancing, feature representation, training strategy, and ensemble learning. This structured analysis helps researchers comprehensively understand the pervasive nature of imbalance across diverse data formats, thereby paving a clearer path toward achieving specific research goals. We provide an overview of relevant open-source libraries, spotlight current challenges, and offer novel insights aimed at fostering future advancements in this critical area of study.
\end{abstract}


\IEEEpeerreviewmaketitle

\input{1.Intro}
\input{3.1Datarebalancing}

\input{3.2Featurerepresentation}
\input{3.3Trainingstrategy}

\input{3.4Ensemble}

\input{4.Data}
\input{2.Evaluation}
\input{9Exp}
\input{6.Packages}
\input{7.Challenges}
\input{8.Conclusion}

\bibliographystyle{IEEEtran}
\bibliography{IEEEabrv,ref}

\input{bio}

\end{document}

%% file: 1.Intro.tex
\section{Introduction}
\label{sec:intro}

The advent of expansive data availability has propelled machine learning (ML) to the forefront of technological advancements in both academia and industry. These ML models are intricately designed to fit specific data distributions and {deployed across a wide range of downstream tasks}, ranging from predictive analytics to automated decision-making systems. {Consequently, their performance is heavily influenced} by the quality and distribution of the data used during training. {When the data is representative, diverse, and carefully preprocessed, models achieve not only high accuracy but also robustness and strong generalization} across different settings and challenges.

However, natural data distributions are inherently complex and frequently flawed. Among the myriad challenges they present, the issue of imbalanced data distribution is particularly prominent, reflecting the widespread and natural disparities encountered across various fields. 
{In financial fraud detection, for example, fraudulent transactions occur far less frequently than legitimate ones, due to long-standing financial regulations and continuous improvements in system integrity. The Credit Card Fraud Detection system exemplifies this, with fraudulent cases comprising only 0.17\% of the total.
In healthcare, imbalance arises from the natural prevalence of diseases in the population. Common conditions dominate medical records, while rare but critical diseases are significantly underrepresented. For instance, the NIH Chest X-ray14 dataset contains over 100,000 images, yet conditions such as Pneumothorax appear in less than 1\% of cases.
In industrial manufacturing, quality control systems are designed to minimize defects through efficient processes and stringent standards. As a result, defects are rare by design, leading to imbalanced datasets. 
The NEU Surface Defect Dataset, for instance, includes only 300 images per defect type, while real-world systems routinely process thousands of non-defective samples.}
{Beyond these general examples, imbalance is also pervasive in broader scientific and engineering applications. In energy management, faults in power grids or renewable systems are rare compared to normal operations. In emergency management, rare but high-impact events such as earthquakes, floods, or industrial accidents are naturally underrepresented. In environmental management, events like oil spills, algal blooms, or pollutant spikes occur sparsely compared to routine monitoring records. In agriculture and horticulture, healthy crops dominate datasets, while rare diseases, pest outbreaks, or nutrient deficiencies are significantly under-sampled. These imbalances are not mere data collection artifacts but stem from structural characteristics of the respective domains. Consequently, these scenarios not only complicate ML model training but also place greater demands on the robustness and reliability of the systems.}

Generally, imbalanced data distributions significantly impact the performance and utility of ML models. {In imbalanced settings, majority classes refer to categories with relatively larger numbers of samples, while minority classes denote categories with comparatively fewer samples. Models often achieve strong performance on majority classes but struggle on minority classes, leading to unclear distribution boundaries.} Consequently, although ML models may exhibit satisfactory overall performance, their effectiveness is notably reduced within these {minority classes}. However, these {minority classes} often hold the greater significance in real-world scenarios. For example, in medical diagnostics, the failure to detect a rare disease due to insufficient data can lead to missed diagnoses and inadequate patient care. Similarly, in financial systems, the inability to identify infrequent instances of fraud can result in significant financial losses and compromised security. This tendency of ML models to overlook rare but critical instances diminishes the utility and safety of automated decision-making systems in practical applications.

In response to these challenges, the ML community has devised a range of methods and we organize them into four fundamental categories—data re-balancing, feature representation, training strategy, and ensemble learning—each corresponding to crucial aspects of the ML process.
Data re-balancing techniques are crucial for adjusting the data distribution to better representation, utilizing approaches such as over-sampling the minority class and under-sampling the majority class. {This adjustment is critical to prevent the model from favoring the majority samples}, aligning with the data preparation phase of ML.
Feature representation strategies enhance the ability to accurately capture and represent the information relevant to minority samples. This improvement is pivotal in the feature engineering phase, enabling models to effectively learn from and make predictions for all samples.
Advanced training strategies adjust the learning algorithms to minimize their inherent bias towards the majority samples. This key adjustment during the training phase ensures that the learning process is inclusive, giving equitable consideration to all samples.
Lastly, ensemble methods, which involve combining several models, correspond to the model integration part of the ML process. These methods leverage the strengths of multiple algorithms to potentially reduce biases that may arise from imbalanced data, thereby enhancing the robustness and accuracy of the final model outputs.
By categorizing methods according to the foundational processes of ML, this taxonomy not only facilitates comprehensive field surveys but also {clarifies the motivations behind these strategies, facilitating the pursuit of specific objectives for specific method designation.}
Additionally, this survey explores the manifestation of imbalances across various data formats, including images, text, and graphs, highlighting the differences, unique challenges and required adaptations for each format. This exploration is critically important as it deepens the understanding of each data format and {facilitates the development of targeted ML strategies} for complex data format scenarios.

{While some existing surveys on imbalanced learning have offered valuable insights, they remain limited in scope. Traditional reviews \cite{he2009learning,krawczyk2016learning,haixiang2017learning} primarily focus on classical approaches such as sampling, cost-sensitive learning, and kernel-based methods, while overlooking recent advances in deep learning. Furthermore, the survey \cite{haixiang2017learning} highlights open challenges and specific application scenarios across domains ranging from management science to engineering, but lacks a systematic ML methodological overview. Other surveys, such as those on deep long-tailed learning \cite{zhang2023deep}, {are restricted to computer vision and therefore provide limited coverage of other data modalities.} In contrast, our survey presents a unified taxonomy that integrates both classical and deep learning methods, with particular emphasis on feature representation and training strategies. We also discuss the data-specific challenges across multiple formats, including image, text, and graph data, and provide comprehensive coverage of evaluation metrics and open-source libraries, making this work a timely and practical resource for researchers and practitioners.}

The contributions of this survey are summarized as follows:
\begin{itemize}
\item We provide a comprehensive survey of the literature addressing imbalanced data learning, offering a structured overview of methods that are rooted in the foundational processes of ML.
\item We conduct a thorough analysis of imbalances across various data formats, including images, text, and graphs, providing an in-depth exploration of the challenges and methodologies specific to each format.
\item We highlight resources available for tackling imbalanced data and explore current challenges and future directions. This discussion is designed to guide researchers {struggling} with imbalance issues, aiding them in developing strategies effectively and efficiently.
\end{itemize}

The structure of this survey is outlined as follows: Section II provides a thorough investigation of methodologies for handling imbalances, organized according to our taxonomy. Section III extensively discusses the manifestation of imbalances across various data formats. 
Section IV offers a detailed investigation of evaluation metrics specific to imbalanced data methods. 
Section V is dedicated to summarizing the resources available for learning from imbalanced data. Finally, Section VI presents the challenges and future directions in this field.

%% file: 3.1Datarebalancing.tex
\section{Methodology}
\label{s3}

\input{tree}

In this section, we delve into the specifics of methods for handling imbalanced data, conducting a systematic analysis across four distinct categories: data re-balancing, feature representation, training strategy, and ensemble learning. We explore the various techniques employed within each category to achieve their respective objectives. Additionally, we provide a detailed overview in Figure \ref{fig:tax}, which serves to encapsulate the scope and relationships among the methods discussed.

\subsection{Data re-balancing}
\label{s31}

\begin{figure*}[t]
\centering
\includegraphics[width=\linewidth]{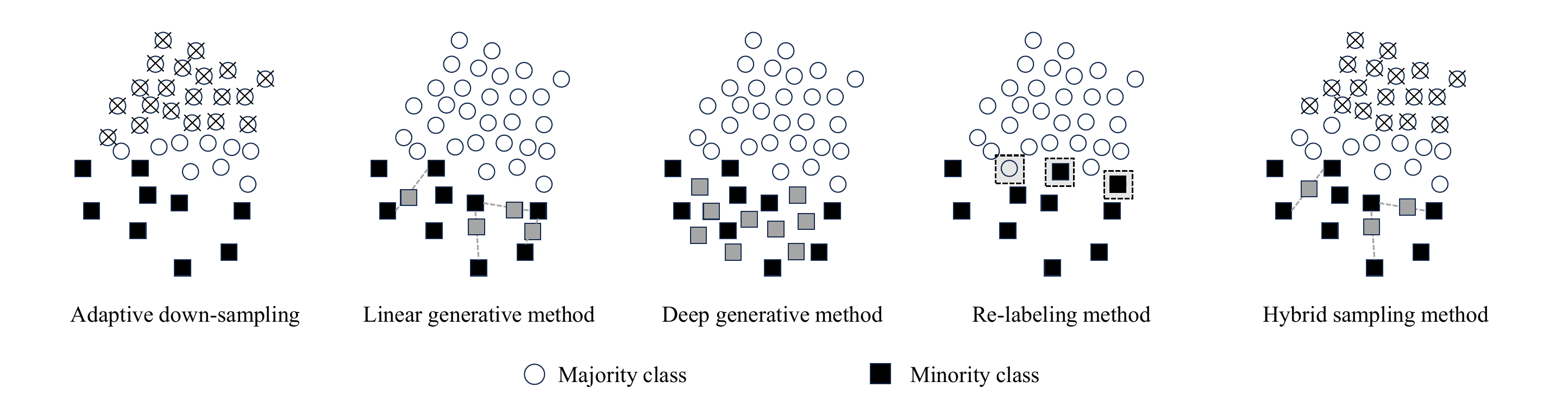}
\caption{{The data re-balancing methods. Adaptive down-sampling methods reduce the number of majority samples to balance the class distribution. Linear generative methods synthesize new minority samples through linear combinations of existing instances. Deep generative methods learn the underlying data distribution and generate synthetic samples accordingly. Relabeling methods introduce new class labels. Hybrid sampling methods simultaneously reduce the majority samples and generate {minority samples} to achieve a balanced dataset.}}
\label{fig_1}
\end{figure*}

{Data re-balancing is a key strategy to address class imbalance by modifying the training data distribution. It operates at the data level, aiming to equalize class representation to prevent models from being biased toward majority classes as shown in Fig. \ref{fig_1}.}

\subsubsection{Data generation}\
\label{s311}

One straightforward approach to address class imbalance and augment minority samples is random oversampling, which involves duplicating randomly selected {minority samples} and adding them to the dataset. However, this fundamental method, while effective in balancing class distributions, can inadvertently lead to classifier overfitting. This occurs because random oversampling generates multiple identical instances, resulting in overly specific rules. To overcome the limitations of random oversampling, more sophisticated techniques have been introduced. Linear generative methods and deep generative models offer a refined solution by creating diverse synthetic samples for minority classes. These advanced techniques enhance the learning process for machine learning models, enabling them to make more accurate predictions for {minority samples}.

\textbf{Linear generative methods} 
Linear generative methods generate synthetic minority samples by interpolating between samples or feature vectors, which includes the Synthetic Minority Oversampling Technique (SMOTE)~\cite{ref_2} and Mixup~\cite{zhang2017mixup}.

\begin{algorithm}[t]
\caption{{SMOTE Oversampling}}
\label{alg:smote}
\resizebox{\linewidth}{!}{
\begin{minipage}{\linewidth}
\begin{algorithmic}[1]
\INPUT {minority samples} $\mathcal{D}_m$, number of neighbors $k$, uniform distribution $\mathcal{U}(0,1)$
\OUTPUT Synthetic samples $\mathcal{D}_{syn}$
\FORALL{$x_i \in \mathcal{D}_m$}
    \STATE Find $k$ nearest neighbors of $x_i$ in $\mathcal{D}_m$
    \STATE Randomly select a neighbor $x_j$
    \STATE $x_{\text{new}} \gets x_i + \lambda \cdot (x_j - x_i)$, $\lambda \sim \mathcal{U}(0,1)$
    \STATE Add $x_{\text{new}}$ to $\mathcal{D}_{syn}$
\ENDFOR
\RETURN $\mathcal{D}_{syn}$
\end{algorithmic}
\end{minipage}}
\end{algorithm}

SMOTE selects k pairs of nearest minority neighbors and performs linear interpolation to create new synthetic minority samples. {The pseudo-code is shown as Algorithm \ref{alg:smote}.} This method effectively mitigates overfitting and enhances classifier learning. Nonetheless, SMOTE generates an equal number of synthetic samples for each original minority sample, disregarding neighboring sample distribution characteristics and potentially causing over-representation of certain classes. To address this, recent advancements include D-SMOTE~\cite{ref_10}, which generates synthetic samples using mean points of nearest neighbors, and N-SMOTE~\cite{ref_11}, which employs structural information in neighbor calculations. Adaptive techniques like Borderline-SMOTE only generate synthetic samples for boundary minority samples~\cite{ref_4}, and ADASYN~\cite{ref_5} adjusts sample generation based on density distribution. Moreover, MSMOTE categorizes {minority samples} into safe, border, and latent noise groups and selectively generates samples for border instances~\cite{ref_12}. With over 100 variants developed to date, SMOTE stands out as one of the most influential and effective methods in the realm of data generation for addressing class imbalance. Given the focus of our paper on a comprehensive survey for imbalanced data, we recommend readers who are particularly interested in SMOTE and its extensive family of variants to refer to a SMOTE survey~\cite{smote-comparison}.

In contrast to SMOTE, Mixup introduces a distinct approach, allowing for interpolations between samples from different classes while simultaneously considering linear interpolations of associated labels using a consistent mixing factor. However, it's important to note that the vanilla Mixup method may significantly disrupt the minority class distribution, potentially blurring the decision boundary. Recognizing this challenge, Remix~\cite{chou2020remix} offers a flexible solution by relaxing the constraint of employing the same mixing factor for both features and labels when generating virtual mixed samples for imbalanced data. In Remix, the mixing factors for labels are thoughtfully determined, accounting for the class sizes of interpolated samples and bestowing greater importance upon minority samples. This strategic choice serves the purpose of shifting the decision boundary toward the majority class and enriching the diversity of information pertaining to the minority class through the introduction of majority samples. 
Nonetheless, it is imperative to acknowledge that the random sampling of interpolated data may exacerbate data imbalance, potentially resulting in suboptimal performance. To address this concern, several sampling methods have been introduced to further bolster Remix's efficacy. MixBoost~\cite{kabra2020mixboost} incorporates an information-theoretic measure in conjunction with the trained classifier, facilitating the selection of instances in pairs. This process ensures that one instance in each pair is in close proximity to the classifier's decision boundary, while the other is positioned further away, contributing to the iterative enhancement of classifier performance. Balanced-MixUp~\cite{galdran2021balanced} employs a dual-strategy approach by concurrently sampling imbalanced instances and balanced classes from the training data. This facilitates the creation of a balanced training distribution, effectively mitigating the risk of under-fitting in minority classes. Label Occurrence-Balanced Mixup~\cite{9746299} incorporates two independent class-balanced samplers, thereby upholding statistically balanced label occurrences for each class. This meticulous approach ensures a more equitable representation of classes in mixed data pairs.

{Besides linear generation methods, recent research has explored kernel-based techniques to enhance representation capacity under class imbalance. These methods introduce non-linearity, making them particularly effective when data distributions or class boundaries are complex and non-linear. For example, Kernel-based SMOTE (K-SMOTE) \cite{mathew2015kernel} generates synthetic minority samples in the kernel-induced feature space of an SVM, leveraging the augmented Gram matrix to better align with the decision boundary and outperforming SMOTE across 51 benchmark datasets. Similarly, KernelADASYN \cite{tang2015kerneladasyn} extends ADASYN by applying kernel density estimation to model the minority class distribution and weighting samples by difficulty, producing more informative synthetic data. These findings highlight the importance of nonlinear transformations in both representation learning and data augmentation, motivating the shift toward deep generative models for more powerful imbalance handling.}

\textbf{Deep generative models.}
Conventional data generation methods typically produce synthetic samples positioned along the line segment connecting {minority samples}. Consequently, these techniques rely on local information and may not capture the overall distribution of the minority class effectively. Deep generative models, celebrated for their capacity to model the probability distribution of high-dimensional data, have gained widespread recognition for their ability to generate high-quality synthetic samples for minority classes.

Wan et al.~\cite{8285168} is a pioneering in the utilization of the generative models. Their work is dedicated to the precise modeling of high-dimensional data distributions via Variational Autoencoder (VAE), with a specific focus on generating synthetic samples for minority classes. VAE maps observed data points into a latent space while simultaneously learning the underlying data structure and probabilistic characteristics. The latent variable is represented by a probability distribution over possible values rather than a single fixed value, which is able to enforce the uncertainty of the model and lead to stable prediction. By employing VAE to generate synthetic samples through the utilization of random noise conforming to a standard normal distribution, this methodology not only ensures exceptional sample diversity but also consistently outperforms conventional data generation methods.

\begin{algorithm}[t]
\caption{{GAN-based Minority Generation}}
\label{alg:gan}
\resizebox{\linewidth}{!}{
\begin{minipage}{\linewidth}
\begin{algorithmic}[1]
\INPUT Real dataset $\mathcal{D} = \mathcal{D}_M \cup \mathcal{D}_m$, noise distribution $P_z$
\OUTPUT Generated synthetic minority dataset $\mathcal{D}_{\text{syn}}$
\STATE Initialize generator $G$ and discriminator $D$
\WHILE{not converged}
    \STATE Sample real data $x \sim \mathcal{D}$ 
    \STATE Sample noise $z \sim P_z$
    \STATE Generate a synthetic minority sample: $\hat{x} \gets G(z)$
    \STATE Train discriminator $D$ to distinguish real samples $x$ from generated $\hat{x}$
    \STATE Train generator $G$ to improve $\hat{x}$ so that $D(\hat{x})$ is predicted as real
\ENDWHILE
\STATE Generate synthetic minority samples: $\mathcal{D}_{\text{syn}} = \{\hat{x}_i = G(z_i) \mid z_i \sim P_z\}$
\RETURN $\mathcal{D}_{\text{syn}}$
\end{algorithmic}
\end{minipage}}
\end{algorithm}

In addition to VAE, another noteworthy contribution to data generation methods is found in the realm of Generative Adversarial Networks (GAN) \cite{goodfellow2020generative}. Douzas et al.~\cite{douzas2018effective} introduced an innovative approach that harnesses the power of conditional Generative Adversarial Networks (cGAN)~\cite{mirza2014conditional} to approximate the genuine data distribution and produce synthetic data specifically tailored for the minority class. {The pseudo-code for GAN is shown in Algorithm \ref{alg:gan}.} A typical cGAN comprises two primary components: a generator and a discriminator. The generator's role is to generate synthetic data based on provided labels, while the discriminator is tasked with distinguishing real data from artificially generated data. The application of GANs to data rebalancing involves an iterative training process. Initially, both the generator and discriminator undergo training to ensure that the discriminator becomes proficient at differentiating between various data samples, and the generator becomes capable of producing synthetic data corresponding to specific class labels. Subsequently, the trained generator is employed to conduct data augmentation for the minority class. This adversarial training procedure enables the network to capture a more accurate data distribution, resulting in the production of more reliable and representative synthetic samples. 

However, training a GAN directly for minority data generation poses significant challenges, primarily due to the scarcity of minority-class samples, which may be insufficient to effectively train a GAN model. In response to this challenge, BAGAN \cite{mariani2018bagan} is proposed to address this issue by proposing an alternative training strategy that incorporates data from both minority and majority classes simultaneously during the adversarial training process. The key innovation lies in class conditioning, which is applied within the latent space. Moreover, BAGAN leverages the discriminator not only for its traditional role of distinguishing fake samples from real ones but also as a classifier. This dual role enables the discriminator to not only discern the authenticity of generated samples but also identify the category of real samples. Consequently, BAGAN learns the underlying features of the specific classification problem by initially considering all available samples across both minority and majority classes. These shared features can then be applied universally across all classes, thereby facilitating the generation of high-quality minorities. Similarly, SCDL-GAN~\cite{cai2019supervised} and RVGAN-TL~\cite{ding2023rvgan} employ the Wasserstein auto-encoder~\cite{tolstikhin2017wasserstein} and VAE, respectively, to capture the distributive characteristics of all classes with guidance from label information before generating samples. EWGAN~\cite{ren2019ewgan} constructs entropy-weighted label vectors for each class to characterize data imbalance and employs the Wasserstein Generative Adversarial Network~\cite{arjovsky2017wasserstein} for stable training. EID-GANs~\cite{li2022eid} tackle extreme data imbalance by designing a novel penalty function that guides the generator to learn features of extreme minorities. Meanwhile, SMOTified-GAN \cite{sharma2022smotified} combines SMOTE with GAN, deploying GAN on pre-sampled minority data produced by SMOTE to enhance sample quality in minority classes. In contrast to the aforementioned GAN framework, GAMO ~\cite{mullick2019generative} offers an alternative solution by integrating a convex generator, classifier, and discriminator into a three-player adversarial game. GAMO's convex generator creates new samples for minority classes through convex combinations of existing instances. Its primary objective is to perplex both the discriminator and the classifier, generating samples that challenge accurate classification. These synthetic samples are strategically positioned near class boundaries, thus aiding in adjusting the classifier's decision boundaries to reduce misclassification of minority class data.

Alongside GANs, diffusion models~\cite{ho2020denoising,wang2022semantic} have gained attention for their robust representation capabilities. These models present a compelling alternative to the intricate adversarial training procedures. They operate on a fundamentally distinct principle—sequentially denoising data—a methodology that has proven to be remarkably stable and effective in generating synthetic samples. Within the context of addressing data imbalance, several pioneering works have harnessed diffusion models to excellent effect.
DiffMix~\cite{oh2023diffmix} extends the utility of diffusion models to nuclei segmentation and classification tasks, which demand precise semantic image and label map pairs, alongside realistic pathology image generation. It diversifies synthetic samples by creating virtual patches and strategically shifting instance mask locations to balance class variances. Additionally, SDM~\cite{wang2022semantic} is employed within DiffMix to synthesize diverse, semantically realistic, and balanced pathology nuclei data, conditioned on generated label maps.
DiffuASR~\cite{liu2023diffusion} addresses data sparsity and the long-tail user problem in sequential recommendation by leveraging diffusion-based pseudo sequence generation. This approach introduces classifier-guided and classifier-free strategies to maintain the original intent of users throughout the diffusion process.
MosaicFusion~\cite{xie2023mosaicfusion} introduces a diffusion-based data augmentation pipeline that simultaneously generates images and masks, particularly beneficial for large vocabulary instance segmentation. This method effectively mitigates distribution imbalances common in imbalanced and open-set scenarios. By running the diffusion process simultaneously on different image canvas regions and conditioning it on various text prompts, MosaicFusion controls the diffusion model to generate multiple objects at specific locations within a single image.
{
PoGDiff~\cite{wang2025pogdiff} introduces a novel fine-tuning method for diffusion models on imbalanced datasets by replacing the ground-truth distribution with a Product of Gaussians that incorporates neighboring text embeddings, significantly enhancing generation accuracy and quality.
DiffROP~\cite{yan2024training} proposes a probabilistic contrastive learning approach to mitigate class bias in diffusion models by reducing distributional overlap among synthetic images, significantly improving generation fidelity for {minority classes}.
{LDMLR \cite{pengxiao2024latent} utilities a backbone model encodes the original imbalanced data into embeddings, and a Denoising Diffusion Implicit Model (DDIM) is trained to generate pseudo-features that enrich the minority representation. These generated features are combined with real embeddings to train the classifier, effectively reducing bias while keeping computation efficient. By leveraging denoising-based iterative refinement in the latent space, LDMLR produces diverse yet realistic features that improve class balance without sacrificing representation quality.}
Recently, BPA~\cite{xu2024rethinking} has been proposed to address the {majority} class accumulation effect in diffusion models trained on imbalanced data; by adaptively assigning class-specific noise priors during training, BPA mitigates sampling bias introduced by uniform noise distributions, leading to improved generation diversity and visual fidelity across both {majority} and {minority classes}.
}

\noindent{\textbf{Discussion.}}
{To better inform practical deployment, we analyze the scalability and effectiveness trade-offs of data generation techniques. 
Linear generative methods, such as SMOTE and its variants (e.g., Borderline-SMOTE, ADASYN), create synthetic samples by interpolating between minority samples and their $k$ nearest neighbors. 
These methods typically operate with a time complexity of $\mathcal{O}(n k d)$, where $n$ is the number of {minority samples}, and $d$ is the feature dimensionality. Owing to their simplicity and low computational overhead, they scale well to relatively simple data structures, such as tabular datasets.
However, the inherent linearity of these methods restricts their capacity to model complex data distributions.}

{In contrast, deep generative methods, including GANs, VAEs, and Diffusion Models, learn to synthesize samples by modeling the underlying data distribution. These approaches incur significantly higher computational costs. GANs and VAEs generally require $\mathcal{O}(T n d)$ operations, where $T$ is the number of training epochs. Diffusion models are even more expensive, with time complexity of $\mathcal{O}(T S nd)$, where $S$ denotes the number of denoising steps per sample. While deep generative models offer improved diversity and fidelity, especially in high dimensional or unstructured data domains (e.g., images or molecular graphs), their scalability is often constrained by training time and hardware demands. 
To empirically evaluate this trade-off, we conduct experiments comparing both effectiveness and efficiency of Linear and Deep Generative Methods. Please refer to Section 5 for details.}

\subsubsection{Adaptive under-sampling}\
\label{s312}

In contrast to data generation techniques that augment the original dataset, under-sampling methods involve the reduction of data within the original dataset. Random under-sampling, as the most fundamental under-sampling approach, entails the random reduction of the majority samples to achieve dataset balance. However, this method may result in the discarding of potentially valuable data, which could bear significance for the inductive process. To enhance the efficacy of data under-sampling, researchers have explored strategies aimed at retaining the most representative {majority samples}. These strategies can be broadly classified into neighbor-based, clustering-based, and evolutionary-based methods.

\textbf{Neighbor-based methods.} Neighbor-based under-sampling methods revolve around the selection of samples based on their proximity to other samples. One notable technique is Tomek Link~\cite{kubat1997addressing}, which identifies pairs of examples as Tomek links if they belong to different classes and are each other's nearest neighbors. Under-sampling using Tomek Link can be executed by either removing all Tomek links from the dataset or specifically targeting the {majority samples} involved in these links. The removal of majority class nearest neighbors serves to reposition the learned decision boundary towards the majority class, thereby fostering the generation of {minority samples}. {The pseudo-code is shown in Algorithm \ref{alg:tomek}.}
One-sided selection (OSS) \cite{kubat1997addressing} extends the concept of Tomek Link to categorize {majority samples} into three groups: noise samples, boundary samples, and safety samples. Subsequently, it eliminates boundary and noise samples from the majority class.
Condensed Nearest Neighbor (CNN)~\cite{ref_6}, on the other hand, leverages the nearest neighbor rule to iteratively decide whether a sample should be retained or removed. Central samples that can be accurately classified by the nearest neighbor rule are discarded. However, this method can be relatively slow due to its need for multiple passes over the training data, and it is sensitive to noise, potentially incorporating noisy samples into the retained subset.
Edited Nearest Neighbor (ENN)\cite{ref_17} operates by removing samples whose class labels differ from the majority of their nearest neighbors. However, its efficacy is limited due to the prevalence of {majority samples} in close proximity to each other. Building upon ENN, the Neighborhood Cleaning Rule (NCR)~\cite{ref_18} combines ENN with k-NN~\cite{guo2003knn} to enhance the removal of noisy samples from the datasets.
The NearMiss family of methods~\cite{ref_19} introduces under-sampling techniques based on the distances between points in the majority class. NearMiss-1 retains {majority samples} whose mean distance to the k nearest minority class points is the lowest. In contrast, NearMiss-2 retains {majority samples} whose mean distance to the k farthest minority class points is the lowest. Finally, NearMiss-3 selects k nearest neighbors in the majority class for each point in the minority class. This approach directly controls the under-sampling ratio through the parameter k, eliminating the need for separate tuning.
NB-Rec~\cite{vuttipittayamongkol2020neighbourhood} investigates the presence of overlapping instances from various classes and introduces the utilization of the k-NN rule to investigate the local neighborhoods of individual samples. It selectively eliminates {majority samples} from the overlapping region without impacting any minority samples, leading to enhanced detection of overlapped minority instances.

\begin{algorithm}[t]
\caption{{Tomek Links Under-sampling}}
\label{alg:tomek}
\resizebox{\linewidth}{!}{
\begin{minipage}{\linewidth}
\begin{algorithmic}[1]
\INPUT Dataset $\mathcal{D} = \{(x_i, y_i)\}$
\OUTPUT Cleaned dataset $\mathcal{D}'$
\FORALL{pairs $(x_i, x_j)$ with $y_i \neq y_j$}
    \IF{$x_j$ is the nearest neighbor of $x_i$ and vice versa}
        \STATE Mark $(x_i, x_j)$ as Tomek Link
    \ENDIF
\ENDFOR
\STATE Remove {majority samples} in Tomek Links
\RETURN $\mathcal{D}'$
\end{algorithmic}
\end{minipage}}
\end{algorithm}

\textbf{Clustering-based methods.} Clustering-based under-sampling methods employ a clustering approach to select samples for under-sampling. The underlying concept is that datasets often comprise various clusters, each exhibiting distinct characteristics. This clustering strategy aims to capture both global information and sub-concepts within the dataset, thereby enhancing sampling quality.
SBC~\cite{yen2009cluster} initiates the process by clustering the dataset into several clusters, each containing both majority and {minority samples}. When a cluster exhibits a higher concentration of {minority samples} and a reduced presence of {majority samples}, it signifies that this cluster aligns more with the characteristics of the minority class. Consequently, SBC identifies an appropriate number of {majority samples} to retain from each cluster, taking into account the ratio of {majority samples} to {minority samples} within the cluster.
In contrast to clustering multiclass samples, CBMP~\cite{zhang2010cluster} takes a direct approach by clustering the majority of samples and subsequently selecting samples that are either nearest to the centroid or chosen randomly. CentersNN~\cite{lin2017clustering} clusters the majority class with the number of clusters equal to the number of data points in the minority class, using cluster centers to represent the majority class. 
DSUS~\cite{ng2014diversified} introduces diversified sensitivity-based under-sampling, which selects valuable samples with high stochastic sensitivity concerning the current trained classifier. To ensure that samples are not exclusively chosen near the decision boundary and do not distort the data distribution, DSUS clusters the majority class and selects only one representative sample from each cluster for inclusion in the selection process. 

However, clustering the majority class used in these methods is a computationally intensive process, which poses a significant challenge. To address this issue, Fast-CBUS~\cite{ofek2017fast} presents an efficient clustering-based under-sampling method that narrows its focus to clustering only the {minority samples}. Within each of these minority class clusters, Fast-CBUS selects {majority samples} that closely surround the minority instances and proceeds to create a dedicated model for each cluster. When classifying a new sample, the method prioritizes predictions made by models associated with clusters that exhibit the closest proximity to the sample. This approach is grounded in the idea that it's beneficial to train models using instances that are challenging to classify, particularly those {majority samples} that are situated near minority instances.

\textbf{Evolutionary-based methods.}
EUS~\cite{garcia2009evolutionary} presents a collection of techniques known as evolutionary under-sampling. These methods incorporate evolutionary algorithms into the process of under-sampling, with the primary aim of enhancing classifier accuracy by reducing instances predominantly from the majority class. To achieve this, EUS employs meticulously designed fitness functions, carefully striking a balance between data reduction, class balance, and classification accuracy.
Following EUS, MapReduce~\cite{triguero2015evolutionary} enhances it to make it feasible for large-scale datasets, even in the presence of extreme data imbalance. The approach involves partitioning the entire training dataset into manageable chunks. For each chunk containing minority data, the goal is to create a balanced subset using EUS, thus utilizing a representative sample from the majority set.
EUSC~\cite{le2021eusc} introduces a two-stage clustering-based surrogate model within the framework of evolutionary under-sampling, facilitating the expedited computation of fitness values. A key innovation is the creation of a surrogate model for binary optimization which is based on the meaning rather than their binary representation.
EUSBoost~\cite{galar2013eusboost, krawczyk2016evolutionary} further enhances the performance of base classifiers by integrating the evolutionary under-sampling approach with Boosting algorithms. This evolutionary strategy prioritizes the selection of the most consequential samples for the classifier learning phase, based on accuracy and an additional diversity term incorporated into the fitness function. Consequently, it effectively addresses the challenges posed by imbalanced scenarios.

In addition to the three categories mentioned above, several alternative approaches have been developed from other perspectives. 
TU~\cite{peng2019trainable} introduces a meta-learning technique to parameterize the data sampler and trains it to optimize classification performance based on the evaluation metric. By incorporating evaluation metric optimization into the data sampling process, this method effectively determines which instances should be retained or discarded for a given classifier and evaluation metric.
SDUS~\cite{9741366} utilizes supervised constructive learning, focusing on sphere neighborhoods (SPN) within the majority class. It offers two strategies: a top-down approach, treating SPN regions equally and favoring regions with smaller sample sizes, and a bottom-up approach, selecting samples proportionally based on SPN inner distribution for global pattern preservation.
RIUS~\cite{hoyos2021relevant} leverages the information preservation principle and calculates cross-information information potential. It then selects the most relevant examples from the majority class, allowing for the extraction of the underlying structure of the majority class with a reduced number of samples.
ECUBoost~\cite{8968753} aims to prevent the loss of informative samples during data pre-processing for boosting-based ensembles. This method employs both confidence and entropy as benchmarks to ensure the validity and structural distribution of the {majority samples} during under-sampling.

\subsubsection{Hybrid sampling}\
\label{s313}

Data generation and under-sampling share the common goal of achieving a balanced class distribution, but each method comes with its own set of limitations. Under-sampling can lead to information loss due to the removal of examples, and data generation techniques can increase the dataset size with potentially synthetic and less informative samples. To achieve a balanced class distribution and optimize classifier performance, researchers have introduced hybrid sampling methods. These approaches often leverage linear generative techniques to over-sample the minority class. However, linear generative methods may fall short in capturing the complex structure of minority samples, particularly in extremely imbalanced scenarios characterized by significant class overlap. To address this challenge, some methods aim to enhance the effectiveness of linear generative techniques by integrating them with adaptive under-sampling strategies. 
For instance, SMOTE+Tomek Links and SMOTE+ENN \cite{batista2004study} initially generate minority samples using SMOTE and subsequently identify and eliminate instances involved in Tomek links or those that have been misclassified.
CUSS~\cite{feng2020cluster} combines SMOTE with the clustering-based under-sampling method to minimize the risk of discarding valuable {majority samples}.
SMOTE-RSB~\cite{ramentol2012smote} and SMOTE-IPF~\cite{saez2015smote} incorporate rough set theory and iterative ensemble-based noise filter known as the iterative-partitioning filter, respectively to prevent the generation of noisy samples that might disrupt the overall data distribution.
BMW-SMOTE~\cite{gao2020ensemble} controls the number of synthetic instances generated based on a weight calculated from the ratio of neighboring majority samples, which effectively reinforces the recognition of the minority class within local regions.

In addition to the approach that combines data generation and sample selection to achieve a balanced data distribution, advanced methods delve into the intricacies of sampling strategies to create diverse training data distributions. For example, BBN~\cite{zhou2020bbn}, SimCAL~\cite{wang2020devil} and TLML~\cite{guo2021long} explore both the original imbalanced distribution and balanced distribution independently. 
Research endeavors~\cite{li2020overcoming, xiang2020learning, cai2021ace, cui2022reslt} categorize classes into groups, each characterized by a balanced distribution. SADE~\cite{zhang2022self} takes an innovative step by introducing the inversely imbalanced distribution to enhance the representation of minority classes in the dataset.
Potential Anchoring~\cite{koziarski2021potential} employs radial basis functions to measure class distribution density and maintains the original shape of class distributions by merging over- and under-sampling techniques. 
DLSA~\cite{xu2022constructing} introduces a streamlined approach by progressively adapting the label space, dynamically transitioning from imbalance to balance to facilitate classification. 
EHSO~\cite{zhu2020ehso} incorporates an evolutionary algorithm to identify the optimal balance between classification performance and the replication ratio of random oversampling. 
Dynamic Sampling~\cite{pouyanfar2018dynamic} leverages sample upgrading, making real-time adjustments to the dataset based on the training results. It randomly removes samples from well performing classes and duplicates samples from poorly performing classes, ensuring that classification model learns relevant information at each iteration. 

\noexpand{{\textbf{Discussion.}}} 
{While effective, the overlapping problem remains a significant challenge in sampling methods. These methods address class overlap in imbalanced data primarily through two strategies: (1) focusing on safe regions and (2) filtering noisy or borderline samples. The first strategy generates synthetic minority samples only within areas where minority instances are densely clustered and far from decision boundaries. This reduces the risk of creating synthetic points in overlapping regions, which could mislead the classifier. The second strategy involves applying noise-filtering techniques such as Edited Nearest Neighbors or Tomek Links, which identify and remove {majority samples} that lie close to or within minority regions. This helps clean the decision boundary and reduces ambiguity caused by overlapping components.} {As for limitations, one key issue is boundary distortion. If down-sampling removes too many majority samples near the decision boundary, it can oversimplify the class separation, leading to underfitting. Additionally, there is the risk of losing valuable majority data during the cleaning process. In scenarios where class overlap is genuine, aggressively removing overlapping majority instances may discard useful information necessary for accurate classification. These trade-offs highlight the need to balance data cleaning with decision boundary preservation in hybrid strategies.}

\subsubsection{Re-labeling}\
\label{s314}

Apart from data generation and under-sampling, researchers have recognized the untapped potential of unlabeled data as a valuable resource. By incorporating self-training and active learning strategies, these unlabeled samples can be effectively labeled to compensate for the lack of imbalanced category supervision information and improve classification ability. 

\textbf{Self-training.}
Self-training is a widely used technique where a model is initially trained on a labeled dataset and then iteratively improves itself by using its predictions (i.e., pseudo-labels) on unlabeled data to generate more training samples. In line with this approach, Yang et al.~\cite{yang2020rethinking} initially experimented with using pseudo-labels from minority classes to address class imbalance and subsequently retrained the model.
Expanding upon this concept, CReST~\cite{wei2021crest} made an important observation that models trained on imbalanced datasets often achieve low recall but high precision on minority classes. This suggests that these models are proficient at generating high-precision pseudo-labels for minority classes. To leverage this insight, CReST introduced a stochastic update strategy for pseudo-label selection, assigning higher selection probabilities to samples predicted as minority classes, which are more likely to be accurate.
To enhance the quality of pseudo-labels, SaR~\cite{Lai_2022_CVPR} introduced a calibration step that adjusts soft labels based on prediction bias before generating one-hot pseudo-labels. This calibration process aims to simultaneously enhance the recall of minority classes and the precision of majority classes.
In addition to utilizing linear classifiers for generating pseudo-labels, DASO~\cite{oh2022daso} explored the use of semantic pseudo-labels. These labels are derived by measuring the similarity between a given class center and unlabeled samples in feature space. DASO's research demonstrated that semantic pseudo-labels excel in predicting minority classes with high recall. As a result, the combination of linear and semantic pseudo-labels offers a promising approach to mitigating overall dataset bias.

\textbf{Active learning.}
Active learning is a technique that seeks to minimize the human labeling effort by automating the selection of the most informative unlabeled samples for human annotation. In the context of imbalanced data, active learning becomes a vital tool for addressing class imbalance and mitigating the potential error labels introduced by self-training. Various active learning strategies have emerged, each focusing on different aspects of informativeness, such as margin, uncertainty, and diversity.

Margin-based selection methods, as exemplified by SVMAL~\cite{ertekin2007active} and AI-WSELM~\cite{qin2021active}, choose the unlabeled instance closest to the classifier's decision boundary as the most informative sample. This approach is rooted in the concept that uncertain or ambiguous samples hold essential information critical for shaping the decision boundaries of a classifier. Hence, they should receive the highest priority for inclusion in the training set.

Uncertainty-based strategies for active learning rely on quantifying sample uncertainty using prediction entropy or confidence metrics. For instance, BootOS~\cite{zhu2007active} employs the entropy of prediction as an indicator of uncertainty for sample selection. It further introduces a novel concept of using max-confidence as an upper bound and min-error as a lower bound for halting the active learning process. Building upon this foundation, several other works have extended the uncertainty-based sampling approach, as evidenced in~\cite{aggarwal2021minority, liu2021comprehensive, dong2020cost, jin2022deep}.

In contrast to these uncertainty-based approaches, CEAL~\cite{fu2011certainty} acknowledges a unique challenge when applying active learning in imbalanced datasets. In such scenarios, the minority class often becomes overwhelmed by the majority class during the labeling process. To mitigate this imbalance, CEAL introduces the concept of "certainty-enhanced neighborhoods." For each unlabeled sample, it recommends considering only the local context where the majority class density is low. This strategic selection process reduces the likelihood of querying majority samples, offering a solution to the issue of minority class overshadowing during active learning.

However, these methods relying solely on a single model's output may not always perform optimally. This is particularly evident in cases where the initial model was trained on limited data and may produce biased sample selections throughout the learning process. To address these limitations, other sampling strategies focus on the diversity~\cite{8595005, dong2020cost, aggarwal2021minority} of selected samples. For example, SAL~\cite{8595005} employs active learning by training a similarity model that recommends unlabeled {minority} class samples for manual labeling.

%% file: tree.tex
\tikzstyle{leaf}=[draw=black,
    rounded corners,minimum height=1em,
    fill=black!15,text opacity=1, align=center,
    fill opacity=.5,  text=black,align=left,font=\footnotesize,
    inner xsep=3pt,
    inner ysep=1pt,
    ]
\tikzstyle{middle}=[draw=black,
    rounded corners,minimum height=1em,
    fill=white!40,text opacity=1, align=center,
    fill opacity=.5,  text=black,align=center,font=\footnotesize,
    inner xsep=3pt,
    inner ysep=1pt,
    ]

\begin{figure*}[!ht]
\centering
\scalebox{1}{
\begin{forest}
  for tree={
    forked edges,
    grow=east,
    reversed=true,
    anchor=base west,
    parent anchor=east,
    child anchor=west,
    base=middle,
    font=\footnotesize,
    rectangle,
    line width=0.5pt,
    draw=black,
    rounded corners,align=left,
    minimum width=2em,
    s sep=5pt,
    inner xsep=3pt,
    inner ysep=1pt,
  },
  where level=1{text width=4.5em}{},
  where level=2{text width=6em,font=\footnotesize}{},
  where level=3{font=\footnotesize}{},
  where level=4{font=\footnotesize}{},
  where level=5{font=\footnotesize}{},
[Methods for imbalance \\learning (Sec. \ref{s3}),  edge=black, text width=9em
    [Data re-balancing \\ (Sec. \ref{s31}),  edge=black, text width=9em
        [Data generation \\ (Sec. \ref{s311}),  text width=9em, edge=black
            [Linear generative method, leaf, text width=16em, edge=black]
            [Deep generative model, leaf, text width=16em, edge=black]
        ]
        [Adaptive down-sampling \\ (Sec. \ref{s312}),  text width=9em, edge=black
            [Neighbor-based method, leaf, text width=16em, edge=black]
            [Clustering-based method, leaf, text width=16em, edge=black]
            [Evolutionary-based method, leaf, text width=16em, edge=black]
        ]
        [Hybrid sampling \\ (Sec. \ref{s313}),  text width=9em, edge=black]
        [Re-labeling \\ (Sec. \ref{s314}),  text width=9em, edge=black
            [Self-training, leaf, text width=16em, edge=black]
            [Active learning, leaf, text width=16em, edge=black]
        ]
    ]
    [Feature representation \\(Sec. \ref{s32}),  edge=black, text width=9em
        [Cost-sensitive learning \\(Sec. \ref{s321}),  text width=12em, edge=black
            [Class-based method, leaf, text width=13em, edge=black]
            [Sample-based method, leaf, text width=13em, edge=black]
            [Weight determination, leaf, text width=13em, edge=black]
        ]
        [Metric learning \\(Sec. \ref{s322}),  text width=12em, edge=black
            [Sample selection, leaf, text width=13em, edge=black]
            [Distance calculation, leaf, text width=13em, edge=black]
        ]
        [Supervised contrastive learning \\(Sec. \ref{s323}),  text width=12em, edge=black
            [Re-balanced sampling, leaf, text width=13em, edge=black]
            [Re-weighting, leaf, text width=13em, edge=black]
            [Pair reconstruction, leaf, text width=13em, edge=black]
        ]
        [Prototype learning \\(Sec. \ref{s324}),  text width=12em, edge=black
            [Discriminative embedding generation, leaf, text width=13em, edge=black]
            [Class information compensation, leaf, text width=13em, edge=black]
            [Critical information extraction, leaf, text width=13em, edge=black]
        ]
        [Transfer learning \\(Sec. \ref{s325}),  text width=12em, edge=black
            [Sample synthesis, leaf, text width=13em, edge=black]
            [Others, leaf, text width=13em, edge=black]
        ]
        [Meta learning \\(Sec. \ref{s326}),  text width=12em, edge=black
            [Weights, leaf, text width=13em, edge=black]
            [Sampling method, leaf, text width=13em, edge=black]
            [Others, leaf, text width=13em, edge=black]
        ]
    ]
    [Training strategy \\ (Sec. \ref{s33}),  edge=black, text width=9em
        [Decoupling training (Sec. \ref{s331}), text width=16em, edge=black]
        [Fine-tuning pre-trained model (Sec. \ref{s332}),  text width=16em, edge=black]
        [Curriculum learning (Sec. \ref{s333}),  text width=16em, edge=black]
        [Posterior recalibration (Sec. \ref{s334}),  text width=16em, edge=black]
    ]
    [Ensemble \\ (Sec. \ref{s34}),  edge=black, text width=9em
        [Bagging-based methods (Sec. \ref{s341}),  text width=16em, edge=black]
        [Boosting-based methods (Sec. \ref{s342}),  text width=16em, edge=black]
        [Cost sensitive-based methods (Sec. \ref{s343}),  text width=16em, edge=black]
        [Knowledge distillation (Sec. \ref{s344}),  text width=16em, edge=black]
    ]
]
\end{forest}}
\caption{\footnotesize{{A taxonomy of imbalanced learning methods.
(1) Data re-balancing methods focus on modifying the input data distribution to mitigate class imbalance during the data preparation phase.
(2) Feature representation methods aim to learn discriminative embeddings that better capture minority class characteristics, improving class separability in the latent space.
(3) Training strategies adjust the learning process to reduce bias toward the majority class during model training.
(4) Ensemble methods integrate multiple models to enhance robustness and generalization, leveraging the diversity of learners to mitigate data imbalance effects during inference.}}}
\label{fig:tax}
\end{figure*}

%% file: 3.2Featurerepresentation.tex
\subsection{Feature representation}
\label{s32}
\begin{figure}[t]
\centering
\includegraphics[width=\linewidth]{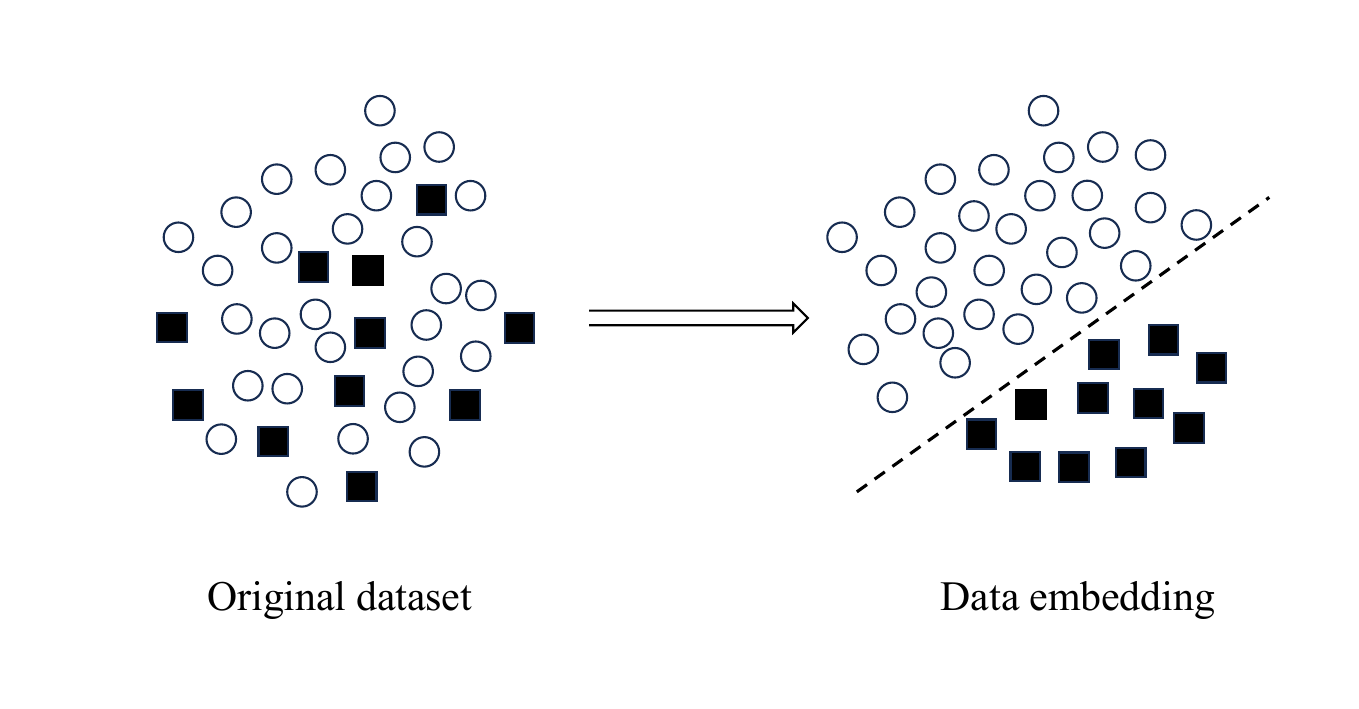}
\caption{{The feature representation methods aim to obtain high-quality and distinctive feature representations for data.}}
\label{fig_2}
\end{figure}

{In addition to re-balancing classifier training, obtaining a high-quality and distinctive feature representation for data is essential for mitigating the impact of class imbalance on model performance as depicted in Fig. \ref{fig_2}}. Several techniques can be employed to enhance feature representation in imbalanced learning, such as cost-sensitive learning, metric learning, supervised contrastive learning, prototype learning, transfer learning, and meta-learning.

\subsubsection{Cost-sensitive learning}\
\label{s321}

Cost-sensitive learning has emerged as a highly effective approach for tackling the imbalance challenges. 
The foundation of cost-sensitive learning is rooted in the concept of cost matrices~\cite{ref_A}. In contrast to conventional machine learning algorithms that treat all classification errors equally, cost matrices introduce the pivotal notion of assigning predefined costs to misclassified samples, i.e., false positives and false negatives. These assigned costs effectively function as weight factors, elevating certain misclassification errors above others in terms of importance. The overarching objective is to minimize the overall expected cost, thereby steering models to place heightened emphasis on minority classes, transcending the mere fixation on average classification errors. In essence, the concept of "costs" embodies the role of "weights" in the context of reweighting errors, {enabling models adjust boundaries and focus on important classes}
{\footnote{In this survey, we utilize the terms ``cost'' and ``weight'' concurrently, following the method used in the original paper.}}.

In the early stages, methods focus on class-specific costs and entail assigning distinct costs or weights to different classes, with these predefined costs typically established based on experiential knowledge or domain expertise~\cite{ref_A, ref_C}. The representative method is MetaCost~\cite{ref_B}, a general approach aimed at enabling non-cost-sensitive algorithms to incorporate cost-sensitive considerations. The key innovation lies in the re-labeling procedure, which begins by training multiple models on bootstrap replicates of the training dataset. Subsequently, the samples are re-labeled based on the minimized cost function. Finally, the classifier is trained using these refreshed labels. Another approach, threshold-moving \cite{zhou2005training}, involves adjusting the output threshold towards the less expensive classes. This adjustment makes it more challenging to misclassify examples with higher associated costs. This method employs the training set to train a neural network, and cost sensitivity is introduced during the test phase by directly modifying the output logits based on the associated costs. In addition to binary classification, Zhou et al.~\cite{zhou2010multi} explored the application of cost-sensitive learning in multi-class classification tasks. They observed that the consistency of cost matrices has a significant impact on classification performance. In cases where the cost matrix lacks consistency, rescaling is recommended, particularly after decomposing the multi-class problem into a series of two-class problems. CSMLP~\cite{castro2013novel} presents a novel approach involving the design of a joint objective function for Multilayer Perceptrons. This method utilizes a single cost parameter to differentiate the importance of class errors. The class-balanced loss~\cite{cui2019class} emphasizes that as the number of samples increases, the additional benefit of adding a new data point diminishes. To address this, it introduces the concept of the effective number to approximate the expected class size and leverages this effective number to calculate weights, rather than relying solely on the sample quantity.

% Focal Loss
\begin{algorithm}[t]
\caption{{Focal Loss Computation}}
\label{alg:focal}
\resizebox{\linewidth}{!}{
\begin{minipage}{\linewidth}
\begin{algorithmic}[1]
\INPUT Logits $z$, true label $y$, hyperparameters $\gamma$ and $\alpha$
\OUTPUT Loss $\mathcal{L}$
\STATE Apply softmax to obtain the predicted probability for the true class:
\STATE $p \gets \text{softmax}(z)_y$
\STATE Down-weight well-classified examples and focus on hard ones using the modulating factor $(1 - p)^\gamma$:
\STATE $\mathcal{L} \gets -\alpha \cdot (1 - p)^\gamma \cdot \log(p)$
\RETURN $\mathcal{L}$
\end{algorithmic}
\end{minipage}}
\end{algorithm}

Besides the class-specific costs, cost-sensitive learning extends its scope to encompass the weighting of individual samples. Focal loss~\cite{lin2017focal} offers a perspective on addressing imbalance problems at the instance level, introducing varying weights for each training sample {as shown in Algorithm \ref{alg:focal}}. Given that class imbalance often amplifies the difficulty of predicting minority classes, focal loss assigns higher costs to more challenging samples while applying lower costs to those that are easier to classify during training. 
Influence-balanced loss~\cite{park2021influence} takes a different approach by employing influence functions to gauge the impact of individual samples on intricate and biased decision boundaries. The loss associated with each sample is proportionally weighted based on the inverse of its influence.
Balanced softmax~\cite{ren2020balanced} extends the conventional softmax function to address imbalanced distributions. It accomplishes this by incorporating the imbalanced conditional prediction probability. This generalized softmax function serves as a form of logit calibration, effectively adjusting each class's logit value in proportion to its label frequencies, thus providing a more balanced representation in the feature space.

Another critical aspect of cost-sensitive learning revolves around the challenge of determining appropriate weights. Predefined weights often necessitate careful tuning of additional weight hyperparameters, introducing complexity to the process. To address this issue, researchers have explored diverse avenues for automatically learning the weights, considering various perspectives, including evolutionary search, gradient-based techniques, distribution transport, and meta-learning.
EDS~\cite{cao2013novel} employs an evolutionary search method to effectively seek optimal misclassification costs in multiclass imbalance scenarios based on an objective function defined with G-mean. This approach optimizes cost parameters through an evolutionary strategy.
ECS-DBN~\cite{zhang2018cost} utilizes adaptive differential evolution to optimize misclassification costs based on the training data. This enhances the effectiveness of deep belief networks by incorporating evaluation measures into the objective function.
% CoSen~\cite{khan2017cost} jointly optimizes the class-dependent costs and the neural network parameters to automatically learn robust feature representations for both the majority and minority classes. 
For gradients-based methods, GDW~\cite{chen2021generalized} addresses the imbalance problem through class-level gradient manipulation. It efficiently obtains class-level weights by developing a two-stage weight generation scheme that involves storing intermediate gradients. 
Equalization Loss~\cite{tan2020equalization} and Equalization Loss v2~\cite{tan2021equalization} determine weights from gradients as well. While Equalization Loss protects the learning of minority classes by blocking some negative gradients, it acknowledges that positive gradients may still be overwhelmed by the negative ones. Therefore, Equalization Loss v2 proposes balancing the positive-to-negative gradient ratio, using the accumulated gradient ratio as an indicator to up-weight positive gradients and down-weight negative gradients.
OT~\cite{guo2022learning} approaches the issue by viewing the training set as an imbalanced distribution over its samples. It transports this distribution to a balanced one using optimal transport. The weights of the training samples are determined as the probability mass of the imbalanced distribution, learned by minimizing the optimal transport distance between the two distributions. 
% AREA~\cite{chen2023area} argues that the number of samples in a class can not accurately measure the size of its spanned space and uses the estimated size of the spanned space for each category to adjust the training weight.
Besides, meta-learning methods are utilized to learn optimal weights through bi-level optimization as well. For instance, LRE~\cite{ren2018learning} presents a novel meta-learning algorithm that learns to assign weights to training examples based on their gradient directions. Meta-Weight-Net~\cite{shu2019meta} automatically learns an explicit loss-weight function, parameterized by a multilayer perceptron, from the data.

\subsubsection{Metric learning}\
\label{s322}
% \cite{Ren_2022_CVPR}
% \cite{Li_2022_CVPR}
% \cite{park2021influence}

Metric learning aims to learn a discriminative feature space by designing task-specific distance metrics between samples, which makes samples of the same class closer together, while samples of different classes are further apart \cite{duan2017deep}, {as shown in Algorithm \ref{alg:triplet}}. It achieves great success in feature representation due to the powerful capacity of understanding the similarity relationship among samples. Thus, metric learning methods has been wildly applied to mitigate the imbalanced classification problems to increase the separability between minority and majority classes and achieve a better classification performance on minorities.

\begin{algorithm}[t]
\caption{{Triplet Loss}}
\label{alg:triplet}
\resizebox{\linewidth}{!}{
\begin{minipage}{\linewidth}
\begin{algorithmic}[1]
\INPUT Anchor $x_a$, Positive $x_p$, Negative $x_n$, Margin $m$, encoder $f$
\OUTPUT Loss $\mathcal{L}$
\STATE Encode each sample into the latent space using $f$
\STATE Compute the distance between the anchor and the positive (same class):
\STATE $d_{ap} \gets \|f(x_a) - f(x_p)\|^2$
\STATE Compute the distance between the anchor and the negative (different class):
\STATE $d_{an} \gets \|f(x_a) - f(x_n)\|^2$
\STATE Enforce that the anchor is closer to the positive than to the negative by at least margin $m$:
\STATE $\mathcal{L} \gets \max(0, d_{ap} - d_{an} + m)$
\RETURN $\mathcal{L}$
\end{algorithmic}
\end{minipage}}
\end{algorithm}

\textbf{Sample selection.}
In metric learning methods, sample selection is the most important role because it directly influences the distances calculation and the quality of the extracted similarity relationship. Therefore, the sample selection of metric learning methods for imbalanced data always emphasizes the minority class \cite{huang2016learning, dong2017class, wang2018iterative, gui2022quadruplet, yan2023borderline} and carefully selects samples to design the distance metrics.  

To balance the training samples used in the training procedure, LMLE \cite{huang2016learning} clusters samples in each class and construct the sample sets by selecting four representative samples from neighboring clusters and classes for each anchor sample. Benefiting to clustering, LMLE captures the significant data variability within each class and greatly enriches the diversity of minority sample sets for distance evaluation. According to four distances defined on the selected sample sets, the triple-header hinge loss constrains the feature representation and enforces a local margin while handling class imbalance problems. 
However, the clustering is sensitive to the entire training data and class size and usually gets limited results for extremely sparse minority classes. 
To address this problem, class rectification loss \cite{dong2017class} construct the triplets by identifying hard-positive and hard-negative samples for minorities in each training batch and designs a batch-wise hard mining metric based on relative, absolute, and distribution information among selected triplets. {Constraining hard-positive samples facilitates the discovery of sparsely represented minority class boundaries, whereas mining hard-negative samples enlarges the margins against confounding impostor classes. Together, these strategies alleviate intra-class compactness and inter-class sparsity in the feature space.}
LSTM-QDM \cite{gui2022quadruplet} utilizes the minority sample to expend the data triplet in the triplet network \cite{cheng2016person} and constructs a quadruplet loss to balance the relationship among anchor, positive, negative and minority samples. Benefiting from the extra minority samples in quadruplet loss and optimizing the softmax loss jointly, LSTM-QDM achieves a stronger representation ability for imbalanced imbalanced time-series data.
Inspired by Borderline-SMOTE, DMFBML \cite{yan2023borderline} separates the samples into three categories: danger, safe and noise based on the ratio of neighbors from the different classes and generates the triplets by the guidance of three categories. Afterward, DMFBML designs a borderline margin loss function to customize the margins among different categories and assign more metric constraints for minority samples, thereby the minority and majority samples in overlapping regions could be separated.

\textbf{Distance calculation.}
Besides introducing minority samples in the sample selection strategy, other works handle the imbalance data and constrain the minority class from distance calculation.
Range loss \cite{zhang2017range} explicitly defines intra-class loss as the harmonic mean of the k-largest distances between samples within each class, and constructs inter-class loss by measuring the minimum distance among class centers. By optimizing these two components, range loss can simultaneously reduce the intra-class variations of different classes and enlarge the inter-class distance.
To treat different classes equally and separate them balanced, DMBK \cite{feng2018learning} first assumes that different classes are sampled from Gaussian densities with different expected values but one identical covariance, and then defines normalized divergence using KL-divergence to represent inter-class divergence. Afterword, inspired by arithmetic-geometric average inequality, DMBK makes use of the geometric mean to constrain the normalized divergences and make all inter-class divergences balanced. This procedure separates all classes in a balanced way and avoids neglected important minorities and inaccurate similarities incurred by imbalanced class distributions.
The Affinity loss function \cite{hayat2019gaussian} proposes to measure class similarities for an input feature in Euclidean space using a Gaussian similarity measure, specifically in terms of the Bergman divergence. Additionally, it utilizes a diversity regularizer to constrain all class centers to be uniformly distributed throughout the feature space. This approach helps the model avoid the majority classes from occupying a disproportionately larger angular space compared to the minority classes, leading to a more balanced embedding space.
Imbalanced metric learning \cite{gautheron2020metric} designs a new weighting method for Mahalanobis distance-base metric learning methods by decomposing the pair of samples based on their labels and introduces the hyper-parameters to control the negative effect of the imbalance. Moreover, imbalanced metric learning derives a generalization bound theoretically involving the proportion of positive samples using the uniform stability framework.

% \textbf{Margin adjustment}
% LDAM \cite{cao2019learning} extends the existing soft margin loss to imbalanced data scenarios by encouraging the minority classes to have larger margins so that the model can improve the generalization error of minority classes. It starts from the classification error bound and theoretically obtain the optimal trade-off of margins among multiple classes which is inversely proportional to the fourth root of the number of examples in each class. With the optimal margins, the LDAM can be extend to any sort of classification loss functions, such as Hinge loss and Cross entropy loss.
% NSL-AAM \cite{} analyses the softmax loss in imbalance setting and proposes an adaptive angle margin penalty which is added to the angle between sample embeddings and the weights of the last linear layer. The angle margin penalty of the class is proportional to the effective sample number of the class, and the larger effective numbers of the class is, the larger angle margin penalty the class suffers.

\subsubsection{Supervised contrastive learning}\
\label{s323}
% BaCon \cite{kang2022comprehensive} 
% PaCo \cite{cui2021parametric} state of the art

\begin{algorithm}[t]
\caption{{Supervised Contrastive Loss}}
\label{alg:supcon}
\resizebox{\linewidth}{!}{
\begin{minipage}{\linewidth}
\begin{algorithmic}[1]
\INPUT Normalized embeddings $\{z_i\}_{i=1}^N$, labels $\{y_i\}_{i=1}^N$, temperature $\tau$
\OUTPUT Loss $\mathcal{L}$
\STATE $\mathcal{L} \gets 0$
\FOR{$i = 1$ to $N$}
    \STATE Define the positive sample: $\mathcal{P}(i) \gets \{p \neq i \mid y_p = y_i\}$ 
    \STATE Define the negative sample: $\mathcal{N}(i) \gets \{n \neq i \mid y_n \neq y_i\}$ 
    \STATE $Z \gets \sum_{j \in \mathcal{P}(i) \cup \mathcal{N}(i)} \exp(\text{sim}(z_i, z_j)/\tau)$
    \FORALL{$p \in \mathcal{P}(i)$}
        \STATE $\mathcal{L} \mathrel{+}= -\frac{1}{|\mathcal{P}(i)|} \log \frac{\exp(\text{sim}(z_i, z_p)/\tau)}{Z}$
    \ENDFOR
\ENDFOR
\STATE $\mathcal{L} \gets \frac{1}{N} \mathcal{L}$
\RETURN $\mathcal{L}$
\end{algorithmic}
\end{minipage}}
\end{algorithm}

Supervised contrastive learning has emerged as a promising research direction in recent years, combining metric learning and self-supervised contrastive learning to leverage the advantages of both methods while overcoming their respective limitations \cite{khosla2020supervised}. 
On the one hand, conventional metric learning methods (e.g., the triplet loss \cite{weinberger2009distance} and N-pair loss \cite{sohn2016improved}) and self-supervised contrastive learning \cite{tian2020contrastive} limit the number of positive samples for each anchor to one, which places more emphasis on optimizing individual pairs and hinders the extraction of richer inter-sample information \cite{khosla2020supervised}. 
On the other hand, while contrastive learning methods have been found to generate a balanced feature space and maintain stable performance, even when the degree of imbalance becomes increasingly severe \cite{yang2020rethinking, kang2021exploring}, the features learned through contrastive learning lack semantic separability since they are learned without class label guidance, leading to poor classification performance. 
To address these limitations and handle the imbalance problem, supervised contrastive learning methods incorporate label information into the contrastive learning process and extend positive examples by including other samples from the same class along with data augmentations. This encourages the model to produce closely aligned representations of all samples from the same class and results in a more robust clustering of the representation space. 

SupCon \cite{khosla2020supervised} first extends contrastive learning to the fully-supervised setting and surpasses the performance of the traditional supervised cross-entropy loss by leveraging class labels with a contrastive loss. {The pseudo-code is shown in Algorithm \ref{alg:supcon}}. Although achieving performance improvement, researchers discover that the supervised contrastive loss still suffers from the imbalance of positive/negative pairs. To mitigate this issue and further enhance supervised contrastive learning for imbalanced data, the improved methods can be categorised into three classes, including {re-balanced sampling, re-weighting, and pair reconstruction}.

\textbf{Re-balanced sampling} methods aim to address the issue of imbalanced positive/negative pairs by sampling techniques. 
KCL \cite{kang2021exploring} improves upon the positive set construction rule in SupCon by randomly selecting k instances for each anchor from the same class to form the positive set, instead of using all the instances from the same class. This modification ensures that there is an equal number of positive samples for each anchor sample, thereby avoiding the dominance of instance-rich classes in the representation learning process.
CA-SupCon \cite{zhang2022class} argues that due to the scarcity of minority samples, SupCon cannot guarantee the separability of minority classes in one mini-batch. To address this issue, CA-SupCon introduces a class-aware sampler that re-balances the data distribution within each mini-batch during training, thus allowing each minority class to have a fair chance to contribute to the feature distance optimization.
GPCL \cite{jiang2021guided} introduces self-training to supervised contrastive learning to provide pseudo-labels for unlabeled data. By incorporating these pseudo-labels with confidence scores, GPCL can make full use of unlabeled data to enrich the contrastive data while filtering out unreliable negative and positive pairs with the same pseudo-labels and low confidence scores in the contrastive loss, respectively.

\textbf{Re-weighting} methods aim to increase the importance of minority pairs in the contrastive loss. For example, PaCo \cite{cui2021parametric} introduces a set of parametric class-wise learnable centers into the contrastive loss. This approach adaptively weighs the samples from different classes and enhances the intensity of pushing samples of the same class closer to their corresponding centers, effectively giving more weight to the minority class pairs.
BCL and TSC \cite{zhu2022balanced, li2022targeted} argue that supervised contrastive learning should ideally result in an embedding where different classes are uniformly distributed on a hypersphere. However, applying the contrastive loss to imbalanced data can lead to poor uniformity, with minority classes having poor separability in the feature space. 
BCL provides evidence for this claim through a theoretical analysis of the lower bound of the contrastive loss. The analysis shows that imbalanced data distribution heavily affects the model and the resulting representations of classes no longer attain a regular simplex configuration, which ultimately hinders the final classification performance.
Therefore, to address this issue, BCL proposes three class-averaging solutions to re-weight the negative pairs in the supervised contrastive loss. This helps to reduce the gradients from negative majority classes and contributes to an asymmetrical geometry configuration. Additionally, to ensure that all classes appear in every mini-batch, BCL introduces class-center representations in the positive and negative pairing, compensating for the batch sampling bias of high-frequency classes.Although motivated by the same insight, it is worth noting that TSC proposes a distinct solution from BCL. TSC aims to maintain a uniform distribution in the feature space for all classes by urging the features of classes closer to the uniformly distributed target features on the vertices of a regular simplex. To maintain semantic relations among classes, TSC proposes an online matching method to ensure that classes that are semantically close to each other are assigned to targets that are also close to each other.

\textbf{Pair reconstruction.} 
In addition to re-balanced sampling and re-weighting techniques, researchers also attempt to leverage the {pair reconstruction} in supervised contrastive learning to mitigate the data imbalance.
SNP-ECC \cite{gao2022ensemble} and MLCC-IBC \cite{gao2023imbalanced} extend the point-to-point contrastive learning approach to a point-to-group approach, and mitigate the imbalance problem by enriching the training samples. They construct contrastive sample groups by randomly sampling from the k-nearest majority/minority neighbors of the anchor sample, respectively. Afterwards, the new-format training data are generated by pairing the anchor sample with various contrastive sample groups. Due to the increased diversity of the contrastive sample groups, the new training data pairs are balanced and are learned using a novel label matching task.
ProCo \cite{yang2022proco} proposes to utilize the category prototype to augment representative samples and generate the adversarial proto-instance via linear interpolation in the feature space. The adversarial proto-instance is designed as a special outlier that can balance the data and encourage ProCo to rectify the decision boundaries of the minority categories during contrastive learning.

\subsubsection{Prototype learning}\
\label{s324}

In addition to exploring correlations among extensive and complex samples via metric learning and supervised contrastive learning, researchers have incorporated the idea of prototype learning \cite{snell2017prototypical} from few-shot learning into the imbalanced data scenario to enhance feature representations of minority classes.
A prototype explicitly represents a summary of all instances belonging to a specific class or intrinsic sample characteristics. In the context of imbalanced data, prototypes can improve minority representation from three aspects: {facilitating the generation of a more discriminative feature space}, {compensating for class information compensation}, and {aiding in the extraction of intrinsic sample information}.

\begin{algorithm}[t]
\caption{{Prototype-based Contrastive Loss}}
\label{alg:proto_contrast}
\resizebox{\linewidth}{!}{
\begin{minipage}{\linewidth}
\begin{algorithmic}[1]
\INPUT Embeddings $\{z_i\}_{i=1}^N$, labels $\{y_i\}_{i=1}^N$, temperature $\tau$
\OUTPUT Loss $\mathcal{L}$
\STATE Compute prototype (mean embedding) for each class $c$: 
\STATE \quad $p_c \gets \frac{1}{|\mathcal{I}_c|} \sum_{i \in \mathcal{I}_c} z_i$, where $\mathcal{I}_c = \{i \mid y_i = c\}$
\STATE $\mathcal{L} \gets 0$
\FOR{$i = 1$ to $N$}
    \STATE Let $p^+ \gets p_{y_i}$ be the prototype of sample $i$'s true class
    \STATE Let $\mathcal{P} = \{p^+\}$ and $\mathcal{N} = \{p_c \mid c \neq y_i\}$ be positive and negative prototypes
    \STATE Compute denominator: $Z \gets \sum_{p \in \mathcal{P} \cup \mathcal{N}} \exp(\text{sim}(z_i, p)/\tau)$
    \STATE $\mathcal{L} \mathrel{+}= -\log \frac{\exp(\text{sim}(z_i, p^+)/\tau)}{Z}$
\ENDFOR
\STATE $\mathcal{L} \gets \frac{1}{N} \mathcal{L}$
\RETURN $\mathcal{L}$
\end{algorithmic}
\end{minipage}}
\end{algorithm}

\textbf{Discriminative embedding generation.}
Class-specific prototypes provide a central target for each class in the feature space. This enables sample embeddings explicitly converge toward the class center, contributing to a more distinctive feature space for classification. 
Hybrid-PSC \cite{wang2021contrastive} improves the supervised contrastive learning method by treating the prototypes as the pairing object for positive/negative pairs, forcing each sample to be pulled towards the prototypes of its class and pushed away from prototypes of all the other classes. {The pseudo-code is shown in Algorithm \ref{alg:proto_contrast}.} This approach allows for more robust and efficient data sampling when dealing with large-scale datasets under a limited memory budget. 
The Affinity loss function \cite{hayat2019gaussian} provides a theoretical analysis of the Softmax loss with a linear prediction layer and treats the linear weight vectors as the class prototypes. Based on this, the affinity loss function can automatically learns the prototypes for each class and conveniently measures feature similarities with class prototypes in Euclidean space using Bergman divergence.
Instead of calculating prototypes from labeled data, TSC \cite{li2022targeted} manually generates a set of targets uniformly distributed on a hyper-sphere as the class prototypes and forces all classes to converge to these distinct distributed targets during training for a uniform distributed feature space.
BCL \cite{zhu2022balanced} uses supervised contrastive loss to train the model, and specifically ensures that each sample is paired with all other class prototypes in the negative pairing. By doing so, BCL compensates for the batch sampling bias of high-frequency classes, which in turn enables comprehensive contractive information from each class to be captured during training.
It is worth noting that the number of class prototypes is not limited to one, and studies \cite{wang2021contrastive, hayat2019gaussian} have shown that adopting a multi-prototype paradigm can capture richer within-class data distribution and promote diversity and discriminability in the feature space.
{For example, Prototype-based Augmentation (ProAug) \cite{hong2024proaug} builds a multi-center prototype dictionary optimized with a category-aware margin loss to capture the multi-modal distribution of each class. Based on this dictionary, ProAug dynamically composes context-similar prototypes via an attention mechanism to generate informative minority features, thereby enriching the representation space and mitigating biased decision boundaries.}

\textbf{Class information compensation.}
As a summary of classes, class prototypes contain basic class information that can help compensate for the information shortage of minority classes by transferring knowledge across classes or generating minority samples directly through data augmentation.
OLTR \cite{liu2019large} utilizes an attention mechanism to adaptively merge all prototypes with each sample embedding. This helps transfer basic class concepts among all classes to the feature embeddings, thereby enhancing the minority embeddings.
PAN \cite{zeng2021pan} addresses the modality-imbalance problem, where the available data for text, image, and audio are all considerably imbalanced, by generating synthetic samples for the minority modalities. It learns semantically consistent prototypes that capture the semantic consistency of the same class across different modalities and uses a gated recurrent unit to fuse the nearest neighbors of the majority modality to generate new samples. 
MPCL \cite{fu2022meta} argues that empirical prototypes are severely biased due to the scarce data in minority classes and proposes a meta-learning method to calibrate the prototypes by contrastive loss. With the assumption that feature embeddings for each class lie in a Gaussian distribution, MPCL estimates the mean and covariance of class distribution by the calibrated prototypes. Then, MPCL samples new samples from the distribution as synthetic minorities to balance the classes.
Similarly, ProCo \cite{yang2022proco} proposes to utilize the class prototype to augment representative samples and generate the adversarial proto-instance to balance the data. This encourages ProCo to rectify the decision boundaries of the minority classes.

\textbf{Critical information extraction.}
Prototypes not only represent entire classes but can also be leveraged to extract intrinsic information from individual samples, which is crucial in handling typical characteristics of different classes. 
For instance, IEM \cite{zhu2020inflated} proposes to use the attention mechanism to learn multiple discriminative feature prototypes for each class, considering both local and global features, as a single prototype may not be sufficient to represent a class. By gradually updating these prototypes during the training procedure, IEM captures the representative local and global semantics of each class and incorporates more discriminative features to improve generalization on minority classes.
MPNet \cite{he2020learning} generates both instance and semantic prototypes shared by all samples, representing local representative features and semantic features, respectively. The learned prototypes are recorded in a memory module and used to represent each sample, addressing the issue of forgetting minority classes during training. Additionally, semantic memory regularization is applied for a separately distributed centroid space.

\subsubsection{Transfer learning}\
\label{s325}

Transfer learning is a technique that leverages salient features and patterns identified in a source domain to enhance performance in a target domain. In the context of imbalanced datasets, minority classes often suffer from a scarcity of labeled data, while majority classes benefit from an abundance of such data. Consequently, researchers concentrate on identifying shared characteristics or attributes among multiple classes and employ transfer learning to transfer common knowledge from well-represented majority classes to underrepresented minority classes.

\textbf{Sample synthesis.}
A substantial portion of transfer learning methods focus on transferring the majority class diversity to minority classes by synthesizing new minority samples.
For instance, FTL \cite{yin2019feature} presents a prototype-based transfer framework, predicated on a Gaussian prior encompassing all classes. Under this assumption, each class is partitioned into mean and variance, representing class-specific and class-generic factors, respectively. The shared variance of major classes, indicating commonalities between categories such as varying poses or lighting conditions, is conveyed to minor classes through the generation of synthetic samples. {The pseudo-code is shown in Algorithm \ref{alg:transfer}.}
LEAP \cite{liu2020deep}models the distribution of intra-class features based on the distribution of angles between features and their corresponding class centers. By assuming that these angles adhere to a Gaussian distribution, the angular variance learned from the majority class is transferred to each minority class, and synthetic samples for minority classes are drawn from the distribution with enhanced intra-class angular diversity.
Similarly, FSA \cite{chu2020feature} employs class activation maps to identify class-specific features from underrepresented classes and fuses them with class-generic features to generate minority samples. RSG \cite{wang2021rsg} devises a {minority class} sample generator module, compatible with any backbone model, to create new samples based on the variation information of majority classes.
In contrast to the aforementioned methods that generate samples from minority instances, M2m \cite{kim2020m2m} directly transfers the diversity of majority classes by translating majority samples into minority samples. By referring to the output of the pre-trained classifier, a small optimized noise is added to the majority sample to facilitate the translation. In this manner, the richer information of majority samples is effectively leveraged to augment the minority classes.
{Sample-Adaptive Feature Augmentation (SAFA) \cite{hong2022safa} extracts transferable, class-irrelevant semantic directions from majority class features using an auto-encoder framework, then translates minority samples along these directions to generate diverse, sample-specific augmented features. A recycling training scheme enforces semantic consistency, contrastive loss ensures transferability, and mode-seeking loss promotes diversity. As a plug-in module, SAFA can be integrated with various networks and loss functions, improving decision boundary calibration without extra inference cost.}

In addition to sample synthesis, several methods endeavor to transfer knowledge in alternative ways.

MetaModelNet \cite{wang2017learning}, rather than transferring majority class knowledge directly, conveys the changing trend of model parameters and introduces a meta-learning method to learn the model dynamics, predicting how model parameters will alter as more training data is progressively incorporated. Specifically, MetaModelNet is trained to forecast many-shot model parameters from few-shot model parameters, which are trained on abundant and limited samples, respectively. By employing the learned model dynamics, MetaModelNet can directly enhance the model parameters trained on minority instances by predicting the parameters when additional minority samples are included.

OLTR \cite{liu2019large} explicitly constructs class centroids and exploits the attention mechanism to amalgamate sample features with class information from various classes, yielding more comprehensive feature representations. This information aggregation not only encompasses minority class information but also transfers basic class concepts among majority classes. Furthermore, the attention mechanism enables the model to adaptively tailor the importance of class information for different samples during end-to-end training, significantly enriching the minority features that lack adequate supervision signals.

GistNet \cite{liu2021gistnet} posits that all classes share a common geometric structure in the feature space. With numerous samples in the majority class, GistNet can encode the class geometry into a set of structure parameters shared by all classes, and the geometric structure of the minority class is restricted via learned structure parameters.

RVGAN-TL \cite{ding2023rvgan} initially generates minority samples using GAN and subsequently applies the TrAdaboost algorithm \cite{tradaboost} to assign different weights to the training samples. Synthetic minority samples that deviate from the real sample distribution are assigned extremely low weights, effectively transferring the distribution of real samples to the minority class and reducing the impact of noisy data.

% \cite{li2021self}

% Add this as the applocation or future work
% Note that some related works focus on addressing the domain class imbalance problem in transfer learning \cite{9655605, 8215613, 8260654}, which occurs when the class distributions between the source and target domains are significantly different. It is important to distinguish the domain class imbalance problem from the class imbalance issue studied in this survey.

\begin{algorithm}[t]
\caption{{Transfer Learning (FTL)}}
\label{alg:transfer}
\resizebox{\linewidth}{!}{
\begin{minipage}{\linewidth}
\begin{algorithmic}[1]
\INPUT Labeled majority class data $\mathcal{D}_M$ and minority class data $\mathcal{D}_m$
\OUTPUT Synthetic {minority samples} $\mathcal{D}_{syn}$
\STATE Estimate class-wise means $\mu_c$ and shared variance $\Sigma$ for $\mathcal{D}_M$
\STATE Estimate class-wise means $\mu_m$ for $\mathcal{D}_m$
\STATE Transfer $\Sigma$ from majority to minority classes
\STATE Generate synthetic samples: $x \sim \mathcal{N}(\mu_m, \Sigma)$
\RETURN $\mathcal{D}_{syn}$
\end{algorithmic}
\end{minipage}}
\end{algorithm}

\subsubsection{Meta-learning}\
\label{s326}

Meta-learning, often referred to as the process of learning to learn, aims to create a model capable of effectively generalizing across a variety of tasks rather than concentrating on instances from a single task. This approach leverages acquired meta-knowledge to address the shortage of training data for individual tasks and has been investigated for adapting models to imbalanced data distributions. Meta-learning methods achieve this goal through bi-level optimization, which consists of both inner and outer loops. The inner loop focuses on task-specific learning by training a model to excel at individual tasks. Meanwhile, the outer loop emphasizes learning across tasks and updates the model based on its performance on the inner loop tasks. To enhance the model's generalization and adaptability to imbalanced data distributions, researchers propose to optimize the model from multiple perspectives, which includes {weights, sampling method, model dynamics, data augmentation and class prototypes}.

\textbf{Weights.}
The majority of meta-learning-based methods \cite{shu2019meta, lee2020l2b, jamal2020rethinking, zhang2021learning, liu2021improving} concentrate on re-weighting methods for imbalanced data. {The pseudo-code is depicted in Algorithm \ref{alg:metalearning}.} These studies highlight the limitations of infeasible manually designed weighting functions and the challenges associated with applying hyper-parameters in loss functions. Consequently, these methods treat the weights for training samples as learnable parameters and optimize them accordingly.

\begin{algorithm}[t]
\caption{{Meta-Weight Learning}}
\label{alg:metalearning}
\resizebox{\linewidth}{!}{
\begin{minipage}{\linewidth}
\begin{algorithmic}[1]
\INPUT Training set $\mathcal{D}_{train}$, validation set $\mathcal{D}_{val}$, model $f$, meta-learner $g$
\OUTPUT Trained model $f$, updated weight function $g$
\FOR{each mini-batch $\{(x_i, y_i)\} \subset \mathcal{D}_{train}$}
    \STATE Compute weights: $w_i \gets g(x_i, y_i)$
    \STATE Compute weighted loss: $\mathcal{L}_{train} = \sum w_i \cdot \ell(f(x_i), y_i)$
    \STATE Simulate gradient step: $f' \gets f - \alpha \cdot \nabla_f \mathcal{L}_{train}$
    \STATE Sample batch $\{(x_j^{val}, y_j^{val})\} \subset \mathcal{D}_{val}$
    \STATE Compute validation loss: $\mathcal{L}_{val} = \sum \ell(f'(x_j^{val}), y_j^{val})$
    \STATE Update $g$: $g \gets g - \beta \cdot \nabla_g \mathcal{L}_{val}$
\ENDFOR
\RETURN $f$, $g$
\end{algorithmic}
\end{minipage}}
\end{algorithm}

LRE \cite{ren2018learning} first uses a clean and unbiased dataset as the meta set (i.e., development set) to guide the training procedure. By performing validation at every training iteration, LRE dynamically assigns importance weights for each batch sample.
Meta-Weight-Net \cite{shu2019meta} automatically learns an explicit loss-weight function, parameterized by a multi-layer perceptron (MLP) from data, resulting in more stable learning performance. 
Bayesian TAML \cite{lee2020l2b} learns a set of class-specific scalars to weight the learning rate of class-specific gradients for each inner loop optimization step, thus giving more attention to minority classes. 
MLCW \cite{jamal2020rethinking} analyzes the imbalance problem from a domain adaptation perspective. They assume that the class-conditioned distribution of training data and test data is not the same, especially for minority classes with insufficient training data. Therefore, MLCW proposes to explicitly account for the differences between conditioned distributions and utilizes meta-learning to estimate the conditional weights for each training example. 
FSR \cite{zhang2021learning} argued against the expensive second-order computation of nested optimizations in meta-learning and proposed a sample re-weighting method. By learning from history to build proxy reward data and employing feature sharing, FSR effectively reduces optimization costs.

\textbf{Sampling method.}
Besides learning adaptive weights for training sets, meta-learning method is deployed to learn a sampling method to re-balance the training data. Balanced Meta-Softmax \cite{ren2020balanced} introduces a balanced softmax method and argues that balance sampling for the mini-batch sampling process might lead to an over-balancing problem. In this scenario, minority classes dominate the optimization process, resulting in local minimums that favor these minority classes. To address this issue, Balanced Meta-Softmax introduces a meta sampler to explicitly learn the optimal sample rate for each class under the guidance of class-balanced meta set.
Instead of co-optimized with the classification model, MESA \cite{liu2020mesa} decouples the model-training and meta-training process and proposes a meta sampler that serves as an adaptive under-sampling solution embedded in the iterative ensemble training process. By incorporating the classification error distribution of ensemble training as its input, MESA achieves high compatibility and is applicable to various statistical and non-statistical classification models, such as decision trees, Naïve Bayes, and k-nearest neighbor classifiers.

\textbf{Model dynamics.}
MetaModelNet \cite{wang2017learning} utilizes meta-learning to learn model dynamics, predicting how model parameters will change as more training data is progressively incorporated. As a result, MetaModelNet can directly enhance model parameters trained on minority instances by predicting the parameters when additional minority samples are included.

\textbf{Data augmentation.}
MetaSAug \cite{li2021metasaug} employs meta-learning for data augmentation. Inspired by ISDA \cite{wang2019implicit}, it calculates class-wise covariance matrices to extract semantic directions and generates synthesized minority instances along these semantically meaningful directions. However, the scarcity of data for minority classes results in inaccurate covariance matrices. To address this issue, MetaSAug incorporates meta-learning and learns the most meaningful class-wise covariance by minimizing the loss on a balanced meta set. Consequently, MetaSAug trains models on an augmentation set enriched with semantically augmented samples, effectively enhancing their performance.

\textbf{Class prototypes.}
MPDT \cite{fu2022meta} focuses on calibrating empirical prototypes that are significantly biased due to scarce data in minority classes. MPDT first introduces meta-prototype contrastive learning to automatically learn the reliable representativeness of prototypes and predict the target distribution of balanced training data based on prototypes. Subsequently, additional features are sampled from the predicted distribution to further balance the overwhelming dominance of majority classes.

%% file: 3.3Trainingstrategy.tex
\subsection{Training strategy}
\label{s33}

\begin{figure}[t]
\centering
\includegraphics[width=\linewidth]{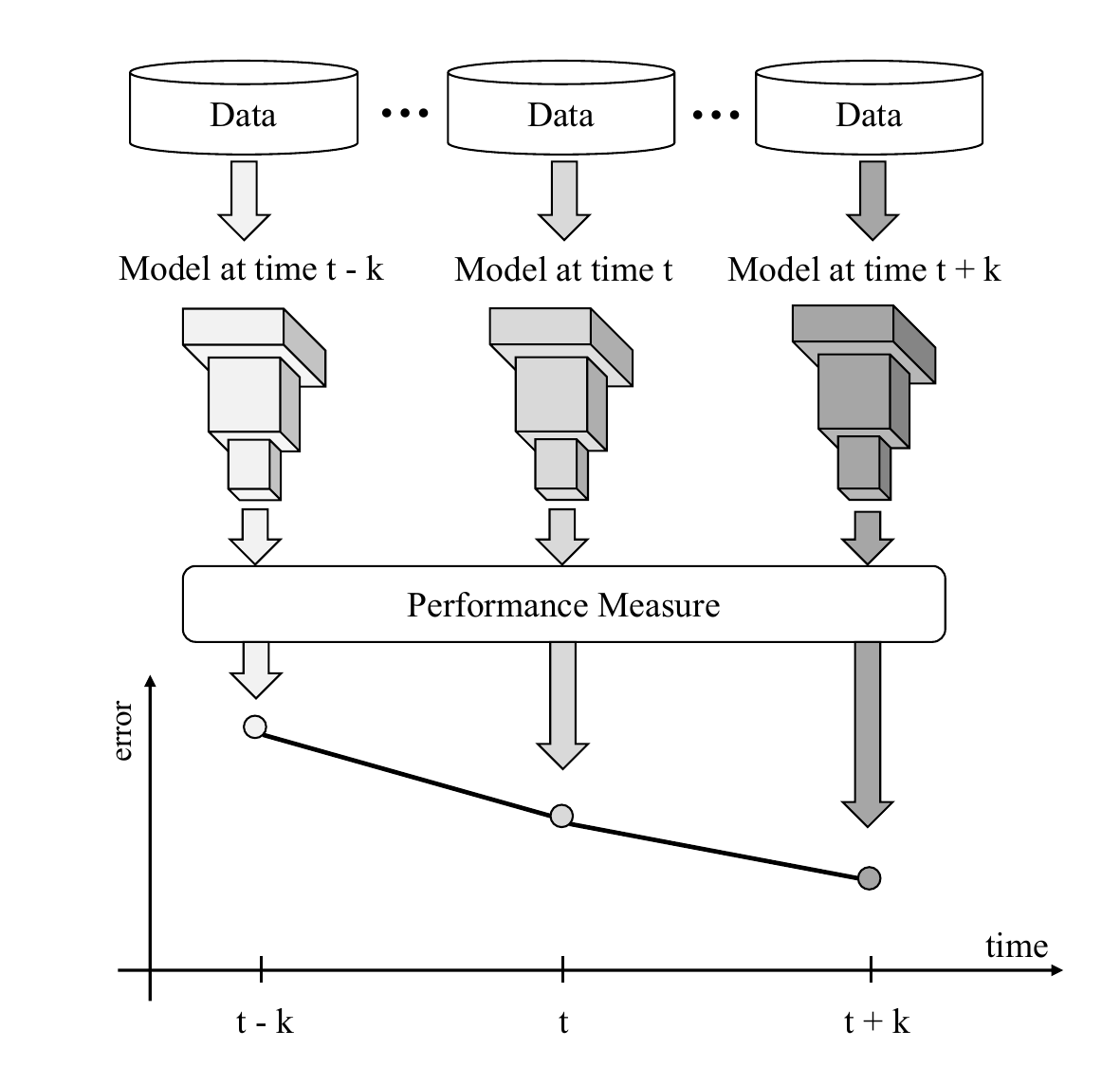}
\caption{{The training strategy exploits different stages of the training pipeline to guide the model toward minimizing its inherent bias toward majority classes. Notably, these strategies often involve the separate use of different data subsets to enhance learning effectiveness.}}
\label{fig_4}
\end{figure}

{Training strategies, as shown in Fig.~\ref{fig_4}, exploit different stages of the training pipeline to guide the model toward minimizing its inherent bias toward majority classes, which includes decoupled training, pretraining, curriculum learning, and post-hoc calibration.}

\subsubsection{Decoupling training}\
\label{s331}

The concept of decoupling training was first formally evaluated by cRT \cite{kang2019decoupling}, which separates the model learning procedure into two distinct processes: representation learning and classification, {as shown in Algorithm \ref{alg:decouple}}.  They systematically explored how imbalanced data affects each process individually and found that data imbalance might not be a significant issue in learning high-quality representations. Moreover, they demonstrated that the model could achieve robust classification performance by only retraining the classifier using class-balanced sampling. 
Similarly, LDAM \cite{cao2019learning} observes empirically that re-weighting and re-sampling are both inferior to the vanilla empirical risk minimization (ERM) algorithm (where all training examples have the same weight) before annealing the learning rate. Thus, LDAM defers the re-balancing or re-sampling training procedure to the classification stage. It initially trains the model using vanilla ERM and anneals the learning rate to obtain a better representation. Subsequently, re-weighting or re-sampling strategies are implemented with a smaller learning rate to adjust the decision boundary and fine-tune the features locally, leading to improved performance compared to one-stage training.

Following the work of cRT and LDAM, numerous methods have adopted the decoupling scheme and focused on enhancing feature representation through supervised and self-supervised learning approaches.
KCL \cite{kang2021exploring} leverages the superiority of supervised contrastive learning and further improve it by balancing the positive pairs, ensuring that there is an equal number of positive samples for each anchor sample. 
In contrast, BaCon \cite{kang2022comprehensive} studies introduce a balance term in the supervised contrastive loss to re-weight the negative pairs. During the classifier fine-tuning phase, they propose a generalized normalization classifier, which is more effective in preserving the length information of feature vectors and achieves more powerful classification ability.
{Instead of using one-hot labels, Logits Retargeting (LORT) \cite{sunrethinking} redistributes label probability mass by assigning a small portion of positive probability and spreading larger negative probabilities across other classes. This "label smooth value" encourages balanced logits magnitudes, reduces perturbations during training, and enhances class separability. Unlike conventional Label Smoothing, LORT sets a relatively large negative probability, which is crucial for improving convergence and discrimination.}

Instead of utilizing the label information in representation learning, SSP \cite{yang2020rethinking} finds that self-supervised method could yields a great initialization that is more label-agnostic from the imbalanced dataset.  
rwSAM \cite{liuself} systematically compares features pre-trained by self-supervised learning and supervised learning, both theoretically and empirically and finds that self-supervised learning can learn label-irrelevant but transferable features that aid in classifying minority classes better than supervised pre-training. To further reduce the influence of biased data distribution on self-supervised pre-training, rwSAM adapts sharpness-aware minimization to improve model generalization and calculates loss weight using kernel density estimation, which does not require training data labels.
SCDL-GAN \cite{cai2019supervised} employs the Wasserstein auto-encoder architecture to represent sample distributions previously. Guided by label information, SCDL-GAN can represent the distributive characteristics of all classes, avoiding biased minority representation and contributing to a well-performing GAN-based minority synthesis.
SSD \cite{li2021self} proposes training an initial model under label supervision and self-supervision simultaneously using instance-balanced sampling. The pre-trained model is then used to provide informative soft labels as supervision signals during the classifier training procedure. By using self-distillation, the inter-class information is integrated into the soft labels and transferred to the minority classes.

\begin{algorithm}[t]
\caption{{Decoupled Training with Balanced Classifier Fine-tuning}}
\label{alg:decouple}
\resizebox{\linewidth}{!}{
\begin{minipage}{\linewidth}
\begin{algorithmic}[1]
\INPUT Imbalanced training data $\mathcal{D}_{\text{imb}}$, balanced data $\mathcal{D}_{\text{bal}}$, backbone $f$, classifier $g$
\OUTPUT Final model $g \circ f$
\FOR{each epoch}
    \FOR{each batch $(x, y) \sim \mathcal{D}_{\text{imb}}$}
        \STATE Extract features $z = f(x)$
        \STATE Compute loss $\mathcal{L}_{\text{rep}} = \ell(g(z), y)$
        \STATE Update both $f$ and $g$ using gradient descent
    \ENDFOR
\ENDFOR
\STATE Freeze backbone $f$
\vspace{0.3em}
\STATE Re-initialize classifier $g$
\FOR{each epoch}
    \FOR{each batch $(x, y) \sim \mathcal{D}_{\text{bal}}$}
        \STATE Extract features $z = f(x)$ using the frozen backbone
        \STATE Compute loss $\mathcal{L}_{\text{cls}} = \ell(g(z), y)$
        \STATE Update classifier $g$ only
    \ENDFOR
\ENDFOR
\RETURN Final model $g \circ f$
\end{algorithmic}
\end{minipage}}
\end{algorithm}

\subsubsection{Fine-tuning pre-trained model}\
\label{s332}

In contrast to the decoupling approach, which trains feature representations and classifiers on the same dataset, fine-tuning from a well pre-trained deep model, which is also called transfer learning, has been found to yield impressive performance for many downstream tasks. This method has been proven to improve model robustness and uncertainty estimates \cite{hendrycks2019using}, while also benefiting imbalanced data representation. Specifically, the model is first pre-trained on a large, related dataset (source domain) to learn rich, unbiased feature representations. Subsequently, fine-tuning is conducted on the target dataset (target domain) to better adapt to the specific task.

FDM \cite{ouyang2016factors} initially examines the impact of imbalance in this paradigm, finding that majority classes have a more significant influence on feature learning, and achieving a more uniform distribution of sample numbers across classes is preferable. To counteract the effects of imbalance, FDM proposes a hierarchical feature learning approach that clusters objects into visually similar class groups and learns deep representations for these groups separately. This facilitates learning specific feature representations for individual classes.
Similarly, taking into account the heavy computation and irrelevant features learned from large-scale datasets, DSTL \cite{cui2018large} proposes to first measure the similarity between the target domain and source domain, selecting the most relevant samples from the source domain for pre-training the model. Subsequently, fine-tuning is performed on a more evenly-distributed subset of the target dataset, balancing the network's efforts among all categories and facilitating the transfer of learned features.
It is worth noting that the source domain and target domain do not require a strict or explicit correlation. By utilizing a large-scale dataset (i.e., ImageNet) as the source domain, IBCC \cite{singh2020imbalanced} can apply the learned basic texture knowledge to a target domain consisting of histopathological images, achieving impressive performance in handling imbalanced medical data.

\subsubsection{Curriculum learning}\
\label{s333}

Curriculum learning \cite{bengio2009curriculum} is the strategy of learning from easy to hard, which can significantly improve the generalization of the deep model. To achieve this, the curriculum schedules focus on sampling, optimization, and re-weighting strategies.
CASED \cite{jesson2017cased} introduces a curriculum sampling method for the task of extreme imbalanced lung nodule detection in chest CT scans. Initially, CASED focuses on learning how to distinguish nodules from their immediate surroundings and gradually incorporates a greater proportion of difficult-to-classify global context until it reaches uniform sampling from the empirical data distribution. The curriculum sampling strategy effectively addresses the imbalance problem but also enhances the model's robustness against nodule annotation quality issues.
CurriculumNet \cite{guo2018curriculumnet} designs the sampling curriculum by measuring the complexity of data using its distribution density in a feature space, and then ranks the complexity in an unsupervised manner. Then the model capability is improved gradually by continuously adding the data with increasing complexity into the training process.
BBN \cite{zhou2020bbn} proposes a bilateral-branch network structure and utilizes each branch to capture the imbalanced distribution and reverse distribution respectively. In the training procedure, a cumulative learning strategy is equipped to first learn the universal patterns of whole dataset and then pay attention to the minority class gradually.

Beside the sampling curriculum from imbalance to balance, DCL \cite{wang2019dynamic} further incorporates metric learning, which emphasizes learning a soft feature embedding to separate different samples in the feature space. With the expectation that the model will first learn suitable feature representations before classifying samples into correct labels, a loss scheduler is proposed to shift the learning objective from metric learning loss to classification loss.

SEDC \cite{zhao2022diagnosing} introduces a dual-curriculum learning strategy that performs feature re-weighting using adaptively updated weighting factors for each sample. By assigning different feature weight and sample weight, SEDC learns the optimal decision boundary by balancing the training contributions between majority and minority classes.

\subsubsection{Posterior re-calibration}\
\label{s334}

Posterior re-calibration refers to techniques that are designed from statistical perspectives to apply post-hoc adjustments to model outputs. The idea of calibrating model outputs is first leveraged in probability estimation where models are expected to generate reliable probability estimates of events (e.g. the probability of a sample falling into a certain class) \cite{wallace2012class}. Recently, significant strides have emerged in adapting posterior calibration techniques for discriminative tasks, specifically addressing challenges posed by imbalanced data. We categorize the existing model calibration methods into two main categories: Weight normalization and Logit re-balancing.

\textbf{Classifier weight normalization.} The weight normalization technique is initially introduced along with decoupling training in \cite{kang2019decoupling}. In their two-stage training approach, they observed that following joint training with instance-balanced sampling, the classifier's weight norms correlate with the cardinality of the classes. Inspired by this observation, they proposed the $\tau$-normalization technique to re-scale the weight matrix in the second traing phase or the inference phase. Following this work, DisAlign \cite{zhang2021disalign} was proposed to address the performance bottleneck in the biased decision boundary of the existing imbalanced methods. It extended the second training phase by introducing additional learnable parameters to the classifier and utilizing a KL divergence-based re-weighted loss. Gaussian Clouded Logit Adjustment \cite{li2022long} discovered that the learned representations within imbalanced datasets remain insufficient, particularly noting a severe compression of embedding spaces associated with the {minority} classes. To this end, it provided another solution to calibrate model output which combines normalization and random perturbation.

\begin{table}[t]
\centering
\caption{{Glossary of Symbols}}
    \resizebox{\linewidth}{!}{
\begin{tabular}{ll}
\toprule
\textbf{Symbol} & \textbf{Description} \\
\midrule
$x$ & Input feature vector (sample) \\
$y$ & Class label corresponding to input $x$ \\
$p_s(y)$ & Prior label distribution in the source (training) set \\
$p_t(y)$ & Prior label distribution in the target (test) set \\
$p_s(x \mid y)$ & Conditional probability of $x$ given label $y$ in the source domain \\
$p_t(x \mid y)$ & Conditional probability of $x$ given label $y$ in the target domain \\
$f_y(x)$ & Logit output for class $y$ given input $x$ \\
% $f_y(x) \leftarrow f_y(x) + \log p_t(y) - \log p_s(y)$ & Logit adjustment rule based on label prior shift \\
\bottomrule
\end{tabular}}
\label{glossary}
\end{table}

\textbf{Logit re-balancing.} This type of methods are based on one common probabilistic depiction of data imbalance: label prior shift. It signifies a difference between the prior label distributions of the training set, denoted as $p_s(y)$, and those of the test set, $p_t(y)$ ({Please refer to Table \ref{glossary} for details}). This difference exists while maintaining identical conditional probabilities of
samples given the class label, expressed as $p_s(x|y) = p_t(x|y)$. Based on this depiction, we may consider applying an additive adjustment to the logit output during inference: $f_y(x) \leftarrow f_y(x) + \log p_t(y) -\log p_s(y)$, which forms the main idea of balanced softmax \cite{hong2021disentangling, ren2020balanced}. For the setting where labels in test set are uniformly distributed, \cite{menon2021long} derived a similar logit adjustment formulation from another perspective. They draw inspiration from an alternative optimization objective, balanced error, which is the average of per-class error rates. Given that the target label distribution may also be imbalanced in practice, LADE \cite{hong2021disentangling} considered that Balanced Softmax does not involve the disentanglement of $p_s(y)$ during the training phase. It extended the original balanced softmax by utilizing regularized DV representation and importance sampling and proposed a novel loss function to further improve the performance. \cite{ye2022identifying} and \cite{tian2020posterior} extended the balanced softmax by adding a hyper-parameter as the scale of adjustment. UNO-IC \cite{tian2020posterior} further combines this parameterized calibration method with temperature scaling and the Noisy-Or model, forming an effective multi-modal fusion algorithm.

Recently, \cite{wang2024unified} conducted a systematical analysis which succeeds in explaining the role of re-weighting and logit-adjustment in a unified manner, as well as some empirical results that are out of the scope of existing theories. As no additional parameters are involved, posterior recalibration manages to provide decent performance gains at a low computational cost, especially in the setting of imbalanced classification.

%% file: 3.4Ensemble.tex
\subsection{Ensemble}
\label{s34}

\begin{figure}[t]
\centering
\includegraphics[width=\linewidth]{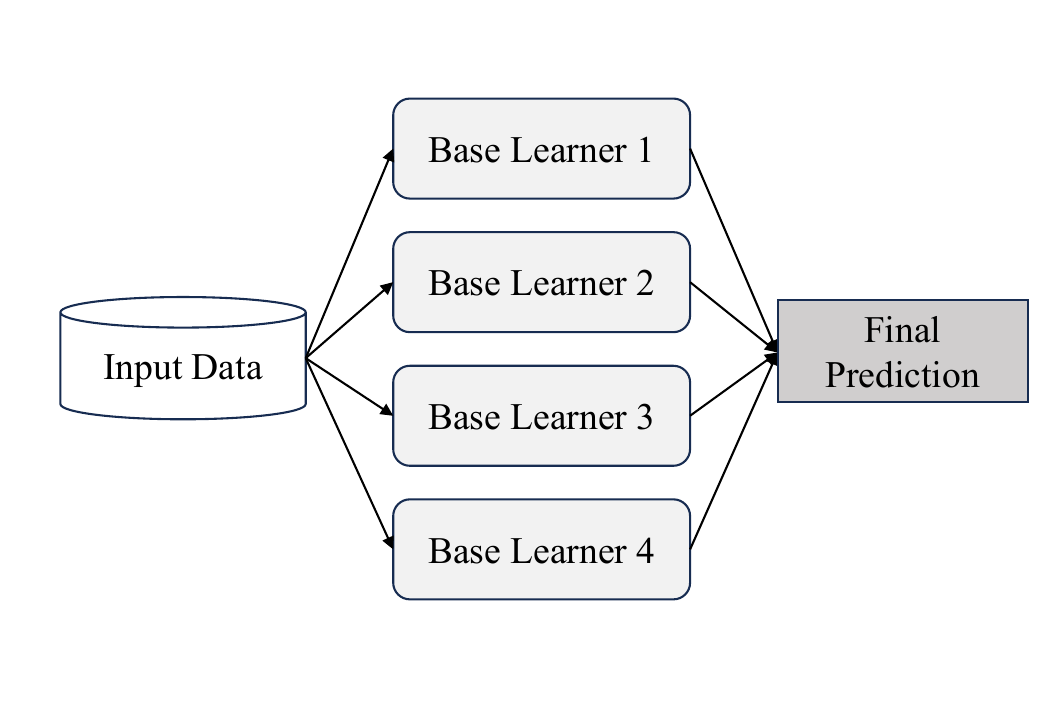}
\caption{{Ensemble methods combine multiple base models and leverage the strengths of diverse learners.}}
\label{fig_3}
\end{figure}

{Ensemble methods combine multiple base models and leverage the strengths of diverse learners as depicted in Fig.~\ref{fig_3}. By aggregating predictions, they help mitigate the bias introduced by imbalanced data and improve the overall robustness and accuracy of the final model outputs.}

\subsubsection{Bagging-based ensemble}\
\label{s341}
% In recent years, bagging-based methods have emerged as effective techniques to deal with imbalanced data by manipulating the training set and creating diverse classifiers. 

% Bagging for Imbalance Learning
\begin{algorithm}[t]
\caption{{Bagging Ensemble}}
\label{alg:bagging}
\resizebox{\linewidth}{!}{
\begin{minipage}{\linewidth}
\begin{algorithmic}[1]
\INPUT Imbalanced dataset $\mathcal{D}$, number of models $K$, base learner $f_k$
\OUTPUT Ensemble classifier $F(x)$
\FOR{$k = 1$ to $K$}
    \STATE Sample balanced subset $\mathcal{D}_k$ from $\mathcal{D}$ via class-wise resampling
    \STATE Train base learner $f_k$ on $\mathcal{D}_k$
\ENDFOR
\STATE Aggregate predictions: $F(x) = \text{Aggregate}(f_1(x), \ldots, f_K(x))$
\RETURN $F(x)$
\end{algorithmic}
\end{minipage}}
\end{algorithm}

% is an ensemble learning technique that combines multiple base classifiers to form a stronger ensemble learner.
% The concept of bagging 
In the context of imbalanced data learning, Bagging, short for Bootstrap Aggregating, involves generating multiple balanced subsets of the imbalanced dataset by resampling techniques like random undersampling and SMOTE, and training base classifiers on these subsets independently \cite{breiman1996bagging}. The final prediction is obtained through the aggregation of predictions made by each base classifier using majority voting or weighted voting {as shown in Algorithm \ref{alg:bagging}}. Apart from rebalancing the class distribution, it aims to counterbalance the bias towards the majority class by introducing diversity among the base classifiers.

Over-Bagging~\cite{ref_22} algorithm replaces the random sampling technique with the random oversampling technique in Bagging algorithm to deal with the imbalance of data. Data balance is achieved by over-sampling the {minority samples}. However, it may generate overfitting and redundant classifiers due to the high similarity between and within bags. This is because random oversampling can create duplicate instances. Hence, SMOTE-Bagging was proposed to alleviate the replication issue. 

SMOTE-Bagging~\cite{ref_22} overcomes this disadvantage by using SMOTE to generate synthetic minority class via interpolating between existing ones. It ensures that the synthetic instances are different from the original ones and promote the diversity of bags. By using a varying resampling rate in each iteration, SMOTE-Bagging also ensures that each bag is different from the others, which further increases the diversity of the ensemble. Specifically, in each bag, the {minority samples} consist of synthetic instances and resampled instances, where the resampling rate starts at 10\% in the first iteration and increases to 100\% in the last iteration. Therefore, SMOTE-Bagging can generate a more diverse ensemble than Over-Bagging, which can lead to better classification performance on the minority class.

% However, the training set used by the base classifier will be too large as each iteration oversampling is faced with all the {majority samples}, this will affect the efficiency of classification learning. The SMOTE Bagging algorithm proposed to solve this problem. 

Similarly, some papers suggest that the Bagging algorithm can balance data by undersampling. Thus, the Under-Bagging \cite{ref_22} algorithm is proposed. On contrary to Over-Bagging, it undersamples the {majority samples} in the process of iteratively generating training sets for base classifiers. Repeated sampling with the {majority samples} can also obtain a large difference in the base classifier, and compared with Over-Bagging, the number of samples per iteration will be less and the efficiency will be higher. However, in the process of under-sampling, it is easy to neglect most of the useful samples, resulting in inaccurate classification results.

% As we know, The Bagging algorithm is simple to implement and has strong generalization ability, thus some studies have proposed using Bagging to deal with data imbalance. Bagging-based classification algorithm with data processing is simpler than Boosting-based algorithm as it does not need to calculate and update weight. In these algorithms, the over-sampling and under-sampling techniques are introduced to rebalance the data, which ensures the difference of the base classifier and the learning accuracy of the ensemble classifier.

Following this research line, numerous variants have been proposed from different viewpoints. For example, Roughly Balanced Bagging (RBBag) was proposed to generate more diverse bags \cite{hido2009roughly}. 
% That is, in each iteration of RBB, a new bag is created by randomly sampling with replacement from the majority class and sampling without replacement from the minority class. The number of positive examples is kept fixed, while the number of negative examples is drawn from a negative binomial distribution. 
That is, in each iteration of RBBag, the number of positive samples is kept fixed, while the number of negative samples  satisfies negative binomial distribution. The diversity of the bags is achieved by varying the number of negative samples, while the rough balance of the bags ensures both the majority and minority classes are equally emphasized. 
From another perspective, Neighbourhood Balanced Bagging (NBBag) \cite{blaszczynski2015neighbourhood} was designed to modify the sampling probabilities of different instances in a way that focuses on the minority samples that are hard to learn while decreasing the probabilities of selecting samples from the majority class. This is achieved by analyzing the neighbourhood of samples, and taking into account the local minority class distribution. It is still a open question that how to resample instances for more accurate and diverse base classifiers and consequently better overall performance. 

\subsubsection{Boosting-based ensemble}\
\label{s342}

% Boosting for Imbalance Learning
\begin{algorithm}[t]
\caption{{Boosting Ensemble}}
\label{alg:boosting}
\resizebox{\linewidth}{!}{
\begin{minipage}{\linewidth}
\begin{algorithmic}[1]
\INPUT Dataset $\mathcal{D} = \{(x_i, y_i)\}$, rounds $T$, base learner $f$
\OUTPUT Weighted ensemble $\{(f_t, \alpha_t)\}$
\STATE Initialize weights $w_i \propto 1/N_{y_i}$
\FOR{$t = 1$ to $T$}
    \STATE Normalize weights $w_i$
    \STATE Train $f_t$ on $\mathcal{D}$ with $w_i$
    \STATE Compute error $\varepsilon_t$; set $\alpha_t = \log((1 - \varepsilon_t)/\varepsilon_t)$
    \STATE Update: $w_i \leftarrow w_i \cdot \exp(\alpha_t)$ if $f_t(x_i) \neq y_i$
\ENDFOR
\RETURN $\{(f_t, \alpha_t)\}$
\end{algorithmic}
\end{minipage}}
\end{algorithm}

Boosting-based ensemble methods aim to enhance the performance of classifiers on imbalanced data through iterative re-weighting of the training samples, to focus on the minority class, {as shown in Algorithm \ref{alg:bagging}}. The most widely used boosting-based ensemble methods are AdaBoost \cite{hastie2009multi}, Gradient Boosting \cite{friedman2001greedy}, and XGBoost \cite{chen2016xgboost}. Especially, in AdaBoost, each base classifier is trained on weighted samples. The weights of misclassified samples are increased with each iteration, while the weights of correctly classified samples are decreased. The final classifier is a weighted combination of base classifiers, where the weights are determined by their classification accuracy.

Based on the Adaboost and the SMOTE algorithm, the SMOTE-Boost~\cite{ref_3} method is proposed. The SMOTE oversampling technique is introduced in each iteration. By adding synthetic minority class examples, each base classifier can pay more attention to the minority class. 
% In the SMOTE-Boost algorithm, data balance is achieved by iterating training sets with different weights of the base classifier to improve the difference of the base classifier and the final classification accuracy. 
However, SMOTE-Boost fails to consider the distribution of minority classes and latent noises when it generates synthetic examples, which can be blind and harmful to classification accuracy. To alleviate this problem, the MSMOTE-Boost~\cite{ref_8} algorithm is proposed as improvement. Instead, it uses the MSMOTE algorithm to process the unbalanced data. In MSMOTE, by the distance between samples, the {minority samples} are divided into three groups: safety samples, boundary samples, and potential noise samples. For security samples, the synthesis algorithm is the same as that of SMOTE; for boundary samples, the nearest sample is chosen; and for potential noise samples, no operation is performed. From another perspective, RAMOBoost \cite{chen2010ramoboost} emphasizes to consider the learnability of {minority samples}. RAMOBoost employs an adaptive sampling strategy at each boosting iteration, to focus on difficult-to-learn {minority samples}. The sampling distribution is updated based on the pseudo-loss of the current hypothesis, which is a measure of the error rate of the hypothesis on the training data. The updated sampling distribution gives more weight to examples that are misclassified by the current hypothesis, which can help the algorithm to focus on difficult-to-learn examples.

Corresponding to the over-sampling boosting algorithms mentioned above is the boosting algorithms combined with under-sampling. The key idea of the RUS-Boost algorithm~\cite{ref_20} is to select samples randomly from the {majority samples} by using the random undersampling technique (RUS) in the iteration process of the AdaBoost algorithm. Compared with the SMOTE-Boost, this algorithm has the advantages of simple implementation and short training time, but it is possible to remove many potentially useful samples in under-sampling. To avoid this problem, the EusBoost~\cite{ref_21} algorithm is proposed. This algorithm uses the evolutionary under-sampling technique, which chooses the most representative samples in the {majority samples} to get more informative dataset, and it introduces the fitness function to ensure the difference of the base classifier.
Another improvement to the RUSBoost algorithm is RHSBoost. 
The key idea of RHSBoost \cite{gong2017rhsboost} is to combine the advantages of RUS and ROSE (Random Over-Sampling Examples) \cite{lunardon2014rose} sampling to generate more balanced and diverse subsets. Similar to SMOTE, at each boosting iteration, a weighted ROSE sampling algorithm is first used to oversample the minority class. Then, RUS is used to further balance the class distribution by randomly removing some instances from the majority class.
Note that ROSE and SMOTE differ in the way they generate synthetic examples. SMOTE generates synthetic examples by interpolating, while ROSE samples from a multivariate kernel density estimate of the minority class, which can maintain the characteristics of the minority class and help avoid overfitting. 
Apart from those algorithms mentioned above, there are also many boosting-based methods proposed in recent years, such as LIUBoost \cite{ahmed2019liuboost} and HUSBoost \cite{popel2018hybrid}, showing boosting-based methods are attractive and promising for imbalance data learning especially with undersampling technique.

\subsubsection{Cost-sensitive ensemble}\
\label{s343}

% Cost-sensitive ensemble methods are a family of techniques that aim to address the class imbalance problem by incorporating the cost of misclassification into the ensemble learning process. 
In scenarios such as fraud detection and medical diagnosis, the misclassification cost of different classes could vary significantly, leading to potential risks or financial losses when traditional methods applied, which simply minimizes the average error rate or maximizes overall accuracy \cite{ling2008cost}.
% Therefore, it becomes crucial to develop techniques that can take into account these costs and make more informed decisions during the classification process.
The key idea of cost-sensitive ensemble methods is to incorporate cost considerations into the construction and combination of base classifiers. To be more specific, these proposals usually differ in the way that they modify the weight update rule.

Among this family, AdaCost \cite{fan1999adacost}, CSB0, CSB1, CSB2 \cite{ting2000comparative}, AdaC1, AdaC2, and AdaC3 \cite{sun2007cost} are the most representative approaches. 
AdaCost introduces a cost adjustment function into the weight updating rule of AdaBoost. This adjustment function increases the weights of costly misclassifications more aggressively, and decreases the weights of costly correct classifications more conservatively. Theoretical analysis also provides an tighter upper bound on the cumulative training misclassification cost than AdaBoost.

CSB0 modifies the weight update rule by multiplying the weight update term by a dedicated cost function parameterized by both classifier prediction and groundtruth. CSB1 improves this vanilla version by taking the confidence of the base learner's prediction into account. 
CSB2, on the other hand, incorporates an additional factor into the weight update rule. This factor is a function of the misclassification cost and the prediction made by the base learner, i.e. the weight of current base learner in cost-sensitive scenario. By utilizing more informative factors, CSB2 is able to further minimize high cost errors.

Likewise the CSB family, AdaC1, AdaC2 and AdaC3 mainly differ in their weight update rules. In AdaC1, the cost function is introduced within the exponent part, multiplying by the weight of base classifier.
But in AdaC2, the cost function is introduced outside the exponent part as another possible attempt. In AdaC3, this modification considers the idea of AdaC1 and AdaC2 at the same time. The weight update formula is modified by introducing the cost function both inside and outside the exponent part. Note that with the change of the cost function, the weight assignment rule for base classifiers is also changing. These details are omitted for simplicity.

% These methods aim to optimize the ensemble such that it minimizes the expected cost of misclassification over the dataset, rather than simply minimizing the error rate or maximizing accuracy.

\subsubsection{Ensemble knowledge distillation} \
\label{s344}

In the era of deep learning, numerous conventional machine learning models such as SVM, KNN, and DT have been replaced by deep learning models, which offer higher accuracy. However, these deep learning models come with a significant drawback: they require substantial computational resources and storage consumption. To address this issue, knowledge distillation was introduced as a compression technique \cite{gou2021knowledge}. 
% Ensemble knowledge distillation, an extension of traditional knowledge distillation, was further proposed in the context of ensemble learning.
Ensemble knowledge distillation extends the concept of traditional knowledge distillation \cite{asif2019ensemble}. Instead of transferring knowledge from a single teacher model, ensemble knowledge distillation transfers knowledge from an ensemble of teacher models. By leveraging the collective knowledge of the ensemble, the student model learns to mimic the collective behavior, resulting in improved robustness and accuracy.

Supervision signals from teacher models can include both high-level and intermediate-level information, namely, soft labels and intermediate layer representations, where the latter is generally considered more informative. 
The diverse capabilities of teachers necessitate selecting the most suitable supervision signals or assigning importance weights based on their expertise. 

For example, in \cite{chebotar2016distilling}, student model is simply distilled with the weighted average of teacher models, where weights are fixed as hyperparameters.  And in \cite{fukuda2017efficient}, two strategies are investigated for utilizing labels from multiple teachers. The first one keeps the same with \cite{chebotar2016distilling}, while the second randomly selects a teacher model at the mini-batch level. For more flexible and finer grained weights, AMTML-KD \cite{liu2020adaptive} associates each teacher with a latent representation to adaptively learn instance-level weights, and gather intermediate-level hints from teachers using a multi-group based strategy. Concurrently, RL-KD also proposes to assign instance-level weights through a reinforcement learning process \cite{yuan2021reinforced}. In RL-KD, the agent receives training instances and applies the teacher selection policy for distillation, where the reward is defined as the performance of the trained student model. Apart from weight assignment, CTKD \cite{zhao2020highlight} designs two dedicated teachers in a collaborative manner. Specifically, one teacher trained from scratch assists the student step by step using its temporary outputs, and the other pre-trained expert teacher guides the student to focus on a critical region. Inspired by the pre-training and fine-tunning paradigm, TMKD \cite{yang2020model} first pre-trains the student, and further fine-tune the pre-trained model on downstream tasks. Both stages reduce the overfitting bias and transfer general knowledge to the student model. 
{Collaborative Global-Local Structure Network (CGL-Net) \cite{wu2024collaborative} addresses the limitations of multi-expert networks for imbalanced classification, where inconsistent expert performance and underutilized data structure hinder generalization. CGL-Net decouples representation learning into two complementary stages: global structure learning, where a student network distills knowledge from diverse experts to achieve balanced category performance; and local structure learning, where augmented data guides the model to capture discriminative regional patterns of individual objects. By hierarchically integrating global and local knowledge, combined with knowledge distillation and multi-expert collaboration, CGL-Net produces robust and controllable representations.}
Note that ensemble knowledge distillation can be also combined with other technique such as parameter pruning and network quantization for more compact model. For instance, \cite{pham2023collaborative} proposes a novel framework that leverages both multi-teacher knowledge distillation and network quantization for learning low bit-width DNNs.

%% file: 4.Data.tex
\section{Imbalance Across Diverse Data Formats}
\label{sec:data}
\subsection{Image data}
\quad The rapid advancement of computer vision benefits greatly from the expansion of datasets. However, keeping data balanced becomes hard as data size grows in real-world image collection efforts. Consequently, data imbalance, especially in the form of {imbalanced} distribution, emerges as a widespread challenge across various computer vision tasks such as image classification, object detection and segmentation, leading to a decline in model performance. In the following sections, we will outline commonly used methods to mitigate the data imbalance problem within these tasks.

\subsubsection{Image Classification}\ 

Image classification stands as one of the foundational tasks in the realm of computer vision. Many image classification datasets as well as various domain-specific collections, exhibit {imbalanced} distribution patterns. In this scenario, models tend to show bias towards the {majority} classes, resulting in lower predicting accuracy in {minority} classes. Numerous studies have delved into resolving the issue of class imbalance in image classification, spanning across the categories we previously discussed. In terms of improving data qualities, resampling \cite{kang2019decoupling, ren2020balanced, wang2019dynamic} and data augmentation \cite{kim2020m2m, tan2021equalization, chu2020feature} are frequently employed, mainly to balance the amount of data in different classes. \cite{kang2019decoupling} evaluates the performance of various sampling strategies for {imbalanced} recognition under both joint and decoupled learning schemes. Dynamic Curriculum Learning (DCL) \cite{wang2019dynamic} integrates the idea of curriculum learning into a sampling scheduler to train the model dynamically from imbalanced to balanced and from easy to hard. Major-to-Minor translation (M2m)  \cite{kim2020m2m} expands the oversampling approach by translating majority samples to minority ones using the perturbation-based optimization. Regarding model enhancements, a variety of techniques spanning model design \cite{wu2020solving, wu2021adversarial}, loss function \cite{huang2016learning, dong2017class, zhang2017range}, and training strategy \cite{kang2021exploring, kang2019decoupling, chu2020feature} have been proposed to improve model performance when dealing with imbalanced data. GLMC \cite{du2023global} effectively improve the generalization of the backbone for {imbalanced} visual recognition by combining curated self-supervised and supervised loss functions.  RoBal \cite{wu2021adversarial} succeed in tackle adversarial robustness under {imbalanced} distribution by leveraging a scale-invariant classifier and data re-balancing via both margin engineering at training stage and boundary adjustment during inference.
Recently, studies like BALLAD \cite{ma2021simple} and PEL \cite{shi2023parameter} employed pretrained contrastive vision-language models such as CLIP to achieve state-of-the-art performance on imbalanced datasets.

\subsubsection{Object Detection}\ 

Object detection involves simultaneously estimating the categories and locations of object instances within a given image. The imbalance problems in object detection can be grouped into four main categories: class imbalance, scale imbalance, spatial imbalance and objective imbalance \cite{oksuz2020imbalance}. Class imbalance refers to the unequal distribution of examples among classes, which can occur both between foreground classes and between foreground and background (negative) classes. To handle this problem, sampling methods are often utilized which re-weights bounding boxes according to their types (i.e. positive or negative)~\cite{girshick2014rich, liu2016ssd}, the prediction hardness~\cite{ross2017focal, li2019gradient} or the IoU metric~\cite{cao2020prime}. In addition to direct sampling methods, some studies introduce additional model architectures to improve the detection criterion~\cite{chen2022residual, chen2019towards}, and some others address imbalance by producing artificial samples within the training dataset~\cite{wang2017fast,tripathi2019learning}.  Scale imbalance arises from the unequal distribution of object scales and input bounding box sizes. Solutions to these issues often involve utilizing models and features with a pyramidal structure. For example, Single Shot Detector (SSD)~\cite{liu2016ssd} is able to exploits features at different levels through a feed-forward pyramidal convolutional network. Its key approach to addressing scale imbalance is the use of multi-scale convolutional bounding box outputs attached to multiple feature maps at the top of the network. Feature Pyramid Networks (FPNs)~\cite{lin2017feature} offer a generic solution for constructing feature pyramids within deep convolutional networks. The construction includes a bottom-up pathway, a top-down pathway, and lateral connections. By integrating these three components, FPNs empower the network to generate feature representations that are both semantically rich and spatially detailed across multiple scales.

\subsubsection{Image Segmentation}\ 

The goal of image segmentation is to divide an image into multiple regions and map them into classes or objects. Segmentation models often suffer from data imbalance, as real-world image segmentation datasets typically exhibit {imbalanced} label distributions, which can lead to a bias towards the majority classes. To mitigate this issue, some studies investigate into class-balancing loss functions. Sudre et al.~\cite{sudre2017generalised} studied the use of different loss functions in deep learning frameworks for segmenting medical images, particularly when dealing with highly unbalanced data. The authors propose using the Generalized Dice overlap~\cite{crum2006generalized} as a robust loss function for such tasks. Bulo et al.~\cite{rota2017loss} introduces a novel concept called loss max-pooling for handling imbalanced training data distributions in semantic image segmentation. This approach adaptively re-weights the contributions of each pixel based on their observed losses, targeting under-performing classification results often encountered for under-represented object classes. Taghanaki et al.~\cite{taghanaki2019combo} propose a new curriculum learning based loss function named Combo loss to tackle the imbalance problem in segmenting multiple organs from medical images. The Combo loss leverage Dice similarity coefficient to deter model parameters from being held at bad local minima and gradually learn better model parameters by penalizing for false positives/negatives using a cross entropy term. Zhong et al.~\cite{zhong2023understanding} explored imbalanced semantic segmentation from the perspective of neural collapse. They introduced a regularization loss on feature centers called Center Collapse Regularizer to encourage the network to learn features closer to the appealing structure.

\subsection{Text data}
Data imbalance also causes problem for many NLP tasks~\cite{henning2022survey,zhao2024retrieval,wang2024qaencoder}, such as sentiment analysis, news classification, and text generation. It can cause the models trained on imbalanced data to be biased towards the majority classes and ignore the minority classes.

\subsubsection{Sentiment Analysis}\

Sentiment analysis is a task that aims to identify the emotional attitude of a text, such as positive, negative, neutral, etc. However, some classes may be more frequent than others in the data, such as positive reviews being more common than negative ones. This creates a problem of class imbalance, which can affect the performance of sentiment classifiers.
For example, the sentiment of tweets collected from Twitter on majority of keywords related to social events, people, organization is highly imbalanced. To alleviate bias, 
re-sampling techniques such as SMOTE are widely applied in \cite{ghosh2019imbalanced,flores2018evaluation,ardianto2020sentiment} to over-sample the minority class. 
% They found that the SMOTE method significantly improved the classifier performance as revealed by precision and recall values.
In addition to social psychology, there are several other factors that can greatly skew sentiment analysis results. One example comes from the work of Al-Azani et al. \cite{al2017using}, where they examined the sentiment analysis of the Syria Tweets dataset. This dataset consisted of 1350 negative and 448 positive tweets, as reported by Mohammad et al. \cite{mohammad2016translation}. It is important to note that this dataset was highly imbalanced due to the ongoing regional instability in Syria.
Compared with static sentiment analysis, real-time sentiment analysis is more challenging due to limited labeled data and sudden sentiment changes caused by real-world events (concept shift). To overcome this, Guerra et al. \cite{guerra2014sentiment} acquire labels and handle concept drift using insights from self-report imbalances. That is, a preference for reporting positive feelings over negative feelings and a preference for reporting extreme feelings over average feelings, which helps generate labeled data on polarizing topics and informs a spike-based feature representation strategy. 
% Their work achieves accuracy of up to 84\% in analyzing live reactions on Twitter, without manual labeling. 
Apart from sentiment analysis of social network, another example can be product reviews. In product reviews, satisfied customers may be less likely to leave a review, while dissatisfied customers are more likely to voice their opinions, leading to a higher number of negative reviews. For instance, Obiedat et al. \cite{obiedat2022sentiment} proposed PSO-SVM with particle swarm optimization (PSO) to find the best feature weights and the best $k$ value for each oversampling technique, and then applied SVM to classify the oversampled and weighted reviews.

\subsubsection{News Classification}\ 

News Classification involves assigning a topic or category to a news article, such as politics, sports, entertainment, etc. The class imbalance phenomenon is the situation where the number of news articles that belong to different categories is not equal. 
Some topics may be more popular or frequent than others, such as sports news being more common than science news. For example, Dogra et al. \cite{dogra2022comparative} aim to extract the banking news and its most correlated news articles from the pool of financial news articles, and divide financial news articles into four classes: banking, non-banking, governmental, and global. The distribution of news articles amongst four classes are 9.16\%, 70.54\%, 4.21\% and 16.09\% respectively, due to the different frequency and level of attention of news events. 
Sometimes, class imbalance phenomenon happens because the class of interest is quite rare than other classes, especially in binary classification problem. For instance, Rivera et al. \cite{rivera2020news} need to classify the news from a local newspaper based on whether they are about a traffic incident or not. However, the class of interest, which is `traffic accident’, only represents 10.30\% of the instances, while the rest are other types of news. 
Another binary classification example can be fake news classification, in this scenario, the class imbalance refers to the number of fake news data is much smaller than the number of true news data. This happens because manual labeling of fake news data is time-consuming and difficult, while true news data is more abundant and accessible. For example, the BanFakeNews dataset \cite{hossain2020banfakenews} contains about 50,000 Bengali news articles, of which only 1,299 are labeled as fake and the rest are labeled as true. Hence, Keya et al. \cite{keya2022augfake} proposed AugFake-BERT, which uses the pre-trained BERT model to insert and substitute words for data augmentation, and train a smaller BERT model from scratch with the augmented dataset for embedding and classification. For multi-class classification problem, the {imbalnced} phenomenon is also evident, for example, Hasib et al. \cite{hasib2023strategies} study Bangla news classification with Bangla Newspaper Dataset \cite{zabir_al_nazi_2020}. The sample of 437,948 news articles are organized into the high imbalanced 32 categories, where the `bangladesh' category has 232,504 amounts of article and the percentage is 53.09\% which is the highest and the lowest one are `askeditor' and `bs-events' categories which had only one article respectively.

% \subsection{Language modeling}

\subsubsection{Text Generation}\

In addition to the aforementioned categories, the growing prevalence of large language models (LLMs) has brought the issue of data imbalance in text generation tasks to the forefront. To address this challenge, researchers have proposed several strategies. One prominent approach involves adjusting the composition of pre-training datasets to balance the proportions of data across tasks, languages, or domains. This method enhances the model's learning efficiency, ensures a comprehensive representation of language features, and aligns the model's capabilities with the specific requirements of the target tasks, thereby achieving balanced performance across diverse scenarios. Raffel et al.~\cite{raffel2020exploring} demonstrated the effectiveness of two such strategies: mixing proportionally to dataset size and temperature-scaled mixing. In the former, datasets are weighted based on their sample sizes, while in the latter, a temperature parameter $T$ is introduced to adjust task-level data distribution. This parameter increases the relative weight of smaller tasks, offering a more balanced mixture. Another promising strategy is leveraging fine-tuned models with limited samples as proxies for data selection. For example, Li et al.~\cite{li2023from} proposed the ``cherry model,'' which operates through three fine-tuning stages: learning from limited experience, evaluation, and retraining based on self-guided insights. This model autonomously identifies and selects high-quality ``cherry samples'' from open-source datasets. Remarkably, on the Alpaca dataset, this method achieved superior performance using only approximately 5\% of the data compared to models trained on the entire dataset. Custom dataset construction and task-specific fine-tuning also serve as effective methods to mitigate data imbalance. For instance, Li et al.~\cite{li2024numinamath} segmented raw data into question-answer pairs, translated and realigned them, and incorporated chain-of-thought (CoT) reasoning and tool-integrated reasoning (TIR) annotations. The resulting mathematical problem dataset underwent two stages of fine-tuning: one on diverse CoT-annotated datasets and another on synthetic TIR datasets, leading to significant improvements in mathematical reasoning tasks. Furthermore, iterative trial-and-error processes have proven effective in addressing data imbalance. An et al.~\cite{an2023learning} utilized multiple LLMs to collect inaccurate reasoning paths, which were subsequently corrected by GPT-4. This Correction-Centric Evolution strategy facilitated fine-tuning, enabling consistent performance improvements across various LLMs and tasks.

% \subsubsection{Relation Extraction} Relation extraction involves identifying and classifying relationships between entities mentioned in text. It aims to extract structured information from unstructured text by determining the type of relationship between entities, such as the interaction between a person and an organization mentioned in a news article. The data imbalance phenomenon of relation extraction is also widely reported in \cite{ye2019exploiting,song2021classifier,mitra2020multi,zhang2019long,yuan2017one,xie2021revisiting,nguyen2014robust}. For example, relation extraction with distant supervision is widely adopted, which automatically generates relation labels by aligning the text with the existing knowledge base. 

% \begin{comment}
% \subsection{Topic Classification}
% \subsubsection{Neural Machine Translation}
% \subsubsection{Event type detection} This task involves determining the type of an event in a text, such as traffic accident, terrorist attack, natural disaster, etc. Some events may be more frequent or salient than others, such as social events being more common than military events.
% \end{comment}

% \subsection{Time series data}

\subsection{Graph data}
Recently, graph data~\cite{shao2024distributed,gao2023accelerating,gao2024contrastive,gao2024rethinking} has gained significant attention for its ability to represent intricate structural relationships among diverse entities, facilitating applications across fields such as chemical analysis, social network modeling, and recommendation systems~\cite{gao2024robgc,gao2023semantic,gao2024graphkdd,gao2024graph}. The extensive use of graph structures has driven considerable advancements in graph analytics, typically categorized into node-level, edge-level, and graph-level tasks based on the analytical objectives. Given these inherent objectives within graph structures, imbalance issues are prevalent across all analytical levels, manifesting in various forms. In the following, we will formulate these distinct imbalance challenges in alignment with each graph objective.

\subsubsection{Node-level imbalance}\ 

\textbf{Node class imbalance.} Node class imbalance refers to the uneven distribution of class labels among nodes within a graph, where certain classes contain fewer labels compared to others. More importantly, the non-IID nature of graph data causes nodes to influence their neighbors through the message-passing paradigm in graph neural networks, further exacerbating the under-representation of minority classes. This interconnectedness intensifies the challenges posed by imbalanced data, as classes with fewer labels struggle to propagate meaningful signals across the graph, potentially diminishing model performance on these underrepresented classes.

To supplement the limited labels of minority classes, data generation has emerged as a widely adopted approach. In linear generative methods, for example, GraphSMOTE applies the SMOTE technique within the graph domain to synthesize new nodes for minority classes and establish edges among them. Other methods, such as GraphMixup \cite{wu2022graphmixup}, GraphENS \cite{park2021graphens}, SNS \cite{gao2023semantic}, and GraphSHA \cite{li2023graphsha}, employ mixup techniques to create more challenging samples by interpolating nodes from different classes, thereby enhancing the generalization capacity of models. In contrast, deep generative methods such as ImGAGN \cite{qu2021imgagn} leverage GAN architectures, where a generator and discriminator work simultaneously to perform data generation and classification. Additionally, methods like GraphSR \cite{zhou2023graphsr} and BIM \cite{zhang2024bim} enhance minority class representation through self-training, expanding the labeled dataset by annotating unlabeled nodes within the original graph. This approach helps mitigate noise introduced by generative models, leading to improved and more robust performance.

\textbf{Node degree imbalance.}
Node degrees in graphs often follow a {imbalanced} distribution, i.e., a significant fraction of nodes are {minority} nodes with
a small degree. For instance, in social networks, a small number of users may have thousands of connections, while the majority maintain only a few. Similarly, in e-commerce recommendation systems, popular items frequently appear in user histories, whereas niche items receive minimal attention. This disparity, referred to as node degree imbalance, creates challenges in effectively exploring the information from low-degree {minority} nodes. Specifically, {majority} nodes with high degrees benefit from rich structural information and neighbors, allowing them to achieve better representations and stronger performance in downstream tasks. In contrast, "{minority}" nodes with low degrees have limited topological information, which constrains their effectiveness.

To mitigate the information scarcity in {minority} nodes, researchers have developed methods to transfer structural patterns from {majority} nodes to {minority} nodes. LTE4G \cite{yun2022lte4g} trains multiple expert models to handle different types of nodes, using knowledge distillation to transfer knowledge from {majority}-node experts to those focused on {minority} nodes. Similarly, Cold Brew leverages a strong teacher model with node-wise structural embeddings to guide a student model, enabling it to uncover missing latent neighborhoods for {minority} nodes. Tail-GNN \cite{liu2021tail} models variable connections between a target node and its neighbors and learns the neighborhood translation from {majority} nodes to enhance the structurally limited {minority} nodes. SAILOR proposed a structural augmentation method to enhance node homophily by adding pseudo-homophilic edges to {minority} nodes, allowing them to acquire more task-relevant information from neighboring nodes.

\textbf{Node topology imbalance.}
Distinct from other imbalance issues that focus on disparities in node or degree quantity, node topology imbalance highlights the positional distribution of labeled nodes within the graph. This problem is rooted in the homophily assumption that influence boundaries created through label propagation should ideally align with true class boundaries. However, the actual placement of labeled nodes often shifts these influence boundaries away from true boundaries, leading to skewed model decision boundaries and degraded performance.

To address topology imbalance, ReNode \cite{chen2021topology} introduces a metric that quantifies this imbalance by measuring influence conflict. It then reweights labeled nodes based on their proximity to class boundaries, thereby improving the performance for nodes located near boundaries. Alternatively, PASTEL \cite{sun2022position} proposes an anchor-based position encoding mechanism that directly optimizes information propagation paths, enhancing the connectivity among nodes within the same class to strengthen the flow of supervisory information.

\subsubsection{Edge-level imbalance}\ 

Besides the node-level imbalances, real-world graphs also exhibit imbalanced edge distributions, which significantly impact various downstream tasks, including inductive recommendation and anomalous transaction detection on financial networks. Specifically, new users or items are continuously added to the system as novel inductive nodes by linking with the existing graph. However, these novel nodes, situated within a few-shot context, typically possess limited edges due to scarce information, posing challenges for effective node representation and accurate link prediction. To address this issue, a recent study \cite{zhu2023few} employs episodic learning to explore transferable knowledge and augment the representation of nodes with sparse connections. Additionally, imbalanced edge distribution prominently affects edge anomaly detection tasks, where anomalies such as abnormal transactions are modeled as minorities within the graph. This challenge is generally addressed by characterizing the normal edge distributions \cite{zhu2023few} with auxiliary information \cite{liu2022deep} and identifying deviations from these norms as indicators of anomalies \cite{li2021live}.

\subsubsection{Graph-level imbalance}\ 

Graph-level datasets, frequently employed in modeling molecular or drug graphs, inherently exhibit an imbalanced distribution. This imbalance is primarily observed in datasets featuring naturally occurring distributions, such as those associated with rare diseases or specific chemical compounds. This scenario poses significant challenges for machine learning models, which may struggle to accurately predict properties or classifications for these infrequent but critical instances. To address these challenges, recent research \cite{wang2022imbalanced} constructs a global graph structure that aggregates information among similar samples to enhance representation. This technique effectively boosts the predictive capabilities of models dealing with rare or infrequent graph types by pooling collective information from analogous instances. Additionally, another study \cite{liu2023semi} focuses on regression tasks and employs additional real, yet unlabeled data to compensate for minorities. By adopting an iterative training process, this model gradually annotates the unlabeled data and includes these data for model training, thereby enhancing model performance systematically over time.

%% file: 2.Evaluation.tex
\section{Evaluation Metrics}
\label{sec:eval}

Evaluating models for imbalanced datasets requires careful consideration of the metrics used to assess the performance. Traditional metrics like accuracy can not provide a reliable measure in this imbalanced contexts because they are biased toward the majority. In datasets where one class significantly outnumbers another, a model can achieve high accuracy simply by favoring the majority, often at the expense of poorly predicting the minority. For example, in a medical diagnosis setting where 95\% of the samples are negative for a disease, a model that always predicts ``negative'' will achieve 95\% accuracy, despite failing to identify any positive cases, which are often the most critical to detect. To counter this bias and accurately assess model effectiveness, it's crucial to use evaluation metrics that accurately reflect performance across all classes, especially when misclassification costs are high. Below, we first describe the fundamental metrics for classification and then introduce the metrics for imbalance scenarios:

\noindent\textbf{{Confusion Matrix}} is a fundamental tool that provides a detailed breakdown of classification results across the actual and predicted classes, and the prediction results are categories into 4 types:

\begin{itemize}
    \item True Positives (TP): Correctly predicted positive observations.
    \item True Negatives (TN): Correctly predicted negative observations.
    \item False Positives (FP): Incorrectly predicted as positive.
    \item False Negatives (FN): Incorrectly predicted as negative.
\end{itemize}

\noindent\textbf{Precision} provides insights into the accuracy of positive predictions.
\begin{equation}
\text{Precision} = \frac{\text{TP}}{\text{TP} + \text{FP}}.
\end{equation}

\noindent\textbf{Recall}, also known as \textbf{Sensitivity}, measures the proportion of actual positives that are correctly identified by the model.
\begin{equation}
\text{Recall} = \frac{\text{TP}}{\text{TP} + \text{FN}}.
\end{equation}
It indicates how effective the model is at detecting positive instances from the data set.

\noindent\textbf{Specificity} measures the proportion of actual negatives that are correctly identified as such. In other words, it is the ability of the test to identify negative results.
\begin{equation}
\text{Specificity} = \frac{\text{TN}}{\text{TN} + \text{FP}}.
\end{equation}

With these fundamental metrics, we can derive more advanced measures to enhance the evaluation of imbalanced datasets:

\noindent\textbf{F1-score} combines precision and recall metrics to balance their influence:
\begin{equation}
\text{F1} = 2 \times \frac{\text{Precision} \times \text{Recall}}{\text{Precision} + \text{Recall}}.
\end{equation}
Precision and recall often have a trade-off relationship: improving precision typically reduces recall and vice versa. This occurs because increasing the threshold for classifying positives can decrease the number of false positives (increasing precision) but also miss more actual positives (decreasing recall). By combining these two metrics, the F1-score offers a single metric that balances this trade-off, providing a holistic view of the model performance.

In multi-class classification scenarios, the F1 score can be calculated in distinct ways, specifically through \textbf{micro} and \textbf{macro} averaging. Micro combines the outcomes of all classes to compute a single average metric, which is particularly useful in imbalanced datasets, as it emphasizes performance on {minority classes}. Conversely, macro calculates the metric separately for each class and then averages these results, ensuring that all classes are given equal importance regardless of their size in the dataset.

\noindent\textbf{Balanced accuracy (BA)} adjusts accuracy to reflect equitable performance across classes:
\begin{equation}
\text{BA} = \frac{\text{Sensitivity} + \text{Specificity}}{2}.
\end{equation}

\noindent\noindent\textbf{G-Mean} is defined as the geometric mean of sensitivity and specificity, and is sensitive to the performance on both classes:
\begin{equation}
\text{G-Mean} = \sqrt{\text{Sensitivity} \times \text{Specificity}}.
\end{equation}

\noindent\textbf{Matthews Correlation Coefficient (MCC)} considers all four quadrants of the confusion matrix and is highly effective for binary classification problems:
\small{
\begin{equation}
\text{MCC} = \frac{\text{TP} \times \text{TN} - \text{FP} \times \text{FN}}{\sqrt{(\text{TP}+\text{FP})(\text{TP}+\text{FN})(\text{TN}+\text{FP})(\text{TN}+\text{FN})}}.
\end{equation}}

\noindent\textbf{Area Under the Receiver Operating Characteristic Curve (AUC-ROC)} is a performance measurement for classification models at various threshold settings. It represents the probability that a classifier will rank a randomly chosen positive instance higher than a randomly chosen negative instance. The ROC curve plots the True Positive Rate (TPR, also known as recall or sensitivity) against the False Positive Rate (FPR) at different threshold levels.
\begin{equation}
\text{AUC-ROC} = \int_{0}^{1} \text{TPR}(\tau) \, d\text{FPR}(\tau).
\end{equation}
A model with perfect predictive accuracy would have an AUC-ROC of 1, indicating it perfectly ranks all positive instances higher than negative ones.

These metrics provide detailed and varied perspectives for assessing the performance of models on imbalanced datasets, enabling the selection of the appropriate measures based on specific evaluation needs.

%% file: 9Exp.tex
\section{Experiments}
\label{sec:expe}
{
To assess the effectiveness of these approaches, we evaluate 20 representative methods across seven benchmark datasets, which vary in both size and imbalance ratios. The dataset statistics are summarized in Table~\ref{tab_data}.}

{
We begin by evaluating 12 conventional machine learning methods on five widely used imbalanced learning benchmarks. Specifically, we compare
\begin{itemize}
\item 3 over-sampling methods (SMOTE~\cite{ref_2}, BorderlineSMOTE~\cite{ref_4}, and ADASYN~\cite{ref_5});
\item 4 under-sampling methods (TomekLinks~\cite{kubat1997addressing}, OneSidedSelection~\cite{kubat1997addressing}, Condensed Nearest Neighbors~\cite{ref_6}, and Edited Nearest Neighbors~\cite{ref_17}); 
\item 2 hybrid sampling methods (SMOTEENN~\cite{batista2004study} and SMOTETomek~\cite{batista2004study}); 
\item 3 ensemble methods (Balanced Bagging~\cite{hido2009roughly}, RUSBoost~\cite{ref_20}, and Balanced Random Forest~\cite{khoshgoftaar2007empirical}). 
\end{itemize}
}

\begin{table}[t]
\setlength{\abovecaptionskip}{1pt}
\renewcommand{\arraystretch}{1.4}
\center
\caption{{The statistic of datasets.}}
\label{tab_data}
\setlength{\tabcolsep}{3.5pt}
\resizebox{0.9\linewidth}{!}{
\begin{tabular}{l|cccc}
\Xhline{1.pt}
Datasets        & \begin{tabular}[c]{@{}c@{}}Imbalance \\ Ratio\end{tabular} & \#Samples & \#Features & \#Classes \\ \hline
ecoli           & 8.6:1                                                      & 336       & 7          & 2         \\
optical\_digits & 9.1:1                                                      & 5,620     & 64         & 2         \\
pen\_digits     & 9.4:1                                                      & 10,992    & 16         & 2         \\
car\_eval\_34   & 12:1                                                      & 1,728     & 21         & 2         \\
letter\_img     & 26:1                                                       & 20,000    & 16         & 2         \\
MINIST          & 10:01                                                      & 33,000    & 784        & 10        \\
FashionMNIST    & 10:01                                                      & 33,000    & 784        & 10        \\
\Xhline{1.pt}
\end{tabular}}
\end{table}

{
For a fair comparison, we use Random Forest as the base classifier, and evaluate performance using three metrics: F1 score, AUC, and Accuracy (ACC). Each experiment is repeated 10 times, and the average performance, standard deviation, and rank of each method are reported in Table~\ref{tab_result1}.}

{
Compared to the base model, all methods generally achieve improved F1 scores, indicating enhanced minority class performance. However, some methods reduce overall accuracy, suggesting diminished performance on the majority class.
Among all methods, SMOTE and SMOTEENN deliver the most robust and consistent results, highlighting the effectiveness of synthetic data generation in compensating for minority class under-representation and improving classifier robustness. In contrast, while under-sampling methods can also enhance performance, they often discard a substantial portion of the majority class data, inevitably leading to information loss. It is important to note that no method consistently outperforms others across all datasets, highlighting the need to select appropriate strategies based on dataset-specific characteristics.}

{
Besides the conventional machine learning methods, we evaluate deep learning-based methods on the widely used benchmarks MNIST and Fashion-MNIST. Following the setup in [1], we impose a 10:1 imbalance ratio to assess the models. The encoder is implemented as a convolutional neural network, followed by a MLP classifier. According to our taxonomy, {we compare 10 kinds of methods including:} 
\begin{itemize}
\item {Linear generation (SMOTE~\cite{ref_2}, ADASYN~\cite{ref_5});}
\item Deep generation (GAN~\cite{mullick2019generative});
\item Cost-sensitive learning (Focal loss~\cite{lin2017focal});
\item Metric learning (Triplet loss~\cite{weinberger2009distance});
\item Supervised contrastive learning (SupCon~\cite{khosla2020supervised});
\item Prototype learning (PSC~\cite{wang2021contrastive});
\item Transfer learning (FTL~\cite{yin2019feature});
\item Decoupling training (DT~\cite{kang2019decoupling});
\item Posterior recalibration ($\tau$-norm~\cite{kang2019decoupling}).
\end{itemize}
}

{
As shown in Table \ref{tab_result3}, Transfer learning and GAN-based data generation achieve the best overall performance, as they can generate additional informative samples by leveraging the majority class distribution. 
However, despite their effectiveness, GANs entail a significantly higher computational cost due to the complexity of training both the generator and discriminator networks. 
As shown in Table~\ref{tab_result2}, training a GAN takes approximately 15 minutes, while the conventional method SMOTE completes in just 2 seconds. Therefore, the choice of data generation strategy should balance performance and computational efficiency, depending on the specific application and deployment constraints.
Notably, decoupled training demonstrates competitive performance with minimal computational overhead. By simply fine-tuning the MLP classifier on a balanced dataset, it emphasizes the impact of training strategies on model effectiveness, offering a highly efficient alternative for imbalanced learning.
}

\begin{table*}[t]
    \setlength{\abovecaptionskip}{1pt}
    \renewcommand{\arraystretch}{1.4}
    \center
    \caption{{The comparison of conventional machine learning methods, with the best-performing results highlighted in bold.}}
    \label{tab_result1}
    \setlength{\tabcolsep}{3.5pt}
    \resizebox{\textwidth}{!}{
\begin{tabular}{l|l|ccc|ccc|ccc|ccc|ccc|c}
\Xhline{1.pt}
\multirow{3}{*}{Category}     & Dataset              & \multicolumn{3}{c|}{ecoli}                                       & \multicolumn{3}{c|}{optical\_digits}                            & \multicolumn{3}{c|}{pen\_digits}                                & \multicolumn{3}{c|}{car\_eval\_34}                              & \multicolumn{3}{c|}{letter\_img}                                & \multicolumn{1}{c}{\multirow{2}{*}{Rank}} \\ \cline{2-17}
                              & Method               & F1                   & AUC                 & ACC                 & F1                  & AUC                 & ACC                 & F1                  & AUC                 & ACC                 & F1                  & AUC                 & ACC                 & F1                  & AUC                 & ACC                 & \multicolumn{1}{c}{}                      \\ \cline{2-18} 
                              & Base                 & 59.31\tiny{±10.60}          & 73.60\tiny{±5.31}          & 92.77\tiny{±1.60}          & 88.85\tiny{±1.97}          & 90.00\tiny{±1.58}          & 98.03\tiny{±0.31}          & 96.92\tiny{±0.48}          & 97.99\tiny{±0.43}          & 99.61\tiny{±0.09}          & 86.77\tiny{±4.50}          & 91.27\tiny{±3.20}          & 98.05\tiny{±0.65}          & 94.90\tiny{±1.01}          & 95.59\tiny{±1.06}          & 99.64\tiny{±0.07}          & \multicolumn{1}{l}{N.A.}                  \\ \hline
\multirow{3}{*}{Over-sampling} & SMOTE                & 60.91\tiny{±8.47}           & 79.10\tiny{±5.85}          & 91.19\tiny{±1.90}          & 94.23\tiny{±1.12}          & 94.63\tiny{±1.01}          & 98.91\tiny{±0.20}          & 98.56\tiny{±0.53}          & 98.71\tiny{±0.47}          & 99.73\tiny{±0.10}          & 89.88\tiny{±2.60}          & 94.69\tiny{±2.51}          & 98.44\tiny{±0.39}          & 95.56\tiny{±1.06}          & 96.14\tiny{±1.06}          & 99.69\tiny{±0.07}          & 4.47                                      \\
                              & BorderlineSMOTE      & 60.73\tiny{±11.97}          & 74.40\tiny{±8.33}          & 91.21\tiny{±2.14}          & 90.64\tiny{±1.84}          & 91.66\tiny{±1.50}          & 98.31\tiny{±0.31}          & 97.84\tiny{±0.64}          & 98.01\tiny{±0.57}          & 99.59\tiny{±0.12}          & 89.79\tiny{±2.07}          & 94.79\tiny{±2.21}          & 98.42\tiny{±0.30}          & 95.63\tiny{±0.95}          & 96.30\tiny{±0.88}          & 99.69\tiny{±0.07}          & 6.80                                       \\
                              & ADASYN               & 56.56\tiny{±9.07}           & 76.83\tiny{±6.16}          & 90.00\tiny{±2.28}          & 92.76\tiny{±1.35}          & 93.49\tiny{±1.10}          & 98.67\tiny{±0.24}          & 98.28\tiny{±0.70}          & 98.41\tiny{±0.68}          & 99.68\tiny{±0.13}          & \textbf{90.27\tiny{±3.00}} & 94.95\tiny{±2.57}          & \textbf{98.50\tiny{±0.45}} & 95.60\tiny{±1.20}          & 96.51\tiny{±1.10}          & 99.69\tiny{±0.08}          & 5.47                                      \\ \hline
\multirow{4}{*}{Down-sampling} & TomekLinks           & \textbf{67.97\tiny{±10.09}} & 81.12\tiny{±6.46}          & \textbf{93.37\tiny{±2.08}} & 89.43\tiny{±1.26}          & 90.45\tiny{±1.03}          & 98.12\tiny{±0.20}          & 97.97\tiny{±0.48}          & 98.04\tiny{±0.42}          & 99.62\tiny{±0.09}          & 87.56\tiny{±3.56}          & 91.79\tiny{±2.91}          & 98.17\tiny{±0.50}          & 95.00\tiny{±1.06}          & 95.62\tiny{±1.03}          & 99.65\tiny{±0.07}          & 7.67                                      \\
                              & OneSidedSelection  & 67.16\tiny{±9.69}           & 81.06\tiny{±7.37}          & 93.27\tiny{±1.58}          & 88.69\tiny{±1.18}          & 89.85\tiny{±0.95}          & 98.00\tiny{±0.19}          & 97.91\tiny{±0.45}          & 97.99\tiny{±0.43}          & 99.61\tiny{±0.08}          & 87.02\tiny{±3.16}          & 91.41\tiny{±2.76}          & 98.09\tiny{±0.43}          & 95.09\tiny{±1.31}          & 95.73\tiny{±1.21}          & 99.66\tiny{±0.09}          & 8.80                                       \\
                              & CondensedNN          & 61.51\tiny{±10.17}          & 78.58\tiny{±7.00}          & 91.68\tiny{±2.27}          & 91.56\tiny{±1.52}          & 97.73\tiny{±0.43}          & 98.23\tiny{±0.35}          & 97.46\tiny{±1.66}          & 98.81\tiny{±0.41}          & 98.44\tiny{±0.36}          & 88.40\tiny{±2.40}          & 95.55\tiny{±2.77}          & 98.13\tiny{±0.40}          & 95.20\tiny{±0.99}          & 97.98\tiny{±0.87}          & 99.64\tiny{±0.07}          & 6.53                                      \\
                              & EditedNN             & 66.11\tiny{±7.64}           & 85.14\tiny{±5.94}          & 91.29\tiny{±2.61}          & 89.76\tiny{±1.57}          & 90.75\tiny{±1.32}          & 98.17\tiny{±0.25}          & 97.94\tiny{±0.52}          & 98.01\tiny{±0.46}          & 99.61\tiny{±0.10}          & 87.56\tiny{±4.87}          & 94.47\tiny{±2.93}          & 97.82\tiny{±0.82}          & 94.92\tiny{±1.08}          & 95.75\tiny{±1.17}          & 99.64\tiny{±0.07}          & 8.07                                      \\ \hline
\multirow{2}{*}{Hybrid sampling}       & SMOTEENN             & 61.65\tiny{±6.33}           & 86.09\tiny{±5.30}          & 88.71\tiny{±2.66}          & \textbf{94.28\tiny{±1.10}} & \textbf{94.71\tiny{±0.97}} & \textbf{98.93\tiny{±0.20}} & 98.56\tiny{±0.66}          & 98.72\tiny{±0.60}          & 99.73\tiny{±0.12}          & 88.14\tiny{±3.54}          & 97.31\tiny{±2.34}          & 98.00\tiny{±0.60}          & 95.40\tiny{±1.19}          & 96.48\tiny{±1.06}          & 99.67\tiny{±0.08}          & 4.47                                      \\
                              & SMOTETomek           & 60.59\tiny{±9.66}           & 79.78\tiny{±7.73}          & 90.99\tiny{±1.74}          & 93.53\tiny{±1.13}          & 94.02\tiny{±0.97}          & 98.80\tiny{±0.20}          & 98.46\tiny{±0.50}          & 98.60\tiny{±0.45}          & 99.71\tiny{±0.09}          & 89.72\tiny{±2.83}          & 94.45\tiny{±2.56}          & 98.42\tiny{±0.42}          & \textbf{95.64\tiny{±1.00}} & 96.23\tiny{±1.00}          & \textbf{99.70\tiny{±0.07}} & 5.33                                      \\ \hline
\multirow{3}{*}{Ensemble}     & BalancedBagging      & 61.46\tiny{±7.59}           & 87.01\tiny{±5.99}          & 88.22\tiny{±3.18}          & 91.52\tiny{±1.30}          & 96.42\tiny{±0.68}          & 98.28\tiny{±0.28}          & 97.59\tiny{±0.83}          & 98.77\tiny{±0.50}          & 99.54\tiny{±0.16}          & 87.20\tiny{±2.90}          & \textbf{98.54\tiny{±0.33}} & 97.30\tiny{±0.61}          & 95.37\tiny{±2.55}          & 98.25\tiny{±0.55}          & 98.23\tiny{±0.27}          & 7.00                                         \\
                              & RUSBoost             & 62.18\tiny{±4.28}           & 84.82\tiny{±4.92}          & 89.31\tiny{±3.22}          & 91.62\tiny{±1.87}          & 92.66\tiny{±1.13}          & 98.89\tiny{±0.53}          & 97.05\tiny{±1.94}          & 98.95\tiny{±0.91}          & 96.79\tiny{±0.46}          & 87.11\tiny{±4.32}          & 97.41\tiny{±2.20}          & 97.34\tiny{±0.84}          & 95.27\tiny{±4.57}          & 96.08\tiny{±1.41}          & 96.99\tiny{±0.73}          & 7.73                                      \\
                              & BalancedRandomForest & 62.84\tiny{±4.28}           & \textbf{89.00\tiny{±2.76}} & 88.22\tiny{±2.75}          & 92.86\tiny{±1.51}          & 96.53\tiny{±0.86}          & 98.58\tiny{±0.30}          & \textbf{98.70\tiny{±0.59}} & \textbf{99.04\tiny{±0.49}} & \textbf{99.75\tiny{±0.11}} & 88.09\tiny{±2.64}          & 97.67\tiny{±0.50}          & 95.92\tiny{±0.63}          & 95.37\tiny{±1.81}          & \textbf{98.53\tiny{±0.72}} & 98.86\tiny{±0.17}          & 4.60                                       \\ \Xhline{1.pt}
\end{tabular}}
\end{table*}

\begin{table}[t]
    \setlength{\abovecaptionskip}{1pt}
    \renewcommand{\arraystretch}{1.4}
    \center
    \caption{{The performance and data augmentation time comparison between representative methods.}}
    \label{tab_result2}
    \setlength{\tabcolsep}{3.5pt}
    \resizebox{0.8\linewidth}{!}{
\begin{tabular}{l|cc|cc}
\Xhline{1.pt}
Dataset & \multicolumn{2}{c|}{MINIST} & \multicolumn{2}{c}{FashionMNIST} \\ \hline
Method  & F1 (\%)     & Aug\_Time (s) & F1 (\%)       & Aug\_Time (s)    \\ \hline
Base    & 90.53±0.14  & N/A           & 74.63±0.10    & N/A              \\
{Focal loss } & { 91.35±0.13  } & { N/A        } & { 79.35±0.17    } & { N/A }              \\
{ADASYN  } & { 91.95+0.55  } & { 2.12          } & { 80.35+0.28     } & { 2.25 }             \\
SMOTE   & 91.18±0.08  & 1.27          & 79.61±0.41    & 1.31              \\
GAN     & 93.46±0.21  & 920.41        & 80.41±0.52    & 933.86           \\ \Xhline{1.pt}
\end{tabular}}
\end{table}

\begin{table}[t]
\setlength{\abovecaptionskip}{1pt}
\renewcommand{\arraystretch}{1.4}
\center
\caption{{The comparison of deep learning-based methods.}}
\label{tab_result3}
\setlength{\tabcolsep}{3.5pt}
\resizebox{\linewidth}{!}{
\begin{tabular}{l|ccc|ccc|c}
\Xhline{1.pt}
Dataset                         & \multicolumn{3}{c|}{MINIST}          & \multicolumn{3}{c|}{FashionMNIST}    & \multirow{2}{*}{Rank} \\ \cline{1-7}
Method                          & F1         & AUC        & ACC        & F1         & AUC        & ACC        &                       \\ \hline
Base                            & 90.53\tiny{±0.14} & 94.35\tiny{±0.07} & 90.63\tiny{±0.12} & 74.63\tiny{±0.10} & 87.85\tiny{±0.05} & 78.13\tiny{±0.09} & N/A                   \\

{SMOTE} & { 91.18\tiny{±0.08} } & { 95.17\tiny{±0.05} } & { 91.30\tiny{±0.08} } & { 79.61\tiny{±0.41} } & { 88.87\tiny{±0.27} } & { 79.97\tiny{±0.49}} & {8.33}                  \\

{ADASYN}                         & {91.95\tiny{±0.55} } & { 95.57\tiny{±0.30} } & { 92.03\tiny{±0.54} } & { 80.35\tiny{±0.28} } & { 89.04\tiny{±0.09} } & { 80.07\tiny{±0.17} } & { 5.67 }                \\

GAN                             & 93.46\tiny{±0.21} & 96.11\tiny{±0.23} & 92.92\tiny{±0.33} & 80.41\tiny{±0.52} & 89.48\tiny{±0.34} & 81.32\tiny{±0.41} & {2.50 }                 \\
Focal loss                      & 91.35\tiny{±0.13} & 95.24\tiny{±0.07} & 91.43\tiny{±0.12} & 79.35\tiny{±0.17} & 88.81\tiny{±0.11} & 79.87\tiny{±0.21} & {8.33}                  \\
Triplet loss                    & 91.47\tiny{±0.29} & 95.56\tiny{±0.11} & 91.50\tiny{±0.12} & 79.06\tiny{±0.59} & 88.22\tiny{±0.65} & 79.80\tiny{±0.65} & {8.50}                 \\
SupCon & 92.96\tiny{±0.60} & 95.58\tiny{±0.11} & 92.70\tiny{±0.10} & 76.37\tiny{±0.22} & 88.89\tiny{±0.13} & 80.00\tiny{±0.22} & {6.17     }             \\
PSC              & 92.43\tiny{±0.15} & 95.85\tiny{±0.10} & 92.53\tiny{±0.19} & 79.86\tiny{±0.66} & 90.22\tiny{±0.39} & 82.40\tiny{±0.71} & {4.33 }                 \\
FTL               & 94.08\tiny{±0.27} & 96.74\tiny{±0.15} & 94.13\tiny{±0.26} & 80.16\tiny{±0.74} & 90.37\tiny{±0.21} & 82.67\tiny{±0.38} & {1.50}                  \\
DT             & 92.82\tiny{±0.70} & 96.04\tiny{±0.38} & 92.87\tiny{±0.68} & 80.40\tiny{±0.26} & 89.22\tiny{±0.18} & 80.60\tiny{±0.33} & {3.83}                  \\
$\tau$-norm         & 92.97\tiny{±0.11} & 96.13\tiny{±0.07} & 93.03\tiny{±0.12} & 78.09\tiny{±0.62} & 88.17\tiny{±0.49} & 78.70\tiny{±0.88} & {5.83}                   \\  
\Xhline{1.pt}
\end{tabular}}
\end{table}

%% file: 6.Packages.tex
\section{Resources}
\label{sec:packages}

\input{table}

With the growing attention on imbalanced data, as shown in Table 1, researchers have developed an expanding array of tools to address the challenges it presents. 
Imbalanced-learn \cite{JMLR:v18:16-365} provides a Python package incorporating a broad set of classical techniques and ensuring full compatibility with scikit-learn, thereby offering a convenient framework for assessing the effectiveness of various methods.
Santos \cite{santos_imbalance_overlap} spearheaded a project summarizing many open-source contributions in this domain, including benchmark datasets, data generation and visualization tools, class overlap-based methods, data complexity measures, and software designed for learning from imbalanced data. However, some tools remain either insufficiently detailed or absent from the summary. For benchmarks, Qin et al. \cite{qin2024iglbench} introduced IGL-Bench, the first comprehensive benchmark dedicated to imbalanced graph learning. IGL-Bench integrates 16 diverse datasets, such as Cora and PubMed, along with 24 state-of-the-art algorithms, addressing both class and topological imbalances in node-level and graph-level tasks. It ensures fair comparisons by employing unified data segmentation rules, standardized imbalance ratio definitions, and fixed splits for training, validation, and testing. Furthermore, it incorporates multi-dimensional evaluation metrics, such as accuracy, balanced accuracy, macro F1, and AUC, providing a rigorous framework for assessing algorithm performance.In terms of software, several notable tools were not included in Santos’ project. One example is \textit{imbalanced-ensemble}, developed by Liu et al. \cite{liu2021imbens}. This open-source Python toolbox employs undersampling techniques in algorithms like BalancedRandomForest and UnderBagging, which randomly sample subsets from the majority class to train individual base classifiers. It also leverages SMOTE in methods such as SmoteBoost and SmoteBagging to synthesize {minority samples}. Additionally, the toolbox uses cost-sensitive approaches to adjust learning objectives, prioritizing minority class performance. The package includes visualization tools, such as Confusion Matrix Heatmaps, to facilitate intuitive performance analysis.Another noteworthy contribution is \textit{Balance}, developed by Sarig et al. \cite{sarig2023balance}. This Python toolkit adjusts and evaluates biased sample data using weighted approaches to align distributions with the target population. Its workflow includes diagnosing initial biases, generating weights via methods like propensity scores, and evaluating the impact of these weights on bias reduction and variance inflation. Balance also supports interactive visualization and diagnostic metrics, such as ASMD, offering researchers a streamlined and flexible API for bias adjustment and analysis.

%% file: table.tex
\begin{table*}[htbp]
\centering
\caption{Overview of Open Source Tools}
\label{tab:open_source_tools}
\renewcommand{\arraystretch}{1.5}  % Adjust row height for better vertical alignment
\resizebox{\linewidth}{!}{
\begin{tabular}{p{4cm}p{5cm}p{4cm}p{5cm}}
\toprule
\textbf{Project Name} & \textbf{Content} & \textbf{Usage} & \textbf{Important Information Notes} \\
\midrule
imbalanced-learn \cite{JMLR:v18:16-365} 
& Benchmark datasets, metrics, data generation, ensemble.
& Python package offering classic methods.
& Compatible with scikit-learn. \\

Open Source Contributions in Learning from Imbalanced and Overlapped Data \cite{santos_imbalance_overlap} 
& Benchmark datasets, data generation, visualization, etc.
& Study class imbalance and class overlap problems
& Focus on unbalanced data distribution and feature space overlap \\
\addlinespace
IGL-Bench \cite{qin2024iglbench}
& First comprehensive benchmark for imbalanced graph learning.
& Evaluate graph neural networks under class and topology imbalance
& Integrates 16 datasets and 24 algorithms, used in real-world applications like social networks, e-commerce, and citation networks \\
\addlinespace
Imbalanced-ensemble \cite{liu2021imbens}
& Python toolbox offering ensemble learning solutions for class imbalance
& Provides 14 ensemble learning algorithms
& Includes undersampling and SMOTE-based techniques, as well as cost-sensitive learning and visualization tools \\
\addlinespace
Balance \cite{sarig2023balance}
& Python toolkit for adjusting sample distributions using weighting
& Analyzes and adjusts sample data bias and evaluates its impact on model performance
& Offers interactive visualization, supports ASMD metrics, low computational complexity, and flexible API design for quick integration \\
\bottomrule
\end{tabular}
}
\end{table*}

%% file: 7.Challenges.tex
\section{Challenges and Future Directions}
\label{sec:challenges}

\subsection{Imbalance in Open World Setting}

Existing researches presume that all classes present in the test set are also included in the training set. However, real-world scenarios frequently adhere to an open-world setting, where new classes may emerge exclusively in the test set. 
{In such scenarios, the imbalance problem becomes exacerbated, as minority classes are not only underrepresented but potentially unavailable during training, leaving models ill-equipped to detect them.
Recent work \cite{liu2019large} emphasizes the need for adaptive mechanisms that can identify and respond to unknown classes. However, two main technical bottlenecks remain:
\textbf{(1) Identifying Unseen Classes}: Models must adapt to the dataset with no labels, which introduces zero-shot learning challenges under imbalance constraints.
\textbf{(2) Overwhelming Confidence in Seen Classes}: Imbalanced training induces a strong bias toward seen (and especially majority) classes. This causes models to misclassify unfamiliar or low-density samples with high confidence into dominant categories, effectively suppressing the emergence of new minority classes.
Future research should focus on developing uncertainty-aware learning frameworks that can detect and incorporate emerging minority classes. Promising directions include open-set recognition combined with continual learning, and the use of generative models to simulate novel class distributions during training.}

\subsection{Imbalanced Multi-Label Problem}

Imbalanced multi-label classification presents unique challenges, differing fundamentally from traditional multi-class scenarios due to the presence of multiple labels per instance. 

{This complexity is especially evident in real-world applications \cite{du2024semi} such as image tagging or medical diagnosis, and presents unique challenges:
(1) Label Overlap Across Imbalance Spectrum: A single instance may be associated with both majority and minority labels, making it difficult for the model to focus on underrepresented labels without being dominated by the gradients of frequent ones. This undermines conventional re-balancing strategies which treat labels independently.
(2) Semantic Entanglement Across Labels: When labels correspond to semantically diverse or even conflicting concepts, multi-label samples introduce ambiguous supervision, making it harder for models to learn clear decision boundaries for each label.}

Potential solutions could include the implementation of refined loss functions tailored to balance label importance, label-specific re-sampling strategies to adjust the representation of underrepresented labels, or advanced regularization techniques to prevent model overfitting on prevalent labels. Moreover, leveraging insights from the existing researches on multi-label learning could provide innovative pathways for algorithmic enhancement, facilitating more accurate predictions across the diverse label distributions in real-world datasets.

\subsection{Imbalanced Regression Problem}

In contrast to classification, regression involves continuous labels and is essential in various real-world settings, such as financial forecasting, climate modeling, and healthcare analytics. However, the imbalance in regression tasks is not as extensively studied as in classification tasks \cite{gong2022ranksim,yang2021delving,gong2022ranksim}. 
{Due to the continuous labels, two challenges are identified:
(1) Undefined Minority Regions: Without discrete classes, it is difficult to explicitly define which target ranges are underrepresented. This lack of clear minority/majority boundaries complicates the design of data-balancing or cost-sensitive techniques.
(2) Ordered Target Dependency: Regression labels are not independent categories but exist on an ordered scale. This structure demands methods that preserve relative distances while addressing imbalance.
Future research should focus on structure-aware solutions that respect label continuity while correcting distribution skew. Promising directions include ranking-based objectives, distribution-aware sampling, and ordinal-regularized loss functions that emphasize rare but critical value regions.}

\subsection{Multi-Modality Imbalance Problem}

{Multi-modality data is fundamental in applications such as healthcare diagnostics, autonomous driving, and multimedia analysis, where fusing diverse sources (e.g., text, image, audio, and sensor signals) enhances model robustness. However, imbalances inherent in multi-modal data \cite{qian2022co} introduce unique challenges not present in unimodal settings. Key challenges include:
(1) Cross-Modality Imbalance: Modalities often differ in data volume, quality, or annotation availability. For example, medical datasets may contain thousands of imaging records but only limited textual notes, causing models to over-rely on the dominant modality and neglect informative signals from the sparse ones.
(2) Fusion-Level Imbalance Misalignment: When modalities are imbalanced, fusion layers may fail to integrate them effectively, especially if one modality dominates the representation space. This leads to suboptimal learning where underrepresented modalities are either ignored or incorrectly weighted during joint decision-making.
Future research should develop imbalance-aware fusion strategies and cross-modal augmentation techniques that explicitly address representational disparities. Transfer learning and cross-modal distillation from rich to sparse modalities also hold promise for improving robustness under modality imbalance.}

\subsection{{Large Language Models for Imbalance Problem}}

{
Large language models (LLMs) provide a powerful foundation for advancing imbalanced learning because of their generative capacity, cross-domain adaptability, and ability to leverage knowledge from large-scale pretraining corpora.
Unlike conventional resampling or augmentation methods, LLMs can generate class-conditional, semantically coherent, and label-faithful data, which makes them particularly suitable for addressing the scarcity of minority classes. 
First, LLM-based augmentation through prompt design and in-context learning can synthesize minority samples across different modalities such as tabular and textual data \cite{kim2024epic,yang2024language,zhao2024improving}. This enrichment of minority classes improves decision boundaries and strengthens feature correlations. Recent studies show that carefully designed prompts enable effective tabular data synthesis under imbalance \cite{kim2024epic,yang2024language} and that LLM-driven text augmentation significantly improves the performance of minority classes \cite{zhao2024improving}.
Besides, meta-learning with LLMs \cite{vassilev2023meta} opens a promising direction for imbalance-aware adaptation. By optimizing prompts or model parameters within a meta-learning framework, LLMs can learn to adapt quickly to minority classes from only a few examples. For instance, Vassilev et al. \cite{vassilev2023meta} propose a meta-learning approach that integrates diverse text representations and then combine it with threshold-moving, achieving robust improvements on highly imbalanced text classification tasks. This line of research highlights the potential of LLM-based meta-learning to complement traditional strategies with scalable and adaptive solutions for learning from minority classes.
}

%% file: 8.Conclusion.tex
\section{Conclusion}
\label{sec:conclusion}
This survey {provides a thorough and up-to-date overview} of imbalanced data learning, organizing the body of existing research into four distinct categories: data re-balancing, feature representation, training strategy, and ensemble learning. It delves into three critical real-world data formats, {highlighting their characteristics} and the unique challenges posed by imbalanced distributions. 

{Beyond synthesizing existing approaches, {we emphasize imbalance learning as both a foundational component and a beneficiary of modern machine learning advancements.} 
Crucially, the rise of generative AI and large-scale pretraining marks a transformative shift in the field. Generative models—such as GANs, VAEs, and diffusion models—enable the synthesis of high-quality, class-conditional samples, offering a principled solution to data scarcity in minority classes.
At the same time, the era of large language models and multimodal foundation models introduces new challenges in distributional bias and representation imbalance across modalities. These trends call for renewed focus on scalable, modality-aware imbalance learning methods to ensure robustness, fairness, and adaptability in this new landscape.}

{By bridging classical techniques with modern paradigms, this survey strives to act as a guiding light for both researchers and practitioners, navigating them through the complex terrain of imbalanced data and encouraging advancements in machine learning. We {hope it will act as a roadmap to inspire} future research and practical innovations in building inclusive, data-efficient, and balanced intelligent systems.}

%% file: bio.tex
\begin{IEEEbiography}[{\includegraphics[width=1in,height=1.25in,clip,keepaspectratio]{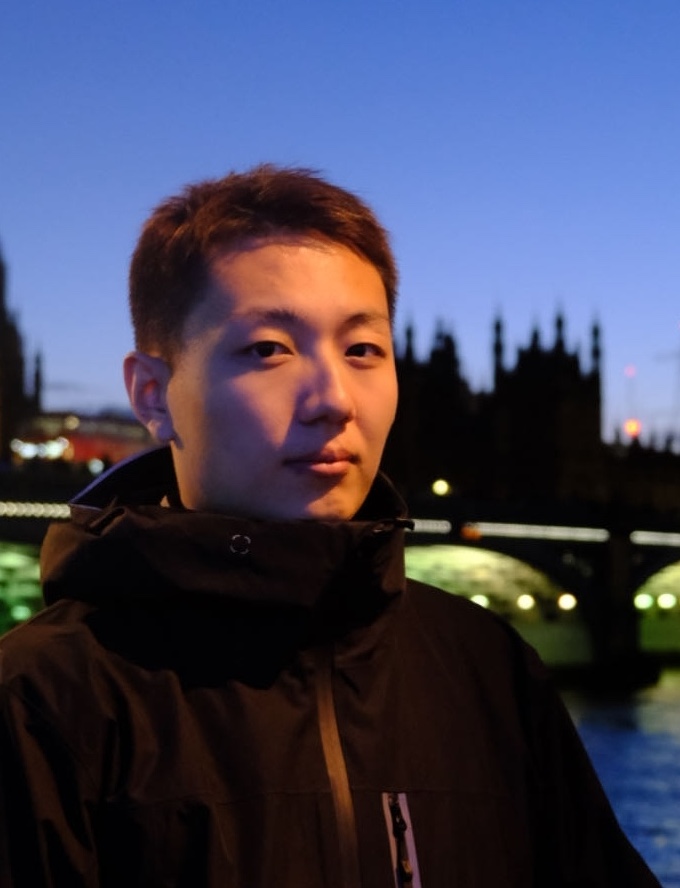}}]
{Xinyi Gao} is a Postdoctoral Researcher at the School of Electrical Engineering and Computer Science, The University of Queensland. He received his Ph.D. in Computer Science from The University of Queensland, following his B.E. and M.E. degrees in Information and Communications Engineering from Xi’an Jiaotong University, China. His research interests include data mining, graph representation learning, and multimedia signal processing.
\end{IEEEbiography}

\begin{IEEEbiography}[{\includegraphics[width=1in,height=1.25in,clip,keepaspectratio]{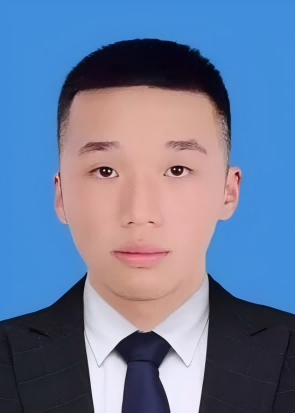}}]
{Dongting Xie} is an undergraduate student at the School of Information Science and Technology, Yunnan University. His research interests include data mining, large language models, and graph representation learning.
\end{IEEEbiography}

\begin{IEEEbiography}[{\includegraphics[width=1in,height=1.25in,clip,keepaspectratio]{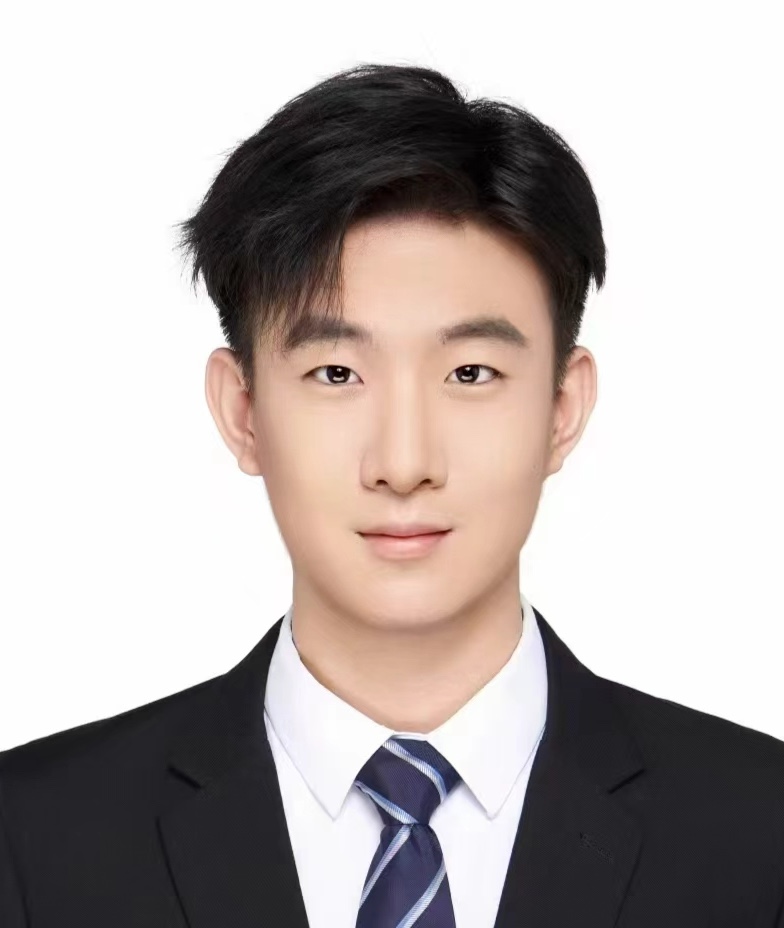}}]
{Yihang Zhang} received the BS degree in Mathematics and Computer Science from Peking University, China in 2024. He is currently pursuing the Master's degree in Data Science at Peking University. His research interests include machine learning and its applications in scientific discoveries.
\end{IEEEbiography}

\begin{IEEEbiography}[{\includegraphics[width=1in,height=1.25in,clip,keepaspectratio]{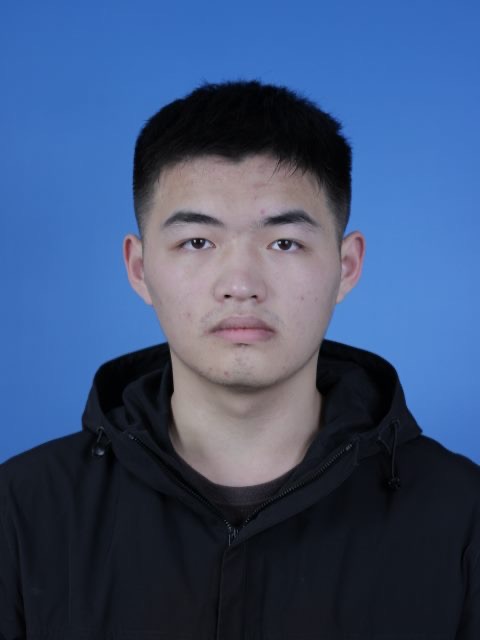}}]
{Zhengren Wang} received the B.S. degree in computer science from UESTC in 2023. He is currently a master student in Academy for Advanced Interdisciplinary Studies, Peking University, advised by Prof. Wentao Zhang. His Research interests include data-centric AI, generative AI and etc. He published papers as the first author in ACL, WWW, IJCAI, etc.
\end{IEEEbiography}

\begin{IEEEbiography}[{\includegraphics[width=1in,height=1.25in,clip,keepaspectratio]{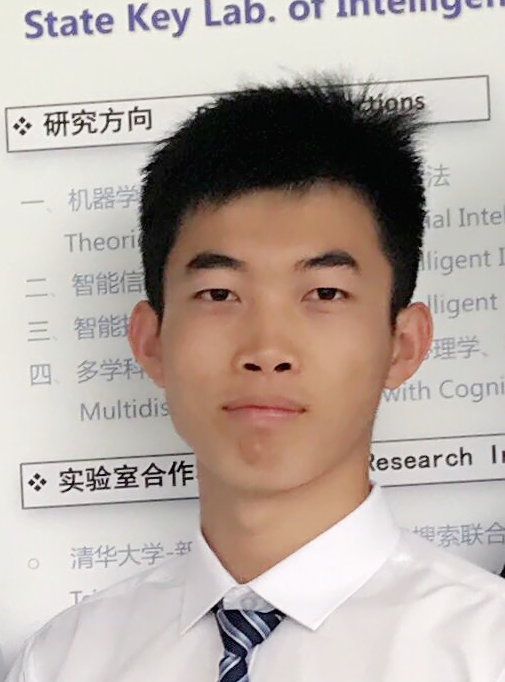}}]
{Chong Chen} is a Senior Researcher and Team Leader at Huawei Cloud. He obtained his Ph.D. and B.E. degrees in Computer Science and Technology from Tsinghua University. His research interests focus on large language model applications, including Search, RAG, NL2SQL, Recommendation, and related fields, with an emphasis on building scalable and intelligent information systems.
\end{IEEEbiography}

\begin{IEEEbiography}[{\includegraphics[width=1in,height=1.25in,clip,keepaspectratio]{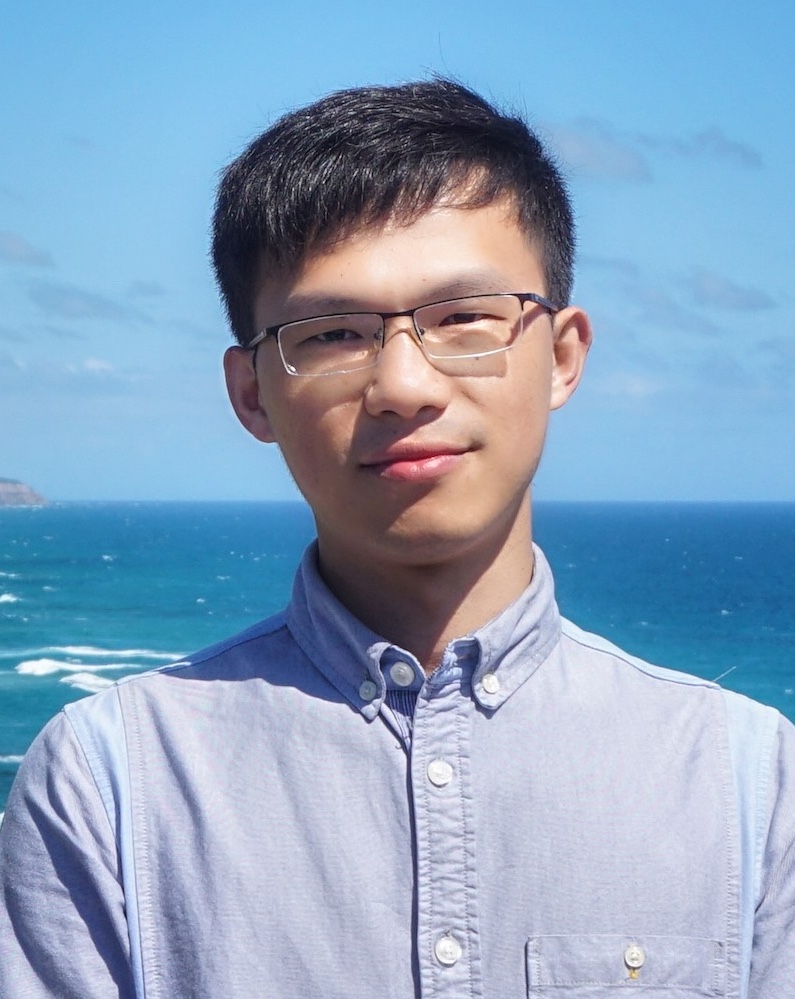}}]
{Conghui He} is a Young Leading Scientist at Shanghai AI Lab and an Adjunct Doctoral Supervisor at the School of AI, SJTU. Recognized as a National-level Young Talent, he received his Ph.D. from Tsinghua University and conducted research at Stanford University and Imperial College London. His work in data-centric AI and high-performance computing has led to 150+ papers in top-tier venues, 9,000+ Google Scholar citations, and open-source projects with 50,000+ GitHub stars.
\end{IEEEbiography}

\begin{IEEEbiography}[{\includegraphics[width=1in,height=1.25in,clip,keepaspectratio]{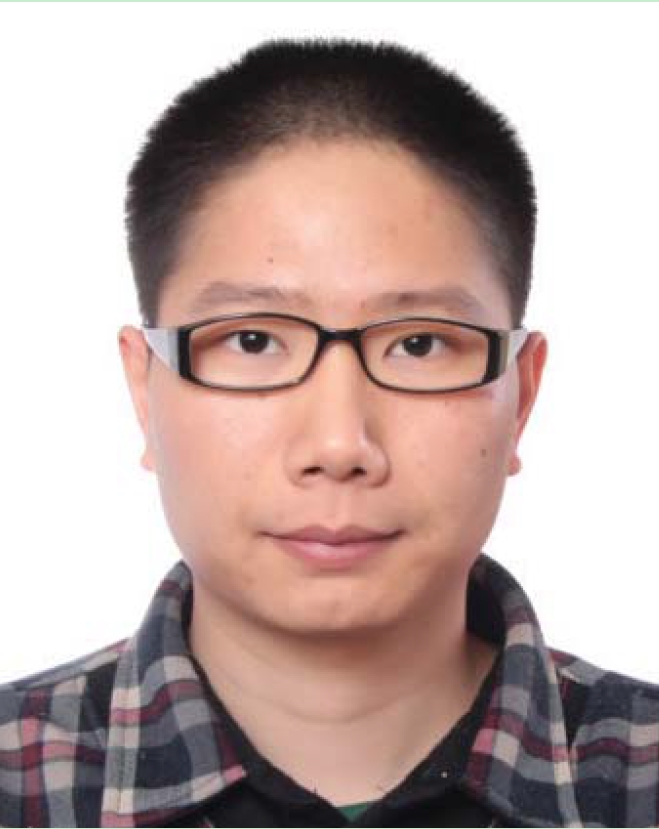}}]
{Hongzhi Yin} received a PhD degree in computer science from Peking University, in 2014. He works as an ARC Future Fellow and Full Professor at The University of Queensland, Australia. He has made notable contributions to recommendation systems,  graph learning, and decentralized and edge intelligence. He has published 300+ papers with an H-index of 75 and received ARC Future Fellowship 2021 and DECRA Award 2016.
\end{IEEEbiography}

\begin{IEEEbiography}[{\includegraphics[width=1in,height=1.25in,clip,keepaspectratio]{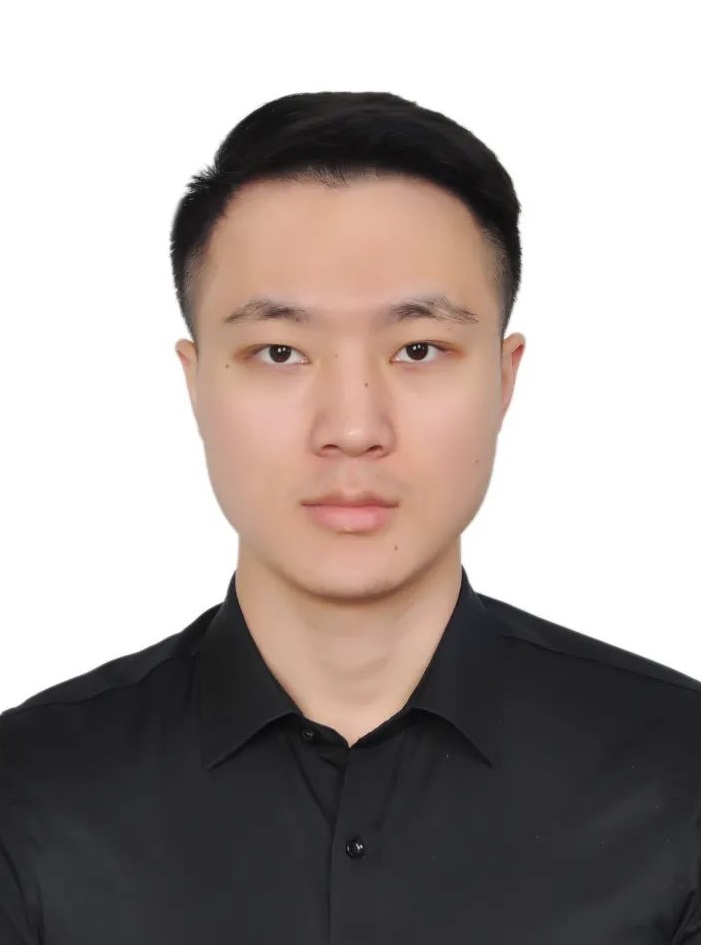}}]
{Wentao Zhang} is an assistant professor in the Center for Machine Learning Research at Peking University. Wentao’s research focuses on Graph ML, Data-centric ML, ML systems and AI4Science. Wentao has published 60+ papers in the top DB, DM, and ML venues. Wentao is the contributor of several system projects, including Angel, SGL, and OpenBox.
\end{IEEEbiography}

%% file: main.bbl
\begin{thebibliography}{100}
\providecommand{\url}[1]{#1}
\csname url@samestyle\endcsname
\providecommand{\newblock}{\relax}
\providecommand{\bibinfo}[2]{#2}
\providecommand{\BIBentrySTDinterwordspacing}{\spaceskip=0pt\relax}
\providecommand{\BIBentryALTinterwordstretchfactor}{4}
\providecommand{\BIBentryALTinterwordspacing}{\spaceskip=\fontdimen2\font plus
\BIBentryALTinterwordstretchfactor\fontdimen3\font minus \fontdimen4\font\relax}
\providecommand{\BIBforeignlanguage}[2]{{%
\expandafter\ifx\csname l@#1\endcsname\relax
\typeout{** WARNING: IEEEtran.bst: No hyphenation pattern has been}%
\typeout{** loaded for the language `#1'. Using the pattern for}%
\typeout{** the default language instead.}%
\else
\language=\csname l@#1\endcsname
\fi
#2}}
\providecommand{\BIBdecl}{\relax}
\BIBdecl

\bibitem{he2009learning}
H.~He and E.~A. Garcia, ``Learning from imbalanced data,'' \emph{IEEE Transactions on knowledge and data engineering}, vol.~21, no.~9, pp. 1263--1284, 2009.

\bibitem{krawczyk2016learning}
B.~Krawczyk, ``Learning from imbalanced data: open challenges and future directions,'' \emph{Progress in artificial intelligence}, vol.~5, no.~4, pp. 221--232, 2016.

\bibitem{haixiang2017learning}
G.~Haixiang, L.~Yijing, J.~Shang, G.~Mingyun, H.~Yuanyue, and G.~Bing, ``Learning from class-imbalanced data: Review of methods and applications,'' \emph{Expert systems with applications}, vol.~73, pp. 220--239, 2017.

\bibitem{zhang2023deep}
Y.~Zhang, B.~Kang, B.~Hooi, S.~Yan, and J.~Feng, ``Deep long-tailed learning: A survey,'' \emph{IEEE transactions on pattern analysis and machine intelligence}, vol.~45, no.~9, pp. 10\,795--10\,816, 2023.

\bibitem{ref_2}
N.~V. Chawla, K.~W. Bowyer, L.~O. Hall, and W.~P. Kegelmeyer, ``Smote: synthetic minority over-sampling technique,'' \emph{Journal of artificial intelligence research}, vol.~16, pp. 321--357, 2002.

\bibitem{zhang2017mixup}
H.~Zhang, M.~Cisse, Y.~N. Dauphin, and D.~Lopez-Paz, ``mixup: Beyond empirical risk minimization,'' \emph{arXiv preprint arXiv:1710.09412}, 2017.

\bibitem{ref_10}
J.~De~La~Calleja and O.~Fuentes, ``A distance-based over-sampling method for learning from imbalanced data sets.'' in \emph{FLAIRS conference}, 2007, pp. 634--635.

\bibitem{ref_11}
V.~Garc{\'\i}a, J.~S{\'a}nchez, and R.~Mollineda, ``On the use of surrounding neighbors for synthetic over-sampling of the minority class,'' in \emph{Proceedings of the 8th conference on Simulation, modelling and optimization}, 2008, pp. 389--394.

\bibitem{ref_4}
H.~Han, W.-Y. Wang, and B.-H. Mao, ``Borderline-smote: a new over-sampling method in imbalanced data sets learning,'' in \emph{International conference on intelligent computing}.\hskip 1em plus 0.5em minus 0.4em\relax Springer, 2005, pp. 878--887.

\bibitem{ref_5}
H.~He, Y.~Bai, E.~A. Garcia, and S.~Li, ``Adasyn: Adaptive synthetic sampling approach for imbalanced learning,'' in \emph{2008 IEEE international joint conference on neural networks (IEEE world congress on computational intelligence)}.\hskip 1em plus 0.5em minus 0.4em\relax IEEE, 2008, pp. 1322--1328.

\bibitem{ref_12}
S.~Hu, Y.~Liang, L.~Ma, and Y.~He, ``Msmote: Improving classification performance when training data is imbalanced,'' in \emph{2009 second international workshop on computer science and engineering}, vol.~2.\hskip 1em plus 0.5em minus 0.4em\relax IEEE, 2009, pp. 13--17.

\bibitem{smote-comparison}
G.~Kov\'acs, ``An empirical comparison and evaluation of minority oversampling techniques on a large number of imbalanced datasets,'' \emph{Applied Soft Computing}, vol.~83, p. 105662, 2019.

\bibitem{chou2020remix}
H.-P. Chou, S.-C. Chang, J.-Y. Pan, W.~Wei, and D.-C. Juan, ``Remix: rebalanced mixup,'' in \emph{Computer Vision--ECCV 2020 Workshops: Glasgow, UK, August 23--28, 2020, Proceedings, Part VI 16}.\hskip 1em plus 0.5em minus 0.4em\relax Springer, 2020, pp. 95--110.

\bibitem{kabra2020mixboost}
A.~Kabra, A.~Chopra, N.~Puri, P.~Badjatiya, S.~Verma, P.~Gupta, and B.~Krishnamurthy, ``Mixboost: Synthetic oversampling using boosted mixup for handling extreme imbalance,'' in \emph{2020 IEEE International Conference on Data Mining (ICDM)}.\hskip 1em plus 0.5em minus 0.4em\relax IEEE, 2020, pp. 1082--1087.

\bibitem{galdran2021balanced}
A.~Galdran, G.~Carneiro, and M.~A. Gonz{\'a}lez~Ballester, ``Balanced-mixup for highly imbalanced medical image classification,'' in \emph{Medical Image Computing and Computer Assisted Intervention--MICCAI 2021: 24th International Conference, Strasbourg, France, September 27--October 1, 2021, Proceedings, Part V 24}.\hskip 1em plus 0.5em minus 0.4em\relax Springer, 2021, pp. 323--333.

\bibitem{9746299}
S.~Zhang, C.~Chen, X.~Zhang, and S.~Peng, ``Label-occurrence-balanced mixup for long-tailed recognition,'' in \emph{ICASSP 2022 - 2022 IEEE International Conference on Acoustics, Speech and Signal Processing (ICASSP)}, 2022, pp. 3224--3228.

\bibitem{mathew2015kernel}
J.~Mathew, M.~Luo, C.~K. Pang, and H.~L. Chan, ``Kernel-based smote for svm classification of imbalanced datasets,'' in \emph{IECON 2015-41ST annual conference of the IEEE industrial electronics society}.\hskip 1em plus 0.5em minus 0.4em\relax IEEE, 2015, pp. 001\,127--001\,132.

\bibitem{tang2015kerneladasyn}
B.~Tang and H.~He, ``Kerneladasyn: Kernel based adaptive synthetic data generation for imbalanced learning,'' in \emph{2015 IEEE congress on evolutionary computation (CEC)}.\hskip 1em plus 0.5em minus 0.4em\relax IEEE, 2015, pp. 664--671.

\bibitem{8285168}
Z.~Wan, Y.~Zhang, and H.~He, ``Variational autoencoder based synthetic data generation for imbalanced learning,'' in \emph{2017 IEEE Symposium Series on Computational Intelligence (SSCI)}, 2017, pp. 1--7.

\bibitem{goodfellow2020generative}
I.~Goodfellow, J.~Pouget-Abadie, M.~Mirza, B.~Xu, D.~Warde-Farley, S.~Ozair, A.~Courville, and Y.~Bengio, ``Generative adversarial networks,'' \emph{Communications of the ACM}, vol.~63, no.~11, pp. 139--144, 2020.

\bibitem{douzas2018effective}
G.~Douzas and F.~Bacao, ``Effective data generation for imbalanced learning using conditional generative adversarial networks,'' \emph{Expert Systems with Applications}, vol.~91, pp. 464--471, 2018.

\bibitem{mirza2014conditional}
M.~Mirza and S.~Osindero, ``Conditional generative adversarial nets,'' \emph{arXiv preprint arXiv:1411.1784}, 2014.

\bibitem{mariani2018bagan}
G.~Mariani, F.~Scheidegger, R.~Istrate, C.~Bekas, and C.~Malossi, ``Bagan: Data augmentation with balancing gan,'' \emph{arXiv preprint arXiv:1803.09655}, 2018.

\bibitem{cai2019supervised}
Z.~Cai, X.~Wang, M.~Zhou, J.~Xu, and L.~Jing, ``Supervised class distribution learning for gans-based imbalanced classification,'' in \emph{2019 IEEE International Conference on Data Mining (ICDM)}.\hskip 1em plus 0.5em minus 0.4em\relax IEEE, 2019, pp. 41--50.

\bibitem{ding2023rvgan}
H.~Ding, Y.~Sun, N.~Huang, Z.~Shen, Z.~Wang, A.~Iftekhar, and X.~Cui, ``Rvgan-tl: A generative adversarial networks and transfer learning-based hybrid approach for imbalanced data classification,'' \emph{Information Sciences}, vol. 629, pp. 184--203, 2023.

\bibitem{tolstikhin2017wasserstein}
I.~Tolstikhin, O.~Bousquet, S.~Gelly, and B.~Schoelkopf, ``Wasserstein auto-encoders,'' \emph{arXiv preprint arXiv:1711.01558}, 2017.

\bibitem{ren2019ewgan}
J.~Ren, Y.~Liu, and J.~Liu, ``Ewgan: Entropy-based wasserstein gan for imbalanced learning,'' in \emph{Proceedings of the AAAI conference on artificial intelligence}, vol.~33, no.~01, 2019, pp. 10\,011--10\,012.

\bibitem{arjovsky2017wasserstein}
M.~Arjovsky, S.~Chintala, and L.~Bottou, ``Wasserstein generative adversarial networks,'' in \emph{International conference on machine learning}.\hskip 1em plus 0.5em minus 0.4em\relax PMLR, 2017, pp. 214--223.

\bibitem{li2022eid}
W.~Li, J.~Chen, J.~Cao, C.~Ma, J.~Wang, X.~Cui, and P.~Chen, ``Eid-gan: Generative adversarial nets for extremely imbalanced data augmentation,'' \emph{IEEE Transactions on Industrial Informatics}, vol.~19, no.~3, pp. 3208--3218, 2022.

\bibitem{sharma2022smotified}
A.~Sharma, P.~Singh, and R.~Chandra, ``Smotified-gan for class imbalanced pattern classification problems,'' \emph{IEEE Access}, vol.~10, pp. 1--1, 01 2022.

\bibitem{mullick2019generative}
S.~S. Mullick, S.~Datta, and S.~Das, ``Generative adversarial minority oversampling,'' in \emph{Proceedings of the IEEE/CVF International Conference on Computer Vision (ICCV)}, October 2019.

\bibitem{ho2020denoising}
J.~Ho, A.~Jain, and P.~Abbeel, ``Denoising diffusion probabilistic models,'' \emph{Advances in neural information processing systems}, vol.~33, pp. 6840--6851, 2020.

\bibitem{wang2022semantic}
W.~Wang, J.~Bao, W.~Zhou, D.~Chen, D.~Chen, L.~Yuan, and H.~Li, ``Semantic image synthesis via diffusion models,'' \emph{arXiv preprint arXiv:2207.00050}, 2022.

\bibitem{oh2023diffmix}
H.-J. Oh and W.-K. Jeong, ``Diffmix: Diffusion model-based data synthesis for nuclei segmentation and classification in imbalanced pathology image datasets,'' \emph{arXiv preprint arXiv:2306.14132}, 2023.

\bibitem{liu2023diffusion}
Q.~Liu, F.~Yan, X.~Zhao, Z.~Du, H.~Guo, R.~Tang, and F.~Tian, ``Diffusion augmentation for sequential recommendation,'' \emph{arXiv preprint arXiv:2309.12858}, 2023.

\bibitem{xie2023mosaicfusion}
J.~Xie, W.~Li, X.~Li, Z.~Liu, Y.~S. Ong, and C.~C. Loy, ``Mosaicfusion: Diffusion models as data augmenters for large vocabulary instance segmentation,'' \emph{arXiv preprint arXiv:2309.13042}, 2023.

\bibitem{wang2025pogdiff}
Z.~Wang, S.~Wei, X.~Huo, and H.~Wang, ``Pogdiff: Product-of-gaussians diffusion models for imbalanced text-to-image generation,'' \emph{arXiv preprint arXiv:2502.08106}, 2025.

\bibitem{yan2024training}
D.~Yan, L.~Qi, V.~T. Hu, M.-H. Yang, and M.~Tang, ``Training class-imbalanced diffusion model via overlap optimization,'' \emph{arXiv preprint arXiv:2402.10821}, 2024.

\bibitem{pengxiao2024latent}
H.~Pengxiao, Y.~Changkun, Z.~Jieming, Z.~Jing, J.~Hong, and L.~Xuesong, ``Latent-based diffusion model for long-tailed recognition,'' in \emph{This CVPR Workshop paper is the Open Access version, provided by the Computer Vision Foundation. Except for this watermark, it is identical to the accepted version; the final published version of the proceedings is available on IEEE Xplore}, 2024.

\bibitem{xu2024rethinking}
C.~Xu, J.~Yan, M.~Yang, and C.~Deng, ``Rethinking noise sampling in class-imbalanced diffusion models,'' \emph{IEEE Transactions on Image Processing}, 2024.

\bibitem{kubat1997addressing}
M.~Kubat, S.~Matwin \emph{et~al.}, ``Addressing the curse of imbalanced training sets: one-sided selection,'' in \emph{Icml}, vol.~97, no.~1.\hskip 1em plus 0.5em minus 0.4em\relax Citeseer, 1997, p. 179.

\bibitem{ref_6}
P.~Hart, ``The condensed nearest neighbor rule (corresp.),'' \emph{IEEE transactions on information theory}, vol.~14, no.~3, pp. 515--516, 1968.

\bibitem{ref_17}
I.~Tomek, ``An experiment with the edited nearest-neighbor rule,'' \emph{IEEE Transactions on Systems, Man, and Cybernetics}, vol. SMC-6, no.~6, pp. 448--452, 1976.

\bibitem{ref_18}
J.~Laurikkala, ``Improving identification of difficult small classes by balancing class distribution,'' in \emph{Conference on artificial intelligence in medicine in Europe}.\hskip 1em plus 0.5em minus 0.4em\relax Springer, 2001, pp. 63--66.

\bibitem{guo2003knn}
G.~Guo, H.~Wang, D.~Bell, Y.~Bi, and K.~Greer, ``Knn model-based approach in classification,'' in \emph{On The Move to Meaningful Internet Systems 2003: CoopIS, DOA, and ODBASE: OTM Confederated International Conferences, CoopIS, DOA, and ODBASE 2003, Catania, Sicily, Italy, November 3-7, 2003. Proceedings}.\hskip 1em plus 0.5em minus 0.4em\relax Springer, 2003, pp. 986--996.

\bibitem{ref_19}
I.~Mani and I.~Zhang, ``knn approach to unbalanced data distributions: a case study involving information extraction,'' in \emph{Proceedings of workshop on learning from imbalanced datasets}, vol. 126.\hskip 1em plus 0.5em minus 0.4em\relax ICML, 2003, pp. 1--7.

\bibitem{vuttipittayamongkol2020neighbourhood}
P.~Vuttipittayamongkol and E.~Elyan, ``Neighbourhood-based undersampling approach for handling imbalanced and overlapped data,'' \emph{Information Sciences}, vol. 509, pp. 47--70, 2020.

\bibitem{yen2009cluster}
S.-J. Yen and Y.-S. Lee, ``Cluster-based under-sampling approaches for imbalanced data distributions,'' \emph{Expert Systems with Applications}, vol.~36, no.~3, pp. 5718--5727, 2009.

\bibitem{zhang2010cluster}
Y.-P. Zhang, L.-N. Zhang, and Y.-C. Wang, ``Cluster-based majority under-sampling approaches for class imbalance learning,'' in \emph{2010 2nd IEEE International Conference on Information and Financial Engineering}.\hskip 1em plus 0.5em minus 0.4em\relax IEEE, 2010, pp. 400--404.

\bibitem{lin2017clustering}
W.-C. Lin, C.-F. Tsai, Y.-H. Hu, and J.-S. Jhang, ``Clustering-based undersampling in class-imbalanced data,'' \emph{Information Sciences}, vol. 409, pp. 17--26, 2017.

\bibitem{ng2014diversified}
W.~W. Ng, J.~Hu, D.~S. Yeung, S.~Yin, and F.~Roli, ``Diversified sensitivity-based undersampling for imbalance classification problems,'' \emph{IEEE transactions on cybernetics}, vol.~45, no.~11, pp. 2402--2412, 2014.

\bibitem{ofek2017fast}
N.~Ofek, L.~Rokach, R.~Stern, and A.~Shabtai, ``Fast-cbus: A fast clustering-based undersampling method for addressing the class imbalance problem,'' \emph{Neurocomputing}, vol. 243, pp. 88--102, 2017.

\bibitem{garcia2009evolutionary}
S.~Garc{\'\i}a and F.~Herrera, ``Evolutionary undersampling for classification with imbalanced datasets: Proposals and taxonomy,'' \emph{Evolutionary computation}, vol.~17, no.~3, pp. 275--306, 2009.

\bibitem{triguero2015evolutionary}
I.~Triguero, M.~Galar, S.~Vluymans, C.~Cornelis, H.~Bustince, F.~Herrera, and Y.~Saeys, ``Evolutionary undersampling for imbalanced big data classification,'' in \emph{2015 IEEE Congress on Evolutionary Computation (CEC)}.\hskip 1em plus 0.5em minus 0.4em\relax IEEE, 2015, pp. 715--722.

\bibitem{le2021eusc}
H.~L. Le, D.~Landa-Silva, M.~Galar, S.~Garcia, and I.~Triguero, ``Eusc: A clustering-based surrogate model to accelerate evolutionary undersampling in imbalanced classification,'' \emph{Applied Soft Computing}, vol. 101, p. 107033, 2021.

\bibitem{galar2013eusboost}
M.~Galar, A.~Fern{\'a}ndez, E.~Barrenechea, and F.~Herrera, ``Eusboost: Enhancing ensembles for highly imbalanced data-sets by evolutionary undersampling,'' \emph{Pattern recognition}, vol.~46, no.~12, pp. 3460--3471, 2013.

\bibitem{krawczyk2016evolutionary}
B.~Krawczyk, M.~Galar, {\L}.~Jele{\'n}, and F.~Herrera, ``Evolutionary undersampling boosting for imbalanced classification of breast cancer malignancy,'' \emph{Applied Soft Computing}, vol.~38, pp. 714--726, 2016.

\bibitem{peng2019trainable}
M.~Peng, Q.~Zhang, X.~Xing, T.~Gui, X.~Huang, Y.-G. Jiang, K.~Ding, and Z.~Chen, ``Trainable undersampling for class-imbalance learning,'' in \emph{Proceedings of the AAAI conference on artificial intelligence}, vol.~33, no.~01, 2019, pp. 4707--4714.

\bibitem{9741366}
Y.~Yan, Y.~Zhu, R.~Liu, Y.~Zhang, Y.~Zhang, and L.~Zhang, ``Spatial distribution-based imbalanced undersampling,'' \emph{IEEE Transactions on Knowledge and Data Engineering}, vol.~35, no.~6, pp. 6376--6391, 2023.

\bibitem{hoyos2021relevant}
J.~Hoyos-Osorio, A.~Alvarez-Meza, G.~Daza-Santacoloma, A.~Orozco-Gutierrez, and G.~Castellanos-Dominguez, ``Relevant information undersampling to support imbalanced data classification,'' \emph{Neurocomputing}, vol. 436, pp. 136--146, 2021.

\bibitem{8968753}
Z.~Wang, C.~Cao, and Y.~Zhu, ``Entropy and confidence-based undersampling boosting random forests for imbalanced problems,'' \emph{IEEE Transactions on Neural Networks and Learning Systems}, vol.~31, no.~12, pp. 5178--5191, 2020.

\bibitem{batista2004study}
G.~E. Batista, R.~C. Prati, and M.~C. Monard, ``A study of the behavior of several methods for balancing machine learning training data,'' \emph{ACM SIGKDD explorations newsletter}, vol.~6, no.~1, pp. 20--29, 2004.

\bibitem{feng2020cluster}
S.~Feng, C.~Zhao, and P.~Fu, ``A cluster-based hybrid sampling approach for imbalanced data classification,'' \emph{Review of Scientific Instruments}, vol.~91, no.~5, 2020.

\bibitem{ramentol2012smote}
E.~Ramentol, Y.~Caballero, R.~Bello, and F.~Herrera, ``Smote-rsb*: a hybrid preprocessing approach based on oversampling and undersampling for high imbalanced data-sets using smote and rough sets theory,'' \emph{Knowledge and information systems}, vol.~33, pp. 245--265, 2012.

\bibitem{saez2015smote}
J.~A. S{\'a}ez, J.~Luengo, J.~Stefanowski, and F.~Herrera, ``Smote--ipf: Addressing the noisy and borderline examples problem in imbalanced classification by a re-sampling method with filtering,'' \emph{Information Sciences}, vol. 291, pp. 184--203, 2015.

\bibitem{gao2020ensemble}
X.~Gao, B.~Ren, H.~Zhang, B.~Sun, J.~Li, J.~Xu, Y.~He, and K.~Li, ``An ensemble imbalanced classification method based on model dynamic selection driven by data partition hybrid sampling,'' \emph{Expert Systems with Applications}, vol. 160, p. 113660, 2020.

\bibitem{zhou2020bbn}
B.~Zhou, Q.~Cui, X.-S. Wei, and Z.-M. Chen, ``Bbn: Bilateral-branch network with cumulative learning for long-tailed visual recognition,'' in \emph{Proceedings of the IEEE/CVF conference on computer vision and pattern recognition}, 2020, pp. 9719--9728.

\bibitem{wang2020devil}
T.~Wang, Y.~Li, B.~Kang, J.~Li, J.~Liew, S.~Tang, S.~Hoi, and J.~Feng, ``The devil is in classification: A simple framework for long-tail instance segmentation,'' in \emph{Computer Vision--ECCV 2020: 16th European Conference, Glasgow, UK, August 23--28, 2020, Proceedings, Part XIV 16}.\hskip 1em plus 0.5em minus 0.4em\relax Springer, 2020, pp. 728--744.

\bibitem{guo2021long}
H.~Guo and S.~Wang, ``Long-tailed multi-label visual recognition by collaborative training on uniform and re-balanced samplings,'' in \emph{Proceedings of the IEEE/CVF Conference on Computer Vision and Pattern Recognition}, 2021, pp. 15\,089--15\,098.

\bibitem{li2020overcoming}
Y.~Li, T.~Wang, B.~Kang, S.~Tang, C.~Wang, J.~Li, and J.~Feng, ``Overcoming classifier imbalance for long-tail object detection with balanced group softmax,'' in \emph{Proceedings of the IEEE/CVF conference on computer vision and pattern recognition}, 2020, pp. 10\,991--11\,000.

\bibitem{xiang2020learning}
L.~Xiang, G.~Ding, and J.~Han, ``Learning from multiple experts: Self-paced knowledge distillation for long-tailed classification,'' in \emph{Computer Vision--ECCV 2020: 16th European Conference, Glasgow, UK, August 23--28, 2020, Proceedings, Part V 16}.\hskip 1em plus 0.5em minus 0.4em\relax Springer, 2020, pp. 247--263.

\bibitem{cai2021ace}
J.~Cai, Y.~Wang, and J.-N. Hwang, ``Ace: Ally complementary experts for solving long-tailed recognition in one-shot,'' in \emph{Proceedings of the IEEE/CVF International Conference on Computer Vision}, 2021, pp. 112--121.

\bibitem{cui2022reslt}
J.~Cui, S.~Liu, Z.~Tian, Z.~Zhong, and J.~Jia, ``Reslt: Residual learning for long-tailed recognition,'' \emph{IEEE transactions on pattern analysis and machine intelligence}, vol.~45, no.~3, pp. 3695--3706, 2022.

\bibitem{zhang2022self}
Y.~Zhang, B.~Hooi, L.~Hong, and J.~Feng, ``Self-supervised aggregation of diverse experts for test-agnostic long-tailed recognition,'' \emph{Advances in Neural Information Processing Systems}, vol.~35, pp. 34\,077--34\,090, 2022.

\bibitem{koziarski2021potential}
M.~Koziarski, ``Potential anchoring for imbalanced data classification,'' \emph{Pattern Recognition}, vol. 120, p. 108114, 2021.

\bibitem{xu2022constructing}
Y.~Xu, Y.-L. Li, J.~Li, and C.~Lu, ``Constructing balance from imbalance for long-tailed image recognition,'' in \emph{European Conference on Computer Vision}.\hskip 1em plus 0.5em minus 0.4em\relax Springer, 2022, pp. 38--56.

\bibitem{zhu2020ehso}
Y.~Zhu, Y.~Yan, Y.~Zhang, and Y.~Zhang, ``Ehso: Evolutionary hybrid sampling in overlapping scenarios for imbalanced learning,'' \emph{Neurocomputing}, vol. 417, pp. 333--346, 2020.

\bibitem{pouyanfar2018dynamic}
S.~Pouyanfar, Y.~Tao, A.~Mohan, H.~Tian, A.~S. Kaseb, K.~Gauen, R.~Dailey, S.~Aghajanzadeh, Y.-H. Lu, S.-C. Chen \emph{et~al.}, ``Dynamic sampling in convolutional neural networks for imbalanced data classification,'' in \emph{2018 IEEE conference on multimedia information processing and retrieval (MIPR)}.\hskip 1em plus 0.5em minus 0.4em\relax IEEE, 2018, pp. 112--117.

\bibitem{yang2020rethinking}
Y.~Yang and Z.~Xu, ``Rethinking the value of labels for improving class-imbalanced learning,'' \emph{Advances in neural information processing systems}, vol.~33, pp. 19\,290--19\,301, 2020.

\bibitem{wei2021crest}
C.~Wei, K.~Sohn, C.~Mellina, A.~Yuille, and F.~Yang, ``Crest: A class-rebalancing self-training framework for imbalanced semi-supervised learning,'' in \emph{Proceedings of the IEEE/CVF conference on computer vision and pattern recognition}, 2021, pp. 10\,857--10\,866.

\bibitem{Lai_2022_CVPR}
Z.~Lai, C.~Wang, S.-c. Cheung, and C.-N. Chuah, ``Sar: Self-adaptive refinement on pseudo labels for multiclass-imbalanced semi-supervised learning,'' in \emph{Proceedings of the IEEE/CVF Conference on Computer Vision and Pattern Recognition (CVPR) Workshops}, June 2022, pp. 4091--4100.

\bibitem{oh2022daso}
Y.~Oh, D.-J. Kim, and I.~S. Kweon, ``Daso: Distribution-aware semantics-oriented pseudo-label for imbalanced semi-supervised learning,'' in \emph{Proceedings of the IEEE/CVF Conference on Computer Vision and Pattern Recognition}, 2022, pp. 9786--9796.

\bibitem{ertekin2007active}
S.~Ertekin, J.~Huang, and C.~L. Giles, ``Active learning for class imbalance problem,'' in \emph{Proceedings of the 30th annual international ACM SIGIR conference on Research and development in information retrieval}, 2007, pp. 823--824.

\bibitem{qin2021active}
J.~Qin, C.~Wang, Q.~Zou, Y.~Sun, and B.~Chen, ``Active learning with extreme learning machine for online imbalanced multiclass classification,'' \emph{Knowledge-Based Systems}, vol. 231, p. 107385, 2021.

\bibitem{zhu2007active}
J.~Zhu and E.~Hovy, ``Active learning for word sense disambiguation with methods for addressing the class imbalance problem,'' in \emph{Proceedings of the 2007 Joint Conference on Empirical Methods in Natural Language Processing and Computational Natural Language Learning (EMNLP-CoNLL)}, 2007, pp. 783--790.

\bibitem{aggarwal2021minority}
U.~Aggarwal, A.~Popescu, and C.~Hudelot, ``Minority class oriented active learning for imbalanced datasets,'' in \emph{2020 25th International Conference on Pattern Recognition (ICPR)}.\hskip 1em plus 0.5em minus 0.4em\relax IEEE, 2021, pp. 9920--9927.

\bibitem{liu2021comprehensive}
W.~Liu, H.~Zhang, Z.~Ding, Q.~Liu, and C.~Zhu, ``A comprehensive active learning method for multiclass imbalanced data streams with concept drift,'' \emph{Knowledge-Based Systems}, vol. 215, p. 106778, 2021.

\bibitem{dong2020cost}
H.~Dong, B.~Zhu, and J.~Zhang, ``A cost-sensitive active learning for imbalance data with uncertainty and diversity combination,'' in \emph{Proceedings of the 2020 12th International Conference on Machine Learning and Computing}, 2020, pp. 218--224.

\bibitem{jin2022deep}
Q.~Jin, M.~Yuan, H.~Wang, M.~Wang, and Z.~Song, ``Deep active learning models for imbalanced image classification,'' \emph{Knowledge-Based Systems}, vol. 257, p. 109817, 2022.

\bibitem{fu2011certainty}
J.~H. Fu and S.~L. Lee, ``Certainty-enhanced active learning for improving imbalanced data classification,'' in \emph{2011 IEEE 11th International Conference on Data Mining Workshops}.\hskip 1em plus 0.5em minus 0.4em\relax IEEE, 2011, pp. 405--412.

\bibitem{8595005}
C.~Zhang, W.~Tavanapong, G.~Kijkul, J.~Wong, P.~C. de~Groen, and J.~Oh, ``Similarity-based active learning for image classification under class imbalance,'' in \emph{2018 IEEE International Conference on Data Mining (ICDM)}, 2018, pp. 1422--1427.

\bibitem{ref_A}
P.~D. Turney, ``Cost-sensitive classification: Empirical evaluation of a hybrid genetic decision tree induction algorithm,'' \emph{Journal of artificial intelligence research}, vol.~2, pp. 369--409, 1994.

\bibitem{ref_C}
V.~S. Sheng and C.~X. Ling, ``Thresholding for making classifiers cost-sensitive,'' in \emph{Aaai}, vol.~6, 2006, pp. 476--81.

\bibitem{ref_B}
P.~Domingos, ``Metacost: A general method for making classifiers cost-sensitive,'' in \emph{Proceedings of the fifth ACM SIGKDD international conference on Knowledge discovery and data mining}, 1999, pp. 155--164.

\bibitem{zhou2005training}
Z.-H. Zhou and X.-Y. Liu, ``Training cost-sensitive neural networks with methods addressing the class imbalance problem,'' \emph{IEEE Transactions on knowledge and data engineering}, vol.~18, no.~1, pp. 63--77, 2005.

\bibitem{zhou2010multi}
------, ``On multi-class cost-sensitive learning,'' \emph{Computational Intelligence}, vol.~26, no.~3, pp. 232--257, 2010.

\bibitem{castro2013novel}
C.~L. Castro and A.~P. Braga, ``Novel cost-sensitive approach to improve the multilayer perceptron performance on imbalanced data,'' \emph{IEEE transactions on neural networks and learning systems}, vol.~24, no.~6, pp. 888--899, 2013.

\bibitem{cui2019class}
Y.~Cui, M.~Jia, T.-Y. Lin, Y.~Song, and S.~Belongie, ``Class-balanced loss based on effective number of samples,'' in \emph{Proceedings of the IEEE/CVF conference on computer vision and pattern recognition}, 2019, pp. 9268--9277.

\bibitem{lin2017focal}
T.-Y. Lin, P.~Goyal, R.~Girshick, K.~He, and P.~Doll{\'a}r, ``Focal loss for dense object detection,'' in \emph{Proceedings of the IEEE international conference on computer vision}, 2017, pp. 2980--2988.

\bibitem{park2021influence}
S.~Park, J.~Lim, Y.~Jeon, and J.~Y. Choi, ``Influence-balanced loss for imbalanced visual classification,'' in \emph{Proceedings of the IEEE/CVF International Conference on Computer Vision}, 2021, pp. 735--744.

\bibitem{ren2020balanced}
J.~Ren, C.~Yu, X.~Ma, H.~Zhao, S.~Yi \emph{et~al.}, ``Balanced meta-softmax for long-tailed visual recognition,'' \emph{Advances in neural information processing systems}, vol.~33, pp. 4175--4186, 2020.

\bibitem{cao2013novel}
P.~Cao, B.~Li, D.~Zhao, and O.~Zaiane, ``A novel cost sensitive neural network ensemble for multiclass imbalance data learning,'' in \emph{The 2013 International Joint Conference on Neural Networks (IJCNN)}.\hskip 1em plus 0.5em minus 0.4em\relax IEEE, 2013, pp. 1--8.

\bibitem{zhang2018cost}
C.~Zhang, K.~C. Tan, H.~Li, and G.~S. Hong, ``A cost-sensitive deep belief network for imbalanced classification,'' \emph{IEEE transactions on neural networks and learning systems}, vol.~30, no.~1, pp. 109--122, 2018.

\bibitem{chen2021generalized}
C.~Chen, S.~Zheng, X.~Chen, E.~Dong, X.~S. Liu, H.~Liu, and D.~Dou, ``Generalized dataweighting via class-level gradient manipulation,'' \emph{Advances in Neural Information Processing Systems}, vol.~34, pp. 14\,097--14\,109, 2021.

\bibitem{tan2020equalization}
J.~Tan, C.~Wang, B.~Li, Q.~Li, W.~Ouyang, C.~Yin, and J.~Yan, ``Equalization loss for long-tailed object recognition,'' in \emph{Proceedings of the IEEE/CVF conference on computer vision and pattern recognition}, 2020, pp. 11\,662--11\,671.

\bibitem{tan2021equalization}
J.~Tan, X.~Lu, G.~Zhang, C.~Yin, and Q.~Li, ``Equalization loss v2: A new gradient balance approach for long-tailed object detection,'' in \emph{Proceedings of the IEEE/CVF conference on computer vision and pattern recognition}, 2021, pp. 1685--1694.

\bibitem{guo2022learning}
D.~Guo, Z.~Li, H.~Zhao, M.~Zhou, H.~Zha \emph{et~al.}, ``Learning to re-weight examples with optimal transport for imbalanced classification,'' \emph{Advances in Neural Information Processing Systems}, vol.~35, pp. 25\,517--25\,530, 2022.

\bibitem{ren2018learning}
M.~Ren, W.~Zeng, B.~Yang, and R.~Urtasun, ``Learning to reweight examples for robust deep learning,'' in \emph{International conference on machine learning}.\hskip 1em plus 0.5em minus 0.4em\relax PMLR, 2018, pp. 4334--4343.

\bibitem{shu2019meta}
J.~Shu, Q.~Xie, L.~Yi, Q.~Zhao, S.~Zhou, Z.~Xu, and D.~Meng, ``Meta-weight-net: Learning an explicit mapping for sample weighting,'' \emph{Advances in neural information processing systems}, vol.~32, 2019.

\bibitem{duan2017deep}
Y.~Duan, J.~Lu, J.~Feng, and J.~Zhou, ``Deep localized metric learning,'' \emph{IEEE Transactions on Circuits and Systems for Video Technology}, vol.~28, no.~10, pp. 2644--2656, 2017.

\bibitem{huang2016learning}
C.~Huang, Y.~Li, C.~C. Loy, and X.~Tang, ``Learning deep representation for imbalanced classification,'' in \emph{Proceedings of the IEEE conference on computer vision and pattern recognition}, 2016, pp. 5375--5384.

\bibitem{dong2017class}
Q.~Dong, S.~Gong, and X.~Zhu, ``Class rectification hard mining for imbalanced deep learning,'' in \emph{Proceedings of the IEEE international conference on computer vision}, 2017, pp. 1851--1860.

\bibitem{wang2018iterative}
N.~Wang, X.~Zhao, Y.~Jiang, Y.~Gao, and K.~BNRist, ``Iterative metric learning for imbalance data classification,'' in \emph{IJCAI}, vol. 2018, 2018, pp. 2805--2811.

\bibitem{gui2022quadruplet}
X.~Gui, J.~Zhang, J.~Tang, H.~Xu, J.~Zou, and S.~Fan, ``A quadruplet deep metric learning model for imbalanced time-series fault diagnosis,'' \emph{Knowledge-Based Systems}, vol. 238, p. 107932, 2022.

\bibitem{yan2023borderline}
M.~Yan and N.~Li, ``Borderline-margin loss based deep metric learning framework for imbalanced data,'' \emph{Applied Intelligence}, vol.~53, no.~2, pp. 1487--1504, 2023.

\bibitem{cheng2016person}
D.~Cheng, Y.~Gong, S.~Zhou, J.~Wang, and N.~Zheng, ``Person re-identification by multi-channel parts-based cnn with improved triplet loss function,'' in \emph{Proceedings of the iEEE conference on computer vision and pattern recognition}, 2016, pp. 1335--1344.

\bibitem{zhang2017range}
X.~Zhang, Z.~Fang, Y.~Wen, Z.~Li, and Y.~Qiao, ``Range loss for deep face recognition with long-tailed training data,'' in \emph{Proceedings of the IEEE International Conference on Computer Vision}, 2017, pp. 5409--5418.

\bibitem{feng2018learning}
L.~Feng, H.~Wang, B.~Jin, H.~Li, M.~Xue, and L.~Wang, ``Learning a distance metric by balancing kl-divergence for imbalanced datasets,'' \emph{IEEE Transactions on Systems, Man, and Cybernetics: Systems}, vol.~49, no.~12, pp. 2384--2395, 2018.

\bibitem{hayat2019gaussian}
M.~Hayat, S.~Khan, S.~W. Zamir, J.~Shen, and L.~Shao, ``Gaussian affinity for max-margin class imbalanced learning,'' in \emph{Proceedings of the IEEE/CVF international conference on computer vision}, 2019, pp. 6469--6479.

\bibitem{gautheron2020metric}
L.~Gautheron, A.~Habrard, E.~Morvant, and M.~Sebban, ``Metric learning from imbalanced data with generalization guarantees,'' \emph{Pattern Recognition Letters}, vol. 133, pp. 298--304, 2020.

\bibitem{khosla2020supervised}
P.~Khosla, P.~Teterwak, C.~Wang, A.~Sarna, Y.~Tian, P.~Isola, A.~Maschinot, C.~Liu, and D.~Krishnan, ``Supervised contrastive learning,'' \emph{Advances in neural information processing systems}, vol.~33, pp. 18\,661--18\,673, 2020.

\bibitem{weinberger2009distance}
K.~Q. Weinberger and L.~K. Saul, ``Distance metric learning for large margin nearest neighbor classification.'' \emph{Journal of machine learning research}, vol.~10, no.~2, 2009.

\bibitem{sohn2016improved}
K.~Sohn, ``Improved deep metric learning with multi-class n-pair loss objective,'' \emph{Advances in neural information processing systems}, vol.~29, 2016.

\bibitem{tian2020contrastive}
Y.~Tian, D.~Krishnan, and P.~Isola, ``Contrastive multiview coding,'' in \emph{Computer Vision--ECCV 2020: 16th European Conference, Glasgow, UK, August 23--28, 2020, Proceedings, Part XI 16}.\hskip 1em plus 0.5em minus 0.4em\relax Springer, 2020, pp. 776--794.

\bibitem{kang2021exploring}
B.~Kang, Y.~Li, S.~Xie, Z.~Yuan, and J.~Feng, ``Exploring balanced feature spaces for representation learning,'' in \emph{International Conference on Learning Representations}, 2021.

\bibitem{zhang2022class}
J.~Zhang, J.~Zou, Z.~Su, J.~Tang, Y.~Kang, H.~Xu, Z.~Liu, and S.~Fan, ``A class-aware supervised contrastive learning framework for imbalanced fault diagnosis,'' \emph{Knowledge-Based Systems}, vol. 252, p. 109437, 2022.

\bibitem{jiang2021guided}
L.~Jiang, S.~Shi, Z.~Tian, X.~Lai, S.~Liu, C.-W. Fu, and J.~Jia, ``Guided point contrastive learning for semi-supervised point cloud semantic segmentation,'' in \emph{Proceedings of the IEEE/CVF international conference on computer vision}, 2021, pp. 6423--6432.

\bibitem{cui2021parametric}
J.~Cui, Z.~Zhong, S.~Liu, B.~Yu, and J.~Jia, ``Parametric contrastive learning,'' in \emph{Proceedings of the IEEE/CVF international conference on computer vision}, 2021, pp. 715--724.

\bibitem{zhu2022balanced}
J.~Zhu, Z.~Wang, J.~Chen, Y.-P.~P. Chen, and Y.-G. Jiang, ``Balanced contrastive learning for long-tailed visual recognition,'' in \emph{Proceedings of the IEEE/CVF Conference on Computer Vision and Pattern Recognition}, 2022, pp. 6908--6917.

\bibitem{li2022targeted}
T.~Li, P.~Cao, Y.~Yuan, L.~Fan, Y.~Yang, R.~S. Feris, P.~Indyk, and D.~Katabi, ``Targeted supervised contrastive learning for long-tailed recognition,'' in \emph{Proceedings of the IEEE/CVF Conference on Computer Vision and Pattern Recognition}, 2022, pp. 6918--6928.

\bibitem{gao2022ensemble}
X.~Gao, X.~Jia, J.~Liu, B.~Xue, Z.~Huang, S.~Fu, G.~Zhang, and K.~Li, ``An ensemble contrastive classification framework for imbalanced learning with sample-neighbors pair construction,'' \emph{Knowledge-Based Systems}, vol. 249, p. 109007, 2022.

\bibitem{gao2023imbalanced}
X.~Gao, Z.~Meng, X.~Jia, J.~Liu, X.~Diao, B.~Xue, Z.~Huang, and K.~Li, ``An imbalanced binary classification method based on contrastive learning using multi-label confidence comparisons within sample-neighbors pair,'' \emph{Neurocomputing}, vol. 517, pp. 148--164, 2023.

\bibitem{yang2022proco}
Z.~Yang, J.~Pan, Y.~Yang, X.~Shi, H.-Y. Zhou, Z.~Zhang, and C.~Bian, ``Proco: Prototype-aware contrastive learning for long-tailed medical image classification,'' in \emph{Medical Image Computing and Computer Assisted Intervention--MICCAI 2022: 25th International Conference, Singapore, September 18--22, 2022, Proceedings, Part VIII}.\hskip 1em plus 0.5em minus 0.4em\relax Springer, 2022, pp. 173--182.

\bibitem{snell2017prototypical}
J.~Snell, K.~Swersky, and R.~Zemel, ``Prototypical networks for few-shot learning,'' \emph{Advances in neural information processing systems}, vol.~30, 2017.

\bibitem{wang2021contrastive}
P.~Wang, K.~Han, X.-S. Wei, L.~Zhang, and L.~Wang, ``Contrastive learning based hybrid networks for long-tailed image classification,'' in \emph{Proceedings of the IEEE/CVF conference on computer vision and pattern recognition}, 2021, pp. 943--952.

\bibitem{hong2024proaug}
Y.~Hong, J.~Zhang, Z.~Sun, and K.~Yan, ``Proaug: Prototype-based augmentation for long-tailed image classification,'' in \emph{ICASSP 2024-2024 IEEE International Conference on Acoustics, Speech and Signal Processing (ICASSP)}.\hskip 1em plus 0.5em minus 0.4em\relax IEEE, 2024, pp. 3035--3039.

\bibitem{liu2019large}
Z.~Liu, Z.~Miao, X.~Zhan, J.~Wang, B.~Gong, and S.~X. Yu, ``Large-scale long-tailed recognition in an open world,'' in \emph{Proceedings of the IEEE/CVF conference on computer vision and pattern recognition}, 2019, pp. 2537--2546.

\bibitem{zeng2021pan}
Z.~Zeng, S.~Wang, N.~Xu, and W.~Mao, ``Pan: Prototype-based adaptive network for robust cross-modal retrieval,'' in \emph{Proceedings of the 44th International ACM SIGIR Conference on Research and Development in Information Retrieval}, 2021, pp. 1125--1134.

\bibitem{fu2022meta}
S.~Fu, H.~Chu, X.~He, H.~Wang, Z.~Yang, and H.~Hu, ``Meta-prototype decoupled training for long-tailed learning,'' in \emph{Proceedings of the Asian Conference on Computer Vision}, 2022, pp. 569--585.

\bibitem{zhu2020inflated}
L.~Zhu and Y.~Yang, ``Inflated episodic memory with region self-attention for long-tailed visual recognition,'' in \emph{Proceedings of the IEEE/CVF Conference on Computer Vision and Pattern Recognition}, 2020, pp. 4344--4353.

\bibitem{he2020learning}
T.~He, D.~Gong, Z.~Tian, and C.~Shen, ``Learning and memorizing representative prototypes for 3d point cloud semantic and instance segmentation,'' in \emph{Computer Vision--ECCV 2020: 16th European Conference, Glasgow, UK, August 23--28, 2020, Proceedings, Part XVIII 16}.\hskip 1em plus 0.5em minus 0.4em\relax Springer, 2020, pp. 564--580.

\bibitem{yin2019feature}
X.~Yin, X.~Yu, K.~Sohn, X.~Liu, and M.~Chandraker, ``Feature transfer learning for face recognition with under-represented data,'' in \emph{Proceedings of the IEEE/CVF conference on computer vision and pattern recognition}, 2019, pp. 5704--5713.

\bibitem{liu2020deep}
J.~Liu, Y.~Sun, C.~Han, Z.~Dou, and W.~Li, ``Deep representation learning on long-tailed data: A learnable embedding augmentation perspective,'' in \emph{Proceedings of the IEEE/CVF conference on computer vision and pattern recognition}, 2020, pp. 2970--2979.

\bibitem{chu2020feature}
P.~Chu, X.~Bian, S.~Liu, and H.~Ling, ``Feature space augmentation for long-tailed data,'' in \emph{Computer Vision--ECCV 2020: 16th European Conference, Glasgow, UK, August 23--28, 2020, Proceedings, Part XXIX 16}.\hskip 1em plus 0.5em minus 0.4em\relax Springer, 2020, pp. 694--710.

\bibitem{wang2021rsg}
J.~Wang, T.~Lukasiewicz, X.~Hu, J.~Cai, and Z.~Xu, ``Rsg: A simple but effective module for learning imbalanced datasets,'' in \emph{Proceedings of the IEEE/CVF Conference on Computer Vision and Pattern Recognition}, 2021, pp. 3784--3793.

\bibitem{kim2020m2m}
J.~Kim, J.~Jeong, and J.~Shin, ``M2m: Imbalanced classification via major-to-minor translation,'' in \emph{Proceedings of the IEEE/CVF conference on computer vision and pattern recognition}, 2020, pp. 13\,896--13\,905.

\bibitem{hong2022safa}
Y.~Hong, J.~Zhang, Z.~Sun, and K.~Yan, ``Safa: Sample-adaptive feature augmentation for long-tailed image classification,'' in \emph{European conference on computer vision}.\hskip 1em plus 0.5em minus 0.4em\relax Springer, 2022, pp. 587--603.

\bibitem{wang2017learning}
Y.-X. Wang, D.~Ramanan, and M.~Hebert, ``Learning to model the tail,'' \emph{Advances in neural information processing systems}, vol.~30, 2017.

\bibitem{liu2021gistnet}
B.~Liu, H.~Li, H.~Kang, G.~Hua, and N.~Vasconcelos, ``Gistnet: a geometric structure transfer network for long-tailed recognition,'' in \emph{Proceedings of the IEEE/CVF International Conference on Computer Vision}, 2021, pp. 8209--8218.

\bibitem{tradaboost}
W.~Dai, Q.~Yang, G.-R. Xue, and Y.~Yu, ``Boosting for transfer learning,'' in \emph{Proceedings of the 24th International Conference on Machine Learning}, ser. ICML '07.\hskip 1em plus 0.5em minus 0.4em\relax New York, NY, USA: Association for Computing Machinery, 2007, p. 193–200.

\bibitem{lee2020l2b}
H.~B. Lee, H.~Lee, D.~Na, S.~Kim, M.~Park, E.~Yang, and S.~J. Hwang, ``Learning to balance: Bayesian meta-learning for imbalanced and out-of-distribution tasks,'' in \emph{ICLR}, 2020.

\bibitem{jamal2020rethinking}
M.~A. Jamal, M.~Brown, M.-H. Yang, L.~Wang, and B.~Gong, ``Rethinking class-balanced methods for long-tailed visual recognition from a domain adaptation perspective,'' in \emph{Proceedings of the IEEE/CVF Conference on Computer Vision and Pattern Recognition}, 2020, pp. 7610--7619.

\bibitem{zhang2021learning}
Z.~Zhang and T.~Pfister, ``Learning fast sample re-weighting without reward data,'' in \emph{Proceedings of the IEEE/CVF International Conference on Computer Vision}, 2021, pp. 725--734.

\bibitem{liu2021improving}
Y.~Liu, B.~Cao, and J.~Fan, ``Improving the accuracy of learning example weights for imbalance classification,'' in \emph{International Conference on Learning Representations}, 2021.

\bibitem{liu2020mesa}
Z.~Liu, P.~Wei, J.~Jiang, W.~Cao, J.~Bian, and Y.~Chang, ``Mesa: boost ensemble imbalanced learning with meta-sampler,'' \emph{Advances in neural information processing systems}, vol.~33, pp. 14\,463--14\,474, 2020.

\bibitem{li2021metasaug}
S.~Li, K.~Gong, C.~H. Liu, Y.~Wang, F.~Qiao, and X.~Cheng, ``Metasaug: Meta semantic augmentation for long-tailed visual recognition,'' in \emph{Proceedings of the IEEE/CVF conference on computer vision and pattern recognition}, 2021, pp. 5212--5221.

\bibitem{wang2019implicit}
Y.~Wang, X.~Pan, S.~Song, H.~Zhang, G.~Huang, and C.~Wu, ``Implicit semantic data augmentation for deep networks,'' \emph{Advances in Neural Information Processing Systems}, vol.~32, 2019.

\bibitem{kang2019decoupling}
B.~Kang, S.~Xie, M.~Rohrbach, Z.~Yan, A.~Gordo, J.~Feng, and Y.~Kalantidis, ``Decoupling representation and classifier for long-tailed recognition,'' \emph{arXiv preprint arXiv:1910.09217}, 2019.

\bibitem{cao2019learning}
K.~Cao, C.~Wei, A.~Gaidon, N.~Arechiga, and T.~Ma, ``Learning imbalanced datasets with label-distribution-aware margin loss,'' \emph{Advances in neural information processing systems}, vol.~32, 2019.

\bibitem{kang2022comprehensive}
N.~Kang, H.~Chang, B.~Ma, and S.~Shan, ``A comprehensive framework for long-tailed learning via pretraining and normalization,'' \emph{IEEE Transactions on Neural Networks and Learning Systems}, 2022.

\bibitem{sunrethinking}
S.~Sun, H.~Lu, J.~Li, Y.~Xie, T.~Li, X.~Yang, L.~Zhang, and J.~Yan, ``Rethinking classifier re-training in long-tailed recognition: Label over-smooth can balance,'' in \emph{The Thirteenth International Conference on Learning Representations}, 2025.

\bibitem{liuself}
H.~Liu, J.~Z. HaoChen, A.~Gaidon, and T.~Ma, ``Self-supervised learning is more robust to dataset imbalance,'' \emph{arXiv preprint arXiv:2110.05025}, 2021.

\bibitem{li2021self}
T.~Li, L.~Wang, and G.~Wu, ``Self supervision to distillation for long-tailed visual recognition,'' in \emph{Proceedings of the IEEE/CVF international conference on computer vision}, 2021, pp. 630--639.

\bibitem{hendrycks2019using}
D.~Hendrycks, K.~Lee, and M.~Mazeika, ``Using pre-training can improve model robustness and uncertainty,'' in \emph{International Conference on Machine Learning}.\hskip 1em plus 0.5em minus 0.4em\relax PMLR, 2019, pp. 2712--2721.

\bibitem{ouyang2016factors}
W.~Ouyang, X.~Wang, C.~Zhang, and X.~Yang, ``Factors in finetuning deep model for object detection with long-tail distribution,'' in \emph{Proceedings of the IEEE conference on computer vision and pattern recognition}, 2016, pp. 864--873.

\bibitem{cui2018large}
Y.~Cui, Y.~Song, C.~Sun, A.~Howard, and S.~Belongie, ``Large scale fine-grained categorization and domain-specific transfer learning,'' in \emph{Proceedings of the IEEE conference on computer vision and pattern recognition}, 2018, pp. 4109--4118.

\bibitem{singh2020imbalanced}
R.~Singh, T.~Ahmed, A.~Kumar, A.~K. Singh, A.~K. Pandey, and S.~K. Singh, ``Imbalanced breast cancer classification using transfer learning,'' \emph{IEEE/ACM transactions on computational biology and bioinformatics}, vol.~18, no.~1, pp. 83--93, 2020.

\bibitem{bengio2009curriculum}
Y.~Bengio, J.~Louradour, R.~Collobert, and J.~Weston, ``Curriculum learning,'' in \emph{Proceedings of the 26th annual international conference on machine learning}, 2009, pp. 41--48.

\bibitem{jesson2017cased}
A.~Jesson, N.~Guizard, S.~H. Ghalehjegh, D.~Goblot, F.~Soudan, and N.~Chapados, ``Cased: curriculum adaptive sampling for extreme data imbalance,'' in \emph{Medical Image Computing and Computer Assisted Intervention- MICCAI 2017: 20th International Conference, Quebec City, QC, Canada, September 11-13, 2017, Proceedings, Part III 20}.\hskip 1em plus 0.5em minus 0.4em\relax Springer, 2017, pp. 639--646.

\bibitem{guo2018curriculumnet}
S.~Guo, W.~Huang, H.~Zhang, C.~Zhuang, D.~Dong, M.~R. Scott, and D.~Huang, ``Curriculumnet: Weakly supervised learning from large-scale web images,'' in \emph{Proceedings of the European conference on computer vision (ECCV)}, 2018, pp. 135--150.

\bibitem{wang2019dynamic}
Y.~Wang, W.~Gan, J.~Yang, W.~Wu, and J.~Yan, ``Dynamic curriculum learning for imbalanced data classification,'' in \emph{Proceedings of the IEEE/CVF international conference on computer vision}, 2019, pp. 5017--5026.

\bibitem{zhao2022diagnosing}
R.~Zhao, X.~Chen, Z.~Chen, and S.~Li, ``Diagnosing glaucoma on imbalanced data with self-ensemble dual-curriculum learning,'' \emph{Medical Image Analysis}, vol.~75, p. 102295, 2022.

\bibitem{wallace2012class}
B.~C. Wallace and I.~J. Dahabreh, ``Class probability estimates are unreliable for imbalanced data (and how to fix them),'' in \emph{2012 IEEE 12th international conference on data mining}.\hskip 1em plus 0.5em minus 0.4em\relax IEEE, 2012, pp. 695--704.

\bibitem{zhang2021disalign}
S.~Zhang, Z.~Li, S.~Yan, X.~He, and J.~Sun, ``Distribution alignment: A unified framework for long-tail visual recognition.'' in \emph{CVPR}, 2021.

\bibitem{li2022long}
M.~Li, Y.-m. Cheung, and Y.~Lu, ``Long-tailed visual recognition via gaussian clouded logit adjustment,'' in \emph{Proceedings of the IEEE/CVF Conference on Computer Vision and Pattern Recognition (CVPR)}, June 2022, pp. 6929--6938.

\bibitem{hong2021disentangling}
Y.~Hong, S.~Han, K.~Choi, S.~Seo, B.~Kim, and B.~Chang, ``Disentangling label distribution for long-tailed visual recognition,'' in \emph{Proceedings of the IEEE/CVF Conference on Computer Vision and Pattern Recognition (CVPR)}, June 2021, pp. 6626--6636.

\bibitem{menon2021long}
A.~K. Menon, S.~Jayasumana, A.~S. Rawat, H.~J. andAndreas Veit, and S.~Kumar, ``Long-tail learning via logit adjustment,'' in \emph{9th International Conference on Learning Representations (ICLR)}, 2021.

\bibitem{ye2022identifying}
H.-J. Ye, H.-Y. Chen, D.-C. Zhan, and W.-L. Chao, ``Identifying and compensating for feature deviation in imbalanced deep learning,'' \emph{arXiv preprint arXiv:2001.01385}, 2020.

\bibitem{tian2020posterior}
J.~Tian, Y.-C. Liu, N.~Glaser, Y.-C. Hsu, and Z.~Kira, ``Posterior re-calibration for imbalanced datasets,'' in \emph{Advances in Neural Information Processing Systems}, H.~Larochelle, M.~Ranzato, R.~Hadsell, M.~Balcan, and H.~Lin, Eds., vol.~33, 2020, pp. 8101--8113.

\bibitem{wang2024unified}
Z.~Wang, Q.~Xu, Z.~Yang, Y.~He, X.~Cao, and Q.~Huang, ``A unified generalization analysis of re-weighting and logit-adjustment for imbalanced learning,'' \emph{Advances in Neural Information Processing Systems}, vol.~36, 2024.

\bibitem{breiman1996bagging}
L.~Breiman, ``Bagging predictors,'' \emph{Machine learning}, vol.~24, pp. 123--140, 1996.

\bibitem{ref_22}
S.~Wang and X.~Yao, ``Diversity analysis on imbalanced data sets by using ensemble models,'' in \emph{2009 IEEE symposium on computational intelligence and data mining}.\hskip 1em plus 0.5em minus 0.4em\relax IEEE, 2009, pp. 324--331.

\bibitem{hido2009roughly}
S.~Hido, H.~Kashima, and Y.~Takahashi, ``Roughly balanced bagging for imbalanced data,'' \emph{Statistical Analysis and Data Mining: The ASA Data Science Journal}, vol.~2, no. 5-6, pp. 412--426, 2009.

\bibitem{blaszczynski2015neighbourhood}
J.~B{\l}aszczy{\'n}ski and J.~Stefanowski, ``Neighbourhood sampling in bagging for imbalanced data,'' \emph{Neurocomputing}, vol. 150, pp. 529--542, 2015.

\bibitem{hastie2009multi}
T.~Hastie, S.~Rosset, J.~Zhu, and H.~Zou, ``Multi-class adaboost,'' \emph{Statistics and its Interface}, vol.~2, no.~3, pp. 349--360, 2009.

\bibitem{friedman2001greedy}
J.~H. Friedman, ``Greedy function approximation: a gradient boosting machine,'' \emph{Annals of statistics}, pp. 1189--1232, 2001.

\bibitem{chen2016xgboost}
T.~Chen and C.~Guestrin, ``Xgboost: A scalable tree boosting system,'' in \emph{Proceedings of the 22nd acm sigkdd international conference on knowledge discovery and data mining}, 2016, pp. 785--794.

\bibitem{ref_3}
N.~V. Chawla, A.~Lazarevic, L.~O. Hall, and K.~W. Bowyer, ``Smoteboost: Improving prediction of the minority class in boosting,'' in \emph{European conference on principles of data mining and knowledge discovery}.\hskip 1em plus 0.5em minus 0.4em\relax Springer, 2003, pp. 107--119.

\bibitem{ref_8}
S.~Hu, Y.~Liang, L.~Ma, and Y.~He, ``Msmote: Improving classification performance when training data is imbalanced,'' in \emph{2009 second international workshop on computer science and engineering}, vol.~2.\hskip 1em plus 0.5em minus 0.4em\relax IEEE, 2009, pp. 13--17.

\bibitem{chen2010ramoboost}
S.~Chen, H.~He, and E.~A. Garcia, ``Ramoboost: Ranked minority oversampling in boosting,'' \emph{IEEE Transactions on Neural Networks}, vol.~21, no.~10, pp. 1624--1642, 2010.

\bibitem{ref_20}
C.~Seiffert, T.~M. Khoshgoftaar, J.~Van~Hulse, and A.~Napolitano, ``Rusboost: A hybrid approach to alleviating class imbalance,'' \emph{IEEE Transactions on Systems, Man, and Cybernetics-Part A: Systems and Humans}, vol.~40, no.~1, pp. 185--197, 2009.

\bibitem{ref_21}
M.~Galar, A.~Fern{\'a}ndez, E.~Barrenechea, and F.~Herrera, ``Eusboost: Enhancing ensembles for highly imbalanced data-sets by evolutionary undersampling,'' \emph{Pattern recognition}, vol.~46, no.~12, pp. 3460--3471, 2013.

\bibitem{gong2017rhsboost}
J.~Gong and H.~Kim, ``Rhsboost: Improving classification performance in imbalance data,'' \emph{Computational Statistics \& Data Analysis}, vol. 111, pp. 1--13, 2017.

\bibitem{lunardon2014rose}
N.~Lunardon, G.~Menardi, and N.~Torelli, ``Rose: a package for binary imbalanced learning.'' \emph{R journal}, vol.~6, no.~1, 2014.

\bibitem{ahmed2019liuboost}
S.~Ahmed, F.~Rayhan, A.~Mahbub, M.~Rafsan~Jani, S.~Shatabda, and D.~M. Farid, ``Liuboost: locality informed under-boosting for imbalanced data classification,'' in \emph{Emerging Technologies in Data Mining and Information Security: Proceedings of IEMIS 2018, Volume 2}.\hskip 1em plus 0.5em minus 0.4em\relax Springer, 2019, pp. 133--144.

\bibitem{popel2018hybrid}
M.~H. Popel, K.~M. Hasib, S.~A. Habib, and F.~M. Shah, ``A hybrid under-sampling method (husboost) to classify imbalanced data,'' in \emph{2018 21st international conference of computer and information technology (ICCIT)}.\hskip 1em plus 0.5em minus 0.4em\relax IEEE, 2018, pp. 1--7.

\bibitem{ling2008cost}
C.~X. Ling and V.~S. Sheng, ``Cost-sensitive learning and the class imbalance problem,'' \emph{Encyclopedia of machine learning}, vol. 2011, pp. 231--235, 2008.

\bibitem{fan1999adacost}
W.~Fan, S.~J. Stolfo, J.~Zhang, and P.~K. Chan, ``Adacost: misclassification cost-sensitive boosting,'' in \emph{Icml}, vol.~99, 1999, pp. 97--105.

\bibitem{ting2000comparative}
K.~M. Ting, ``A comparative study of cost-sensitive boosting algorithms,'' in \emph{Proc. of the 17th International Conference on Machine Learning (ICML), 2000}, 2000.

\bibitem{sun2007cost}
Y.~Sun, M.~S. Kamel, A.~K. Wong, and Y.~Wang, ``Cost-sensitive boosting for classification of imbalanced data,'' \emph{Pattern recognition}, vol.~40, no.~12, pp. 3358--3378, 2007.

\bibitem{gou2021knowledge}
J.~Gou, B.~Yu, S.~J. Maybank, and D.~Tao, ``Knowledge distillation: A survey,'' \emph{International Journal of Computer Vision}, vol. 129, pp. 1789--1819, 2021.

\bibitem{asif2019ensemble}
U.~Asif, J.~Tang, and S.~Harrer, ``Ensemble knowledge distillation for learning improved and efficient networks,'' \emph{arXiv preprint arXiv:1909.08097}, 2019.

\bibitem{chebotar2016distilling}
Y.~Chebotar and A.~Waters, ``Distilling knowledge from ensembles of neural networks for speech recognition.'' in \emph{Interspeech}, 2016, pp. 3439--3443.

\bibitem{fukuda2017efficient}
T.~Fukuda, M.~Suzuki, G.~Kurata, S.~Thomas, J.~Cui, and B.~Ramabhadran, ``Efficient knowledge distillation from an ensemble of teachers.'' in \emph{Interspeech}, 2017, pp. 3697--3701.

\bibitem{liu2020adaptive}
Y.~Liu, W.~Zhang, and J.~Wang, ``Adaptive multi-teacher multi-level knowledge distillation,'' \emph{Neurocomputing}, vol. 415, pp. 106--113, 2020.

\bibitem{yuan2021reinforced}
F.~Yuan, L.~Shou, J.~Pei, W.~Lin, M.~Gong, Y.~Fu, and D.~Jiang, ``Reinforced multi-teacher selection for knowledge distillation,'' in \emph{Proceedings of the AAAI Conference on Artificial Intelligence}, vol.~35, no.~16, 2021, pp. 14\,284--14\,291.

\bibitem{zhao2020highlight}
H.~Zhao, X.~Sun, J.~Dong, C.~Chen, and Z.~Dong, ``Highlight every step: Knowledge distillation via collaborative teaching,'' \emph{IEEE Transactions on Cybernetics}, vol.~52, no.~4, pp. 2070--2081, 2020.

\bibitem{yang2020model}
Z.~Yang, L.~Shou, M.~Gong, W.~Lin, and D.~Jiang, ``Model compression with two-stage multi-teacher knowledge distillation for web question answering system,'' in \emph{Proceedings of the 13th International Conference on Web Search and Data Mining}, 2020, pp. 690--698.

\bibitem{wu2024collaborative}
F.~Wu, Z.~Liu, Z.~Zhang, J.~Liu, and L.~Wang, ``Collaborative global-local structure network with knowledge distillation for imbalanced data classification,'' \emph{IEEE Transactions on Circuits and Systems for Video Technology}, 2024.

\bibitem{pham2023collaborative}
C.~Pham, T.~Hoang, and T.-T. Do, ``Collaborative multi-teacher knowledge distillation for learning low bit-width deep neural networks,'' in \emph{Proceedings of the IEEE/CVF Winter Conference on Applications of Computer Vision}, 2023, pp. 6435--6443.

\bibitem{wu2020solving}
T.-Y. Wu, P.~Morgado, P.~Wang, C.-H. Ho, and N.~Vasconcelos, ``Solving long-tailed recognition with deep realistic taxonomic classifier,'' in \emph{Computer Vision--ECCV 2020: 16th European Conference, Glasgow, UK, August 23--28, 2020, Proceedings, Part VIII 16}.\hskip 1em plus 0.5em minus 0.4em\relax Springer, 2020, pp. 171--189.

\bibitem{wu2021adversarial}
T.~Wu, Z.~Liu, Q.~Huang, Y.~Wang, and D.~Lin, ``Adversarial robustness under long-tailed distribution,'' in \emph{Proceedings of the IEEE/CVF conference on computer vision and pattern recognition}, 2021, pp. 8659--8668.

\bibitem{du2023global}
F.~Du, P.~Yang, Q.~Jia, F.~Nan, X.~Chen, and Y.~Yang, ``Global and local mixture consistency cumulative learning for long-tailed visual recognitions,'' in \emph{Proceedings of the IEEE/CVF Conference on Computer Vision and Pattern Recognition}, 2023, pp. 15\,814--15\,823.

\bibitem{ma2021simple}
T.~Ma, S.~Geng, M.~Wang, J.~Shao, J.~Lu, H.~Li, P.~Gao, and Y.~Qiao, ``A simple long-tailed recognition baseline via vision-language model,'' \emph{arXiv preprint arXiv:2111.14745}, 2021.

\bibitem{shi2023parameter}
J.-X. Shi, T.~Wei, Z.~Zhou, X.-Y. Han, J.-J. Shao, and Y.-F. Li, ``Parameter-efficient long-tailed recognition,'' \emph{arXiv preprint arXiv:2309.10019}, 2023.

\bibitem{oksuz2020imbalance}
K.~Oksuz, B.~C. Cam, S.~Kalkan, and E.~Akbas, ``Imbalance problems in object detection: A review,'' \emph{IEEE transactions on pattern analysis and machine intelligence}, vol.~43, no.~10, pp. 3388--3415, 2020.

\bibitem{girshick2014rich}
R.~Girshick, J.~Donahue, T.~Darrell, and J.~Malik, ``Rich feature hierarchies for accurate object detection and semantic segmentation,'' in \emph{Proceedings of the IEEE conference on computer vision and pattern recognition}, 2014, pp. 580--587.

\bibitem{liu2016ssd}
W.~Liu, D.~Anguelov, D.~Erhan, C.~Szegedy, S.~Reed, C.-Y. Fu, and A.~C. Berg, ``Ssd: Single shot multibox detector,'' in \emph{Computer Vision--ECCV 2016: 14th European Conference, Amsterdam, The Netherlands, October 11--14, 2016, Proceedings, Part I 14}.\hskip 1em plus 0.5em minus 0.4em\relax Springer, 2016, pp. 21--37.

\bibitem{ross2017focal}
T.-Y. Ross and G.~Doll{\'a}r, ``Focal loss for dense object detection,'' in \emph{proceedings of the IEEE conference on computer vision and pattern recognition}, 2017, pp. 2980--2988.

\bibitem{li2019gradient}
B.~Li, Y.~Liu, and X.~Wang, ``Gradient harmonized single-stage detector,'' in \emph{Proceedings of the AAAI conference on artificial intelligence}, vol.~33, no.~01, 2019, pp. 8577--8584.

\bibitem{cao2020prime}
Y.~Cao, K.~Chen, C.~C. Loy, and D.~Lin, ``Prime sample attention in object detection,'' in \emph{Proceedings of the IEEE/CVF conference on computer vision and pattern recognition}, 2020, pp. 11\,583--11\,591.

\bibitem{chen2022residual}
J.~Chen, D.~Liu, B.~Luo, X.~Peng, T.~Xu, and E.~Chen, ``Residual objectness for imbalance reduction,'' \emph{Pattern Recognition}, vol. 130, p. 108781, 2022.

\bibitem{chen2019towards}
K.~Chen, J.~Li, W.~Lin, J.~See, J.~Wang, L.~Duan, Z.~Chen, C.~He, and J.~Zou, ``Towards accurate one-stage object detection with ap-loss,'' in \emph{Proceedings of the IEEE/CVF Conference on Computer Vision and Pattern Recognition}, 2019, pp. 5119--5127.

\bibitem{wang2017fast}
X.~Wang, A.~Shrivastava, and A.~Gupta, ``A-fast-rcnn: Hard positive generation via adversary for object detection,'' in \emph{Proceedings of the IEEE conference on computer vision and pattern recognition}, 2017, pp. 2606--2615.

\bibitem{tripathi2019learning}
S.~Tripathi, S.~Chandra, A.~Agrawal, A.~Tyagi, J.~M. Rehg, and V.~Chari, ``Learning to generate synthetic data via compositing,'' in \emph{Proceedings of the IEEE/CVF Conference on Computer Vision and Pattern Recognition}, 2019, pp. 461--470.

\bibitem{lin2017feature}
T.-Y. Lin, P.~Doll{\'a}r, R.~Girshick, K.~He, B.~Hariharan, and S.~Belongie, ``Feature pyramid networks for object detection,'' in \emph{Proceedings of the IEEE conference on computer vision and pattern recognition}, 2017, pp. 2117--2125.

\bibitem{sudre2017generalised}
C.~H. Sudre, W.~Li, T.~Vercauteren, S.~Ourselin, and M.~Jorge~Cardoso, ``Generalised dice overlap as a deep learning loss function for highly unbalanced segmentations,'' in \emph{Deep Learning in Medical Image Analysis and Multimodal Learning for Clinical Decision Support: Third International Workshop, DLMIA 2017, and 7th International Workshop, ML-CDS 2017, Held in Conjunction with MICCAI 2017, Qu{\'e}bec City, QC, Canada, September 14, Proceedings 3}.\hskip 1em plus 0.5em minus 0.4em\relax Springer, 2017, pp. 240--248.

\bibitem{crum2006generalized}
W.~R. Crum, O.~Camara, and D.~L. Hill, ``Generalized overlap measures for evaluation and validation in medical image analysis,'' \emph{IEEE transactions on medical imaging}, vol.~25, no.~11, pp. 1451--1461, 2006.

\bibitem{rota2017loss}
S.~Rota~Bulo, G.~Neuhold, and P.~Kontschieder, ``Loss max-pooling for semantic image segmentation,'' in \emph{Proceedings of the IEEE conference on computer vision and pattern recognition}, 2017, pp. 2126--2135.

\bibitem{taghanaki2019combo}
S.~A. Taghanaki, Y.~Zheng, S.~K. Zhou, B.~Georgescu, P.~Sharma, D.~Xu, D.~Comaniciu, and G.~Hamarneh, ``Combo loss: Handling input and output imbalance in multi-organ segmentation,'' \emph{Computerized Medical Imaging and Graphics}, vol.~75, pp. 24--33, 2019.

\bibitem{zhong2023understanding}
Z.~Zhong, J.~Cui, Y.~Yang, X.~Wu, X.~Qi, X.~Zhang, and J.~Jia, ``Understanding imbalanced semantic segmentation through neural collapse,'' in \emph{Proceedings of the IEEE/CVF conference on computer vision and pattern recognition}, 2023, pp. 19\,550--19\,560.

\bibitem{henning2022survey}
S.~Henning, W.~Beluch, A.~Fraser, and A.~Friedrich, ``A survey of methods for addressing class imbalance in deep-learning based natural language processing,'' \emph{arXiv preprint arXiv:2210.04675}, 2022.

\bibitem{zhao2024retrieval}
P.~Zhao, H.~Zhang, Q.~Yu, Z.~Wang, Y.~Geng, F.~Fu, L.~Yang, W.~Zhang, and B.~Cui, ``Retrieval-augmented generation for ai-generated content: A survey,'' \emph{arXiv preprint arXiv:2402.19473}, 2024.

\bibitem{wang2024qaencoder}
Z.~Wang, Q.~Yu, S.~Wei, Z.~Li, F.~Xiong, X.~Wang, S.~Niu, H.~Liang, and W.~Zhang, ``Qaencoder: Towards aligned representation learning in question answering system,'' \emph{arXiv preprint arXiv:2409.20434}, 2024.

\bibitem{ghosh2019imbalanced}
K.~Ghosh, A.~Banerjee, S.~Chatterjee, and S.~Sen, ``Imbalanced twitter sentiment analysis using minority oversampling,'' in \emph{2019 IEEE 10th international conference on awareness science and technology (iCAST)}.\hskip 1em plus 0.5em minus 0.4em\relax IEEE, 2019, pp. 1--5.

\bibitem{flores2018evaluation}
A.~C. Flores, R.~I. Icoy, C.~F. Pe{\~n}a, and K.~D. Gorro, ``An evaluation of svm and naive bayes with smote on sentiment analysis data set,'' in \emph{2018 International Conference on Engineering, Applied Sciences, and Technology (ICEAST)}.\hskip 1em plus 0.5em minus 0.4em\relax IEEE, 2018, pp. 1--4.

\bibitem{ardianto2020sentiment}
R.~Ardianto, T.~Rivanie, Y.~Alkhalifi, F.~S. Nugraha, and W.~Gata, ``Sentiment analysis on e-sports for education curriculum using naive bayes and support vector machine,'' \emph{Jurnal Ilmu Komputer dan Informasi}, vol.~13, no.~2, pp. 109--122, 2020.

\bibitem{al2017using}
S.~Al-Azani and E.-S.~M. El-Alfy, ``Using word embedding and ensemble learning for highly imbalanced data sentiment analysis in short arabic text,'' \emph{Procedia Computer Science}, vol. 109, pp. 359--366, 2017.

\bibitem{mohammad2016translation}
S.~M. Mohammad, M.~Salameh, and S.~Kiritchenko, ``How translation alters sentiment,'' \emph{Journal of Artificial Intelligence Research}, vol.~55, pp. 95--130, 2016.

\bibitem{guerra2014sentiment}
P.~C. Guerra, W.~Meira~Jr, and C.~Cardie, ``Sentiment analysis on evolving social streams: How self-report imbalances can help,'' in \emph{Proceedings of the 7th ACM international conference on Web search and data mining}, 2014, pp. 443--452.

\bibitem{obiedat2022sentiment}
R.~Obiedat, R.~Qaddoura, A.-Z. Ala’M, L.~Al-Qaisi, O.~Harfoushi, M.~Alrefai, and H.~Faris, ``Sentiment analysis of customers’ reviews using a hybrid evolutionary svm-based approach in an imbalanced data distribution,'' \emph{IEEE Access}, vol.~10, pp. 22\,260--22\,273, 2022.

\bibitem{dogra2022comparative}
V.~Dogra, S.~Verma, K.~Verma, N.~Jhanjhi, U.~Ghosh, and D.-N. Le, ``A comparative analysis of machine learning models for banking news extraction by multiclass classification with imbalanced datasets of financial news: challenges and solutions,'' \emph{International Journal of Interactive Multimedia and Artificial Intelligence}, 2022.

\bibitem{rivera2020news}
G.~Rivera, R.~Florencia, V.~Garc{\'\i}a, A.~Ruiz, and J.~P. S{\'a}nchez-Sol{\'\i}s, ``News classification for identifying traffic incident points in a spanish-speaking country: A real-world case study of class imbalance learning,'' \emph{Applied Sciences}, vol.~10, no.~18, p. 6253, 2020.

\bibitem{hossain2020banfakenews}
M.~Z. Hossain, M.~A. Rahman, M.~S. Islam, and S.~Kar, ``Banfakenews: A dataset for detecting fake news in bangla,'' \emph{arXiv preprint arXiv:2004.08789}, 2020.

\bibitem{keya2022augfake}
A.~J. Keya, M.~A.~H. Wadud, M.~Mridha, M.~Alatiyyah, and M.~A. Hamid, ``Augfake-bert: handling imbalance through augmentation of fake news using bert to enhance the performance of fake news classification,'' \emph{Applied Sciences}, vol.~12, no.~17, p. 8398, 2022.

\bibitem{hasib2023strategies}
K.~M. Hasib, N.~A. Towhid, K.~O. Faruk, J.~Al~Mahmud, and M.~Mridha, ``Strategies for enhancing the performance of news article classification in bangla: Handling imbalance and interpretation,'' \emph{Engineering Applications of Artificial Intelligence}, vol. 125, p. 106688, 2023.

\bibitem{zabir_al_nazi_2020}
\BIBentryALTinterwordspacing
Z.~A. Nazi, ``Bangla newspaper dataset,'' 2020. [Online]. Available: \url{https://www.kaggle.com/dsv/1576225}
\BIBentrySTDinterwordspacing

\bibitem{raffel2020exploring}
C.~Raffel, N.~Shazeer, A.~Roberts, K.~Lee, S.~Narang, M.~Matena, and P.~J. Liu, ``Exploring the limits of transfer learning with a unified text-to-text transformer,'' \emph{Journal of Machine Learning Research}, vol.~21, no. 140, pp. 1--67, 2020.

\bibitem{li2023from}
M.~Li, Y.~Zhang, Z.~Li, J.~Chen, L.~Chen, N.~Cheng, and J.~Xiao, ``From quantity to quality: Boosting llm performance with self-guided data selection for instruction tuning,'' \emph{arXiv preprint arXiv:2308.12032}, 2023.

\bibitem{li2024numinamath}
J.~Li, E.~Beeching, L.~Tunstall, B.~Lipkin, R.~Soletskyi, S.~Huang, and S.~Polu, ``Numinamath: The largest public dataset in ai4maths with 860k pairs of competition math problems and solutions,'' 2024, hugging Face repository.

\bibitem{an2023learning}
S.~An, Z.~Ma, Z.~Lin, N.~Zheng, J.~G. Lou, and W.~Chen, ``Learning from mistakes makes llm better reasoner,'' \emph{arXiv preprint arXiv:2310.20689}, 2023.

\bibitem{shao2024distributed}
Y.~Shao, H.~Li, X.~Gu, H.~Yin, Y.~Li, X.~Miao, W.~Zhang, B.~Cui, and L.~Chen, ``Distributed graph neural network training: A survey,'' \emph{ACM Computing Surveys}, vol.~56, no.~8, pp. 1--39, 2024.

\bibitem{gao2023accelerating}
X.~Gao, W.~Zhang, J.~Yu, Y.~Shao, Q.~V.~H. Nguyen, B.~Cui, and H.~Yin, ``Accelerating scalable graph neural network inference with node-adaptive propagation,'' in \emph{IEEE 40th International Conference on Data Engineering}, 2024, pp. 3042--3055.

\bibitem{gao2024contrastive}
X.~Gao, Y.~Li, T.~Chen, G.~Ye, W.~Zhang, and H.~Yin, ``Contrastive graph condensation: Advancing data versatility through self-supervised learning,'' \emph{arXiv preprint arXiv:2411.17063}, 2024.

\bibitem{gao2024rethinking}
X.~Gao, T.~Chen, W.~Zhang, J.~Yu, G.~Ye, Q.~V.~H. Nguyen, and H.~Yin, ``Rethinking and accelerating graph condensation: A training-free approach with class partition,'' in \emph{Proceedings of the ACM Web Conference}, 2025.

\bibitem{gao2024robgc}
X.~Gao, H.~Yin, T.~Chen, G.~Ye, W.~Zhang, and B.~Cui, ``Robgc: Towards robust graph condensation,'' \emph{{arXiv}}, 2024.

\bibitem{gao2023semantic}
X.~Gao, W.~Zhang, T.~Chen, J.~Yu, H.~Q.~V. Nguyen, and H.~Yin, ``Semantic-aware node synthesis for imbalanced heterogeneous information networks,'' in \emph{Proceedings of the 32nd ACM International Conference on Information and Knowledge Management}, 2023, pp. 545--555.

\bibitem{gao2024graphkdd}
X.~Gao, T.~Chen, W.~Zhang, Y.~Li, X.~Sun, and H.~Yin, ``Graph condensation for open-world graph learning,'' in \emph{Proceedings of the 30th ACM SIGKDD Conference on Knowledge Discovery and Data Mining}, 2024, pp. 851--862.

\bibitem{gao2024graph}
X.~Gao, J.~Yu, T.~Chen, G.~Ye, W.~Zhang, and H.~Yin, ``Graph condensation: A survey,'' \emph{IEEE Transactions on Knowledge and Data Engineering}, 2025.

\bibitem{wu2022graphmixup}
L.~Wu, J.~Xia, Z.~Gao, H.~Lin, C.~Tan, and S.~Z. Li, ``Graphmixup: Improving class-imbalanced node classification by reinforcement mixup and self-supervised context prediction,'' in \emph{Joint European conference on machine learning and knowledge discovery in databases}.\hskip 1em plus 0.5em minus 0.4em\relax Springer, 2022, pp. 519--535.

\bibitem{park2021graphens}
J.~Park, J.~Song, and E.~Yang, ``Graphens: Neighbor-aware ego network synthesis for class-imbalanced node classification,'' in \emph{International conference on learning representations}, 2021.

\bibitem{li2023graphsha}
W.-Z. Li, C.-D. Wang, H.~Xiong, and J.-H. Lai, ``Graphsha: Synthesizing harder samples for class-imbalanced node classification,'' in \emph{Proceedings of the 29th ACM SIGKDD conference on knowledge discovery and data mining}, 2023, pp. 1328--1340.

\bibitem{qu2021imgagn}
L.~Qu, H.~Zhu, R.~Zheng, Y.~Shi, and H.~Yin, ``Imgagn: Imbalanced network embedding via generative adversarial graph networks,'' in \emph{Proceedings of the 27th ACM SIGKDD conference on knowledge discovery \& data mining}, 2021, pp. 1390--1398.

\bibitem{zhou2023graphsr}
M.~Zhou and Z.~Gong, ``Graphsr: A data augmentation algorithm for imbalanced node classification,'' in \emph{Proceedings of the AAAI Conference on artificial intelligence}, vol.~37, no.~4, 2023, pp. 4954--4962.

\bibitem{zhang2024bim}
W.~Zhang, X.~Gao, L.~Yang, M.~Cao, P.~Huang, J.~Shan, H.~Yin, and B.~Cui, ``Bim: improving graph neural networks with balanced influence maximization,'' in \emph{IEEE 40th International Conference on Data Engineering (ICDE)}, 2024, pp. 2931--2944.

\bibitem{yun2022lte4g}
S.~Yun, K.~Kim, K.~Yoon, and C.~Park, ``Lte4g: Long-tail experts for graph neural networks,'' in \emph{Proceedings of the 31st ACM International Conference on Information \& Knowledge Management}, 2022, pp. 2434--2443.

\bibitem{liu2021tail}
Z.~Liu, T.-K. Nguyen, and Y.~Fang, ``Tail-gnn: Tail-node graph neural networks,'' in \emph{Proceedings of the 27th ACM SIGKDD conference on knowledge discovery \& data mining}, 2021, pp. 1109--1119.

\bibitem{chen2021topology}
D.~Chen, Y.~Lin, G.~Zhao, X.~Ren, P.~Li, J.~Zhou, and X.~Sun, ``Topology-imbalance learning for semi-supervised node classification,'' \emph{Advances in Neural Information Processing Systems}, vol.~34, pp. 29\,885--29\,897, 2021.

\bibitem{sun2022position}
Q.~Sun, J.~Li, H.~Yuan, X.~Fu, H.~Peng, C.~Ji, Q.~Li, and P.~S. Yu, ``Position-aware structure learning for graph topology-imbalance by relieving under-reaching and over-squashing,'' in \emph{Proceedings of the 31st ACM International Conference on Information \& Knowledge Management}, 2022, pp. 1848--1857.

\bibitem{zhu2023few}
X.~Zhu, P.~Luo, Z.~Zhao, T.~Xu, A.~Lizhiyu, Y.~Yu, X.~Li, and E.~Chen, ``Few-shot link prediction for event-based social networks via meta-learning,'' in \emph{International Conference on Database Systems for Advanced Applications}.\hskip 1em plus 0.5em minus 0.4em\relax Springer, 2023, pp. 31--41.

\bibitem{liu2022deep}
J.~Liu, F.~Xia, X.~Feng, J.~Ren, and H.~Liu, ``Deep graph learning for anomalous citation detection,'' \emph{IEEE Transactions on Neural Networks and Learning Systems}, vol.~33, no.~6, pp. 2543--2557, 2022.

\bibitem{li2021live}
Z.~Li, H.~Wang, P.~Zhang, P.~Hui, J.~Huang, J.~Liao, J.~Zhang, and J.~Bu, ``Live-streaming fraud detection: A heterogeneous graph neural network approach,'' in \emph{Proceedings of the 27th ACM SIGKDD Conference on Knowledge Discovery \& Data Mining}, 2021, pp. 3670--3678.

\bibitem{wang2022imbalanced}
Y.~Wang, Y.~Zhao, N.~Shah, and T.~Derr, ``Imbalanced graph classification via graph-of-graph neural networks,'' in \emph{Proceedings of the 31st ACM International Conference on Information \& Knowledge Management}, 2022, pp. 2067--2076.

\bibitem{liu2023semi}
G.~Liu, T.~Zhao, E.~Inae, T.~Luo, and M.~Jiang, ``Semi-supervised graph imbalanced regression,'' in \emph{Proceedings of the 29th ACM SIGKDD Conference on Knowledge Discovery and Data Mining}, 2023, pp. 1453--1465.

\bibitem{khoshgoftaar2007empirical}
T.~M. Khoshgoftaar, M.~Golawala, and J.~Van~Hulse, ``An empirical study of learning from imbalanced data using random forest,'' in \emph{19th IEEE international conference on tools with artificial intelligence (ICTAI 2007)}, vol.~2.\hskip 1em plus 0.5em minus 0.4em\relax IEEE, 2007, pp. 310--317.

\bibitem{JMLR:v18:16-365}
\BIBentryALTinterwordspacing
G.~Lema{{\^i}}tre, F.~Nogueira, and C.~K. Aridas, ``Imbalanced-learn: A python toolbox to tackle the curse of imbalanced datasets in machine learning,'' \emph{Journal of Machine Learning Research}, vol.~18, no.~17, pp. 1--5, 2017. [Online]. Available: \url{http://jmlr.org/papers/v18/16-365}
\BIBentrySTDinterwordspacing

\bibitem{santos_imbalance_overlap}
M.~Santos, ``Open-source imbalance-overlap,'' {https://github.com/miriamspsantos/open-source-imbalance-overlap}.

\bibitem{qin2024iglbench}
J.~Qin, H.~Yuan, Q.~Sun, L.~Xu, J.~Yuan, P.~Huang, and P.~S. Yu, ``Igl-bench: Establishing the comprehensive benchmark for imbalanced graph learning,'' \emph{arXiv preprint arXiv:2406.09870}, 2024.

\bibitem{liu2021imbens}
Z.~Liu, J.~Kang, H.~Tong, and Y.~Chang, ``Imbens: Ensemble class-imbalanced learning in python,'' \emph{arXiv preprint arXiv:2111.12776}, 2021.

\bibitem{sarig2023balance}
T.~Sarig, T.~Galili, and R.~Eilat, ``balance--a python package for balancing biased data samples,'' \emph{arXiv preprint arXiv:2307.06024}, 2023.

\bibitem{du2024semi}
G.~Du, J.~Zhang, N.~Zhang, H.~Wu, P.~Wu, and S.~Li, ``Semi-supervised imbalanced multi-label classification with label propagation,'' \emph{Pattern Recognition}, vol. 150, p. 110358, 2024.

\bibitem{gong2022ranksim}
Y.~Gong, G.~Mori, and F.~Tung, ``Ranksim: Ranking similarity regularization for deep imbalanced regression,'' \emph{arXiv preprint arXiv:2205.15236}, 2022.

\bibitem{yang2021delving}
Y.~Yang, K.~Zha, Y.~Chen, H.~Wang, and D.~Katabi, ``Delving into deep imbalanced regression,'' in \emph{International conference on machine learning}.\hskip 1em plus 0.5em minus 0.4em\relax PMLR, 2021, pp. 11\,842--11\,851.

\bibitem{qian2022co}
Y.~Qian, C.~Zhang, Y.~Zhang, Q.~Wen, Y.~Ye, and C.~Zhang, ``Co-modality graph contrastive learning for imbalanced node classification,'' \emph{Advances in Neural Information Processing Systems}, vol.~35, pp. 15\,862--15\,874, 2022.

\bibitem{kim2024epic}
J.~Kim, T.~Kim, and J.~Choo, ``Epic: Effective prompting for imbalanced-class data synthesis in tabular data classification via large language models,'' \emph{Advances in Neural Information Processing Systems}, vol.~37, pp. 31\,504--31\,542, 2024.

\bibitem{yang2024language}
J.~Y. Yang, G.~Park, J.~Kim, H.~Jang, and E.~Yang, ``Language-interfaced tabular oversampling via progressive imputation and self-authentication,'' in \emph{The Twelfth International Conference on Learning Representations}, 2024.

\bibitem{zhao2024improving}
H.~Zhao, H.~Chen, T.~A. Ruggles, Y.~Feng, D.~Singh, and H.-J. Yoon, ``Improving text classification with large language model-based data augmentation,'' \emph{Electronics}, vol.~13, no.~13, p. 2535, 2024.

\bibitem{vassilev2023meta}
A.~Vassilev, H.~Jin, and M.~Hasan, ``Meta learning with language models: Challenges and opportunities in the classification of imbalanced text,'' \emph{arXiv preprint arXiv:2310.15019}, 2023.

\end{thebibliography}
